%% file: _main.tex
\input{_constants}
\arxiv 

\pdfoutput=1
\documentclass[10pt,twocolumn,letterpaper]{article}
\input{cvpr_header}
\begin{document}
\title{\paperTitle}
\author{\authorBlock}
\maketitle

\input{00_abstract}
\input{01_intro}

\input{02_related}

\input{03_method}

\input{04_experiment}

\input{10_conclusion}

{\small
\bibliographystyle{ieee_fullname}
\bibliography{11_references}
}

\ifarxiv \clearpage \input{12_appendix} \fi

\end{document}

%% file: _constants.tex
\def\paperTitle{A Lightweight Domain Adaptive Absolute Pose Regressor Using \textsc{Barlow Twins} Objective}

\def\authorBlock{
    Praveen Kumar Rajendran$^{1}$ \qquad Quoc-Vinh Lai-Dang$^{1}$ \qquad Luiz Felipe Vecchietti$^{2}$ \qquad Dongsoo Har$^{1}$ \\
    {$^{1}$}KAIST, Daejeon, South Korea \qquad {$^{2}$}IBS, Daejeon, South Korea\\
    {\tt\small \{praveenkumar, ldqvinh, dshar\}@kaist.ac.kr \qquad lfelipesv@ibs.re.kr}
}

\newif\ifreview 
\newif\ifarxiv \newcommand{\arxiv}{\arxivtrue}
\newif\ifcamera 
\newif\ifrebuttal 

%% file: 00_abstract.tex
\begin{abstract}
Identifying the camera pose for a given image is a challenging problem with applications in robotics, autonomous vehicles, and augmented/virtual reality. Lately, learning-based methods have shown to be effective for absolute camera pose estimation. However, these methods are not accurate when generalizing to different domains. In this paper, a domain adaptive training framework for absolute pose regression is introduced. In the proposed framework, the scene image is augmented for different domains by using generative methods to train parallel branches using \textsc{Barlow Twins} objective. The parallel branches leverage a lightweight CNN-based absolute pose regressor architecture. Further, the efficacy of incorporating spatial and channel-wise attention in the regression head for rotation prediction is investigated. Our method is evaluated with two datasets, Cambridge landmarks and 7Scenes. The results demonstrate that, even with using roughly 24 times fewer FLOPs, 12 times fewer activations, and 5 times fewer parameters than MS-Transformer, our approach outperforms all the CNN-based architectures and achieves performance comparable to transformer-based architectures. Our method ranks $2^{nd}$ and $4^{th}$ with the Cambridge Landmarks and 7Scenes datasets, respectively. In addition, for augmented domains not encountered during training, our approach significantly outperforms the MS-transformer. Furthermore, it is shown that our domain adaptive framework achieves better performance than the single branch model trained with the identical CNN backbone with all instances of the unseen distribution.

\end{abstract}

%% file: 01_intro.tex
\section{Introduction}
\label{sec:intro}

Absolute camera pose estimation, e.g., predicting the 3-Dimensional position and orientation of a camera for a given image, is a challenging problem with numerous practical applications in computer vision, ranging from mobile robots~\cite{alatise2017pose} and autonomous vehicles~\cite{bresson2017simultaneous, kim2020pose} to augmented/virtual reality~\cite{chatzopoulos2017mobile}. Traditional camera pose estimation techniques such as structure-from-motion(SfM) enabled by perspective-n-point(PnP) and RANSAC~\cite{fischler1981random} tend to exploit multi-view geometry scene constraints, using 2D-3D correspondences for predictions~\cite{hartley2003multiple}. The SfM-based methods are usually dependent on image descriptors such as SIFT\cite{lowe2004distinctive}, ORB\cite{rublee2011orb}, and FAST\cite{philbin2007object}. However, these descriptors suffer poor performance in handling scenes with abrupt changes of illumination, occlusions, or repeated structures. In addition, the computational cost is significantly high. Another alternative solution to tackle this problem is image retrieval (IR) methods~\cite{torii201524, arandjelovic2016netvlad, sarlin2019coarse, taira2018inloc, noh2017large, sattler2019understanding}. For the IR methods such as NetVLAD~\cite{arandjelovic2016netvlad}, similar images to the query image are searched in the dataset and are used for feature extraction, matching, and pose estimation. However, the IR methods take relatively large running time.

Learning-based approaches have shown credible results for a variety of tasks~\cite{kim2020machine, vecchietti2020sampling, vecchietti2020batch, mishra2021socially}. Thus, to tackle the shortcomings of traditional camera pose estimation methods, lately, end-to-end pose regressors involving deep learning techniques have been investigated. The first absolute pose regressor(APR), PoseNet~\cite{kendall2015posenet}, predicts the camera pose for a given image using an end-to-end model consisting of GoogLeNet~\cite{szegedy2015going} convolutional neural network (CNN) backbone. Following the PoseNet~\cite{kendall2015posenet}, multiple research works~\cite{kendall2016modelling, kendall2017geometric, walch2017image, cai2019hybrid, naseer2017deep, melekhov2017image, brahmbhatt2018geometry, blanton2020extending, wang2020atloc, shavit2021we, shavit2021learning, shavit2022camera} investigating end-to-end methods have been proposed. Given their fixed architecture, these methods have a fixed memory footprint regardless of the size of the scene and their running time is an order of magnitude faster than traditional structure-based methods. However, the accuracy of these methods is still not superior to structure-based methods that make use of 3D information during inference.

Typically, the aforementioned end-to-end methods are trained in a supervised fashion using a dataset consisting of images and ground truth poses for a single scene. Because of that, using the same trained model for images collected in different scenes is non-trivial. Recently, methods that are trained using a dataset with data obtained from multiple scenes have been proposed to alleviate this problem. Multi-Scene PoseNet (MSPN)~\cite{blanton2020extending} and MS-Transformer~\cite{shavit2021learning} train a model with multiple scene data. These absolute pose regressors (APRs) based on multiple scenes aim to learn general embeddings for different scenes when compared to scene-specific APRs. During inference, it is observed that multiple scene APRs increase the robustness when tested with unseen scene images. To improve the performance and robustness of end-to-end camera pose estimation, methods that use as input several images captured sequentially have also been investigated~\cite{brahmbhatt2018geometry, clark2017vidloc}.

Scene-specific APRs are not designed to solve domain invariance by nature, thus they lack the ability to generalize well to different domains of the same scene. Despite being trained with multiple scene data, the MSPN and MS-Transformer also do not encourage domain invariance to different domains. Unlike these approaches, our method aims to address the domain invariance of APRs. In this paper, we hypothesize that using image pairs for a given pose under different conditions as input and using an objective for domain invariance should improve the accuracy of a given APR in unseen domains when compared to the APR that is trained only in the real distribution. To check our hypothesis, we introduce a robust domain adaptive framework for absolute pose regression. In our work, scene images are augmented for different domains by using generative methods to train parallel branches leveraging a contrastive \textsc{Barlow Twins} objective inspired by~\cite{zbontar2021barlow}. This objective has the goal of reducing the difference of embeddings for images taken from the same pose under different domains. \textit{The proposed training framework is general and can be applied to different CNN-based APRs for improving domain invariance.} Additionally, the number of parallel branches used for training can be adjusted for different tasks. The proposed training framework is shown in Fig.~\ref{fig:01_fig-overall_DA}(a) for three parallel branches with shared weights, one for the original image and two for domain augmentations. Domain augmentations are performed by processing the original image using generative adversarial network (GAN) methods such as ManiFest \cite{pizzati2021manifest} and CoMoGAN \cite{pizzati2021comogan}. While the proposed framework ensures the robustness of the model by using parallel branches during training, for inference, as shown in Fig.~\ref{fig:01_fig-overall_DA}(b), the model is loaded as the single branch since parallel branches share weights thus not adding computational complexity when compared to other methods.

For our study, we focus on using an APR with a lightweight CNN-based architecture for embedded applications. The pretrained MobileNetV3-Large~\cite{howard2019searching} backbone is leveraged for learning domain invariant embeddings. These embeddings are then used by a regression head to predict the camera pose, e.g., translation and rotation vectors. For the rotation vector prediction, adding spatial and channel-wise attention to focus on important features related to a specific view of the scene is investigated. The proposed training framework as well as APR architectures are evaluated by using two datasets, the 7Scenes dataset~\cite{glocker2013real}, which contains multiple indoor scenes, and the Cambridge Landmarks dataset~\cite{kendall2015posenet}, which contains multiple outdoor scenes. The results show that our method outperforms CNN-based architectures while achieving comparable performance to transformer-based architectures. It is noted that our method is highly efficient regarding FLOPs, number of parameters, and memory requirements when compared to baseline methods. Ablation studies are performed to investigate the importance of each of the proposed components to the final performance. Additionally, we evaluate our method in three domains unseen during training and show that it achieves superior performance when compared to baseline methods. Throughout this work, we use the term $\mathcal{DA}$-model to refer to the proposed \textbf{D}omain \textbf{A}daptive training framework while $\mathcal{SB}$-model is used to refer to a \textbf{S}ingle \textbf{B}ranch APR training framework. Our main contributions can be outlined as follows.

\input{./figs/01_fig-overall_DA.tex}

\begin{itemize}
    \item We propose a domain adaptive training framework for absolute pose regressors by using a contrastive \textsc{Barlow Twins} objective along with an L2 loss to train parallel branches. The \textsc{Barlow Twins} loss enforces invariance and redundancy reduction on learned embeddings for the same pose images under different domains. The framework is general and can be applied to different absolute pose regressors. 
    \item Shared weights are maintained in parallel branches during training, thus, during inference, our model can be loaded as a single branch model keeping the same runtime complexity when compared to CNN-based absolute pose regressors. 
    \item The addition of spatial and channel-wise attention to the regression head responsible for rotation prediction is investigated. The attention layer incorporated improves rotation prediction in our framework while introducing only about 11K additional parameters($\sim$ 0.1MB) to the network.
    
    \item We propose the use of a lightweight MobileNetV3-Large backbone for the experiments using the proposed domain adaptive framework. It is shown that, even with the use of a lightweight architecture, our method outperforms other CNN-based baseline methods and achieves comparable performance to transformer-based architectures in domains seen during training while outperforming all baseline methods on domains unseen during training. 
    
\end{itemize}

%% file: figs/01_fig-overall_DA.tex
\begin{figure}[t]
    \includegraphics[width=\columnwidth,keepaspectratio]{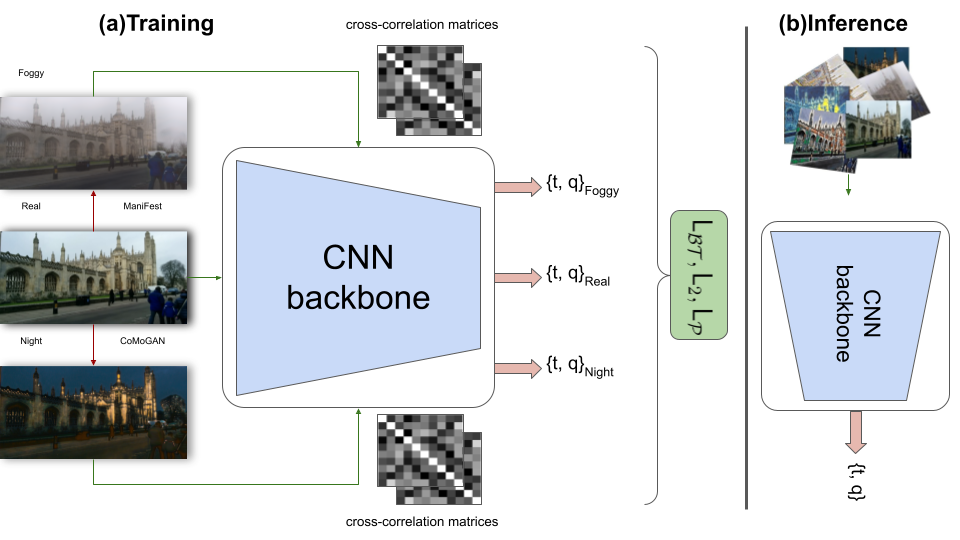}
    \setlength{\abovecaptionskip}{-15pt}
    \caption{Overview of the proposed framework. (a) Domain adaptive training framework. Three parallel branches with shared weights are trained for images related to the same pose under different domains. (b) Inference framework. 
    }
    \label{fig:01_fig-overall_DA}
\end{figure}

%% file: 02_related.tex
\section{Related Work}
\label{sec:related}

In this section, end-to-end deep learning methods for camera pose estimation are reviewed. Additional related work regarding structure-based localization and image retrieval methods is presented in Appendix~H.

\noindent
\textbf{Relative Pose Estimation:} Deep learning-based regression is not only being used in absolute pose regression tasks but also adopted in another body of work, the relative camera pose estimation investigated by RPNet~\cite{en2018rpnet} and others~\cite{melekhov2017relative, laskar2017camera, yang2020rcpnet, rajendran2022relmobnet}. In a relative pose estimation problem, the input given is a pair of images with different poses, and the output is the relative transformation (translation and rotation vectors) between the poses represented by the input images. To tackle this problem, Melekhov et al\cite{melekhov2017relative, laskar2017camera} investigated relative pose estimation with end-to-end deep learning methods. The RPNet \cite{en2018rpnet} is proposed as an end-to-end approach to obtain the full translation vector while previous approaches obtained a translation vector up to a scale factor. Following RPNet, the RCPNet \cite{yang2020rcpnet} used a ResNet-based model with a learnable pose loss objective. The RelMobNet \cite{rajendran2022relmobnet} involved the use of a lightweight CNN backbone and different training strategies to improve computational efficiency for relative pose estimation. Inspired by the methods that predict keypoint locations in a scene \cite{sun2018integral, suwajanakorn2018discovery, kundu20183d, luvizon2019human}, DirectionNet\cite{chen2021wide} proposed the prediction of a discrete distribution over camera poses and demonstrated promising results. Recent research~\cite{Rockwell2022} investigated Vision Transformers (ViT) to estimate the relative pose.

\noindent
\textbf{Absolute Pose Estimation:} In this subsection, we review methods that, similarly to the method proposed in this work, automatically learn features for absolute pose estimation. We define these methods as absolute pose regressors (APRs). The end-to-end APR using a CNN to estimate a 6 Degree of Freedom (DoF) camera pose was PoseNet \cite{kendall2015posenet}. The model is trained on a dataset of a single landmark scene. The regression objective is defined to minimize the deviation for translation and rotation vectors, in which a hyperparameter is used to balance between the translation and rotation losses. Geometric PoseNet \cite{kendall2017geometric} increases the robustness of the PoseNet by learning how to balance the losses automatically as part of the training process. The impact of geometric reprojection error-based objective is also investigated in \cite{kendall2017geometric}.

Following the PoseNet, the end-to-end camera pose estimation is further investigated by combining CNNs and RNNs \cite{walch2017image}. Specifically, Walsh et al \cite{walch2017image} apply long-short term memories on top of the learned CNN features for structured dimensionality reduction to enhance localization performance. Bayesian PoseNet \cite{kendall2016modelling} improves localization accuracy by using a Bayesian CNN to predict the uncertainty of the model during estimation. GPoseNet \cite{cai2019hybrid} uses a Gaussian Process regressor to learn the probability distribution of the 6DoF camera position at different poses. SVS-Pose \cite{naseer2017deep} is a method for data augmentation in 3D space for better pose coverage, which further improves camera pose estimation. MapNet \cite{brahmbhatt2018geometry} combines images with multimodal sensory inputs such as visual odometry and GPS for camera localization. However, it should be noted that the MapNet  relies on the use of multiple sequential images to predict the final pose. IPRNet \cite{shavit2021we} exploits the use of general features obtained by models trained for visual similarity to predict the camera pose. AtLoc \cite{wang2020atloc} exploits an attention mechanism on top of CNN to focus on informative features to predict the camera pose. MSPN~\cite{blanton2020extending} suggests that a single model can be used for multiple scenes instead of scene-specific models. MS-Transformer\cite{shavit2021learning} presents a transformer-based architecture for multiple scenes, providing state-of-the-art results. The prediction of translation vectors in MS-Tranformer is further improved by adding a camera pose encoder and optimization steps as presented in \cite{shavit2022camera}.

\input{./figs/02_fig-architecture.tex}

%% file: figs/02_fig-architecture.tex
\begin{figure*}[t]
 \setlength{\textfloatsep}{0pt }
 \setlength{\abovecaptionskip}{0pt} 
 \setlength{\belowcaptionskip}{0pt} 
\centering
    \includegraphics[width=0.8\textwidth,keepaspectratio]{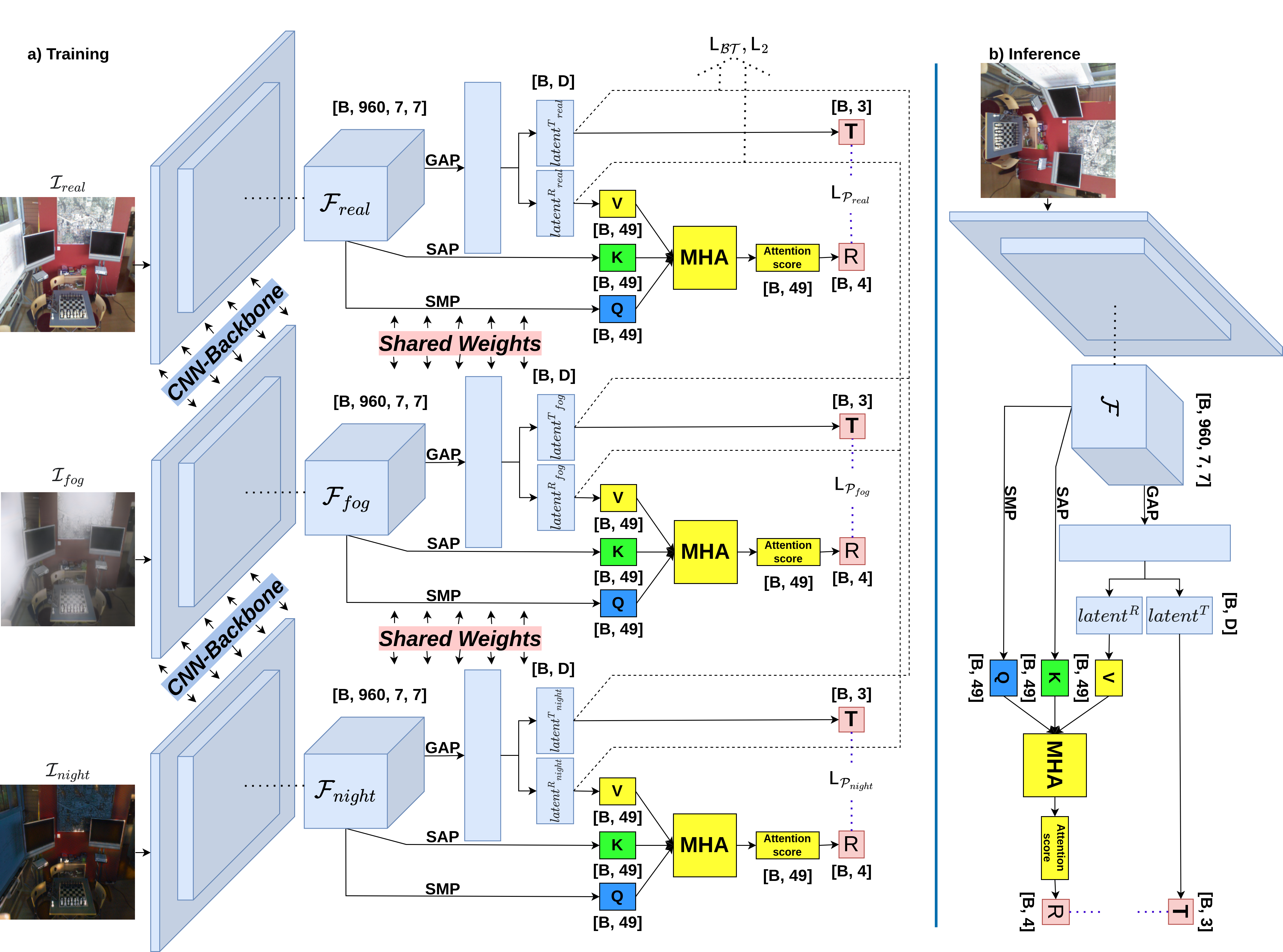}
    \caption{\textbf{Left:} $\mathcal{DA}$-model learns domain invariant features by using a training objective composed of a $\mathsf{L}_{2}$ loss and a \textsc{Barlow Twins} $\mathsf{L}_{\mathcal{BT}}$ loss. Additionally, a pose loss $\mathsf{L}_{\mathcal{P}}$ is applied to optimize pose predictions. \textbf{Right:} The inference stage is performed as a single branch model as three parallel branches share network weights during training.}\label{fig:02_fig-architecture}
\end{figure*}

%% file: 03_method.tex
\section{Proposed Method}
\label{sec:method}

The proposed method is based on two objectives: (i) creation of a domain adaptive framework that can leverage the use of images corresponding to the same pose under different domains during training; (ii) use of the proposed framework with a lightweight APR architecture while keeping competitive performance with baseline methods. For (i), the use of \textsc{Barlow Twins} objective~\cite{zbontar2021barlow} is proposed to reduce the difference of embeddings under different domains. For (ii), the use of a MobileNetV3-Large~\cite{howard2019searching} backbone and the addition of a multi-head attention(MHA)~\cite{vaswani2017attention} module are exploited to improve rotation prediction.

\noindent
\textbf{Domain Adaptive (\texorpdfstring{$\mathcal{DA}$}-) APR Framework:} The proposed framework is general and is to be used with APR architectures. Unlike previous APR approaches, the framework differs by incorporating a domain adaptive objective into the model training. This objective requires the availability of multiple images of the same pose under different domains (light, texture). In this work, the number $N$ of multiple images obtained for each pose is set to 3. To obtain these images, we use two methods based on GANs: ManiFest \cite{pizzati2021manifest} and CoMoGAN~\cite{pizzati2021comogan}. The ManiFest~\cite{pizzati2021manifest} is a few-shot learning-based image translation GAN that learns a context-aware representation of a target domain, and, in this work, is used to generate augmented foggy-view images. The output image after using the ManiFest is defined as $\mathcal{I}_{fog}$ as shown in Eq.~(1). The CoMoGAN~\cite{pizzati2021comogan} is a continuous image translation method used for time-lapse generation, and, in this work, is used to generate augmented night-view images. The output image after using the CoMoGAN is defined as $\mathcal{I}_{night}$ as shown in Eq.~(2).

{\small
\begin{gather}
    \mathcal{I}_{fog} = ManiFest(\mathcal{I}_{real}), \label{eq:manifest}\\
    \mathcal{I}_{night} = CoMoGAN(\mathcal{I}_{real}), \label{eq:comogan}
\end{gather}
}
where $\mathcal{I}_{real}$ is the real image sample contained in the dataset for a given pose $\mathcal{P}_{GT} \thinspace \in \thinspace \mathbb{R}^{7}$. The pose $\mathcal{P}_{GT}$ is given as $\mathcal{P}_{GT} = [t,q]$, where $t \thinspace \in \thinspace \mathbb{R}^{3}$ is the 3-Dimensional translation vector and $q \thinspace \in \thinspace \mathbb{R}^{4}$ is a 4-Dimensional quarternion representing the rotation vector. 
After this preprocessing step, we have $N=3$ images for the same pose $\mathcal{P}_{GT}$ under different domains (real, foggy, night). An APR receives an image as input and provides the predicted pose $\widehat{\mathcal{P}}$ as output for the target scene. During training, we have $N=3$ parallel branches that provide as outputs $\widehat{\mathcal{P}_{real}}$, $\widehat{\mathcal{P}_{fog}}$, and $\widehat{\mathcal{P}_{night}}$ as given in Eqs.~(3), (4), and (5) as following

{\small 
\begin{gather} 
\widehat{\mathcal{P}_{real}} = APR(\mathcal{I}_{real}), \label{eq:pred-pose-real} \\
\widehat{\mathcal{P}_{fog}} = APR(\mathcal{I}_{fog}), \label{eq:pred-pose-fog} \\
\widehat{\mathcal{P}_{night}} = APR(\mathcal{I}_{night}). \label{eq:pred-pose-night}
\end{gather}
}

The parallel branches share the same architecture and network weights. The detailed visualization of the APR architecture utilized in this work for the prediction of $\widehat{\mathcal{P}_{real}}$, $\widehat{\mathcal{P}_{fog}}$, and $\widehat{\mathcal{P}_{night}}$ is shown in Fig.~\ref{fig:02_fig-architecture}. The three parallel branches using MobileNetV3-Large CNN backbones are presented. Details regarding MobileNetV3-Large architecture is given in Appendix~A. The MobileNet-V3-Large CNN backbones encode the three input images into three latent feature maps ($\mathcal{F}_{real}$, $\mathcal{F}_{fog}$, $\mathcal{F}_{night}$) $\in \thinspace \mathbb{R}^{960 \times 7 \times 7}$. These feature maps are used to obtain two latent vectors, one for translation and one for rotation, for each input image, via Global Average Pooling(GAP) and multi-layer perceptron(MLP). These latent vectors $\in \thinspace \mathbb{R}^{D}$ are defined as ${latent^{ T}}_{real}$,  ${latent^{ T}}_{fog}$,  ${latent^{ T}}_{night}$,  ${latent^{ R}}_{real}$,  ${latent^{ R}}_{fog}$, and ${latent^{ R}}_{night}$, where ${latent^{T}}$ is the latent vector used to predict the translation vector and ${latent^{R}}$ is the latent vector used to predict the rotation vector. It is important to note that, given a variable number of domains for the same pose, this methodology can be expanded for training with additional parallel branches or for creating contrastive image pairs for training.

For the translation vector, the latent vector ${latent^{T}}$ of each subgroup (real, foggy, night) is the input for a MLP that finally predicts the translation vector $t$. For the rotation vector, the latent vector ${latent^{R}}$ of each subgroup is combined with the features obtained by the MobileNet-V3-Large CNN backbone to be passed in a multi-head attention layer as follows. The latent vector ${latent^{R}}$ is passed to an MLP layer to get the Value $V \in \thinspace \mathbb{R}^{49}$ of the MHA layer. Query $Q \in \thinspace \mathbb{R}^{49}$ and Key $K \in \thinspace \mathbb{R}^{49}$ vectors are obtained by applying spatial-point-wise max pooling (SMP) and  spatial-point-wise average pooling (SAP) to the feature maps ($\mathcal{F}_{real}$, $\mathcal{F}_{fog}$, $\mathcal{F}_{night}$), respectively. The intuition behind this step is related to restoring features that are important to rotation prediction in a given scene. It is important to mention that the addition of the MHA layer using this formulation introduces approximately 11k new parameters to the APR architecture. The attention score from the MHA layer is then used to predict the rotation vector represented as the quaternion $q$.

\noindent
\textbf{Training Objective:} As shown in Fig.~\ref{fig:02_fig-architecture}, ${latent^{ T}}_{real}$ and ${latent^{ R}}_{real}$ are the latent embeddings of $\mathcal{I}_{real}$ image. The
${latent^{ T}}_{fog}$ and ${latent^{ R}}_{fog}$ are the latent embeddings of $\mathcal{I}_{fog}$ image,
and, ${latent^{ T}}_{night}$ and ${latent^{ R}}_{night}$ are the latent embeddings of $\mathcal{I}_{night}$ image. For improved domain invariance, our objective is to increase similarity among ${latent^{ T}}_{real}$, ${latent^{ T}}_{fog}$, and ${latent^{ T}}_{night}$, and to increase similarity among ${latent^{R}}_{real}$, ${latent^{R}}_{fog}$, and ${latent^{R}}_{night}$. This is achieved by the use of \textsc{Barlow Twins} objective ($\mathsf{L}_{\mathcal{BT}}$). It attempts to make the cross-correlation matrix($\mathcal{C} \in \thinspace \mathbb{R}^{D \times D}$) computed from twin embeddings in the same procedure of \cite{zbontar2021barlow}, as close to the identity matrix as possible. More details about calculating the cross-correlation matrix can be found in Appendix~C. The $\mathsf{L}_{\mathcal{BT}}$ enforces reduction of redundancy within an embedding and invariance between two embeddings by estimating the deviation of the cross-correlation matrix($\mathcal{C}$) from the identity matrix. This deviation is defined as a loss and is backpropagated to update the network weights. The general \textsc{Barlow Twins} objective ($\mathsf{L}_{\mathcal{BT}}$) used in this work is given as
{\small	
\begin{gather} 
    \mathsf{L}_{\mathcal{BT}} \triangleq \underbrace{\alpha_{1} \sum\limits_{i}^{D} {{(1-\mathcal{C}_{ii})}^{2}}}_\text{invariance term} + \underbrace{\alpha_{2}\lambda \sum\limits_{i}^{D} \sum\limits_{j \neq i}^{D} {\mathcal{C}_{ij}}^{2}}_\text{redundancy reduction term}. \label{eq:BT-loss-general}
\end{gather}
}
where $\alpha_{1}$, $\alpha_{2}$ are our empirical weighting parameters while $\lambda$ is a standard weight parameter suggested by~\cite{zbontar2021barlow}. The $\mathcal{C}_{ii}$ represents diagonal elements while $\mathcal{C}_{ij}$ represents off-diagonal elements of the matrix $\mathcal{C}$. Indexing the matrix is done by using $i$, $j$ along dimension $\mathbb{R}^{D \times D}$. From~(\ref{eq:BT-loss-general}) it is clear that diagonal elements of the cross-correlation matrix are forced to be closer to 1 while the off-diagonal elements of the cross-correlation matrix are forced to be 0. To impose domain invariance on all latent embeddings, four cross-correlation matrices $\mathcal{C} \in \thinspace \mathbb{R}^{D \times D}$, e.g., two for translation latent pairs and two for rotation latent pairs, are calculated and obtained as follows
{\small
\begin{gather}
\mathcal{C}^{T^{fog}} = \mathcal{C}({latent^{ T}}_{real}, {latent^{ T}}_{fog}), \label{eq:cross-correl-1} \\
\mathcal{C}^{T^{night}} = \mathcal{C}({latent^{ T}}_{real}, {latent^{ T}}_{night}), \label{eq:cross-correl-2}\\
\mathcal{C}^{R^{fog}} = \mathcal{C}({latent^{ R}}_{real}, {latent^{ R}}_{fog}), \label{eq:cross-correl-3}\\
\mathcal{C}^{R^{night}} = \mathcal{C}({latent^{ R}}_{real}, {latent^{ R}}_{night}). \label{eq:cross-correl-4}
\end{gather}
}
The \textsc{Barlow Twins} objective \cite{zbontar2021barlow} in~(\ref{eq:BT-loss-general}) is applied for the four cross-correlation matrices in (\ref{eq:cross-correl-1}, \ref{eq:cross-correl-2}, \ref{eq:cross-correl-3}, \ref{eq:cross-correl-4}) and combined as {$\mathsf{L}_{\mathcal{BT}_{total}}$}. To further enforce domain invariance, $\mathsf{L}_{2}$ losses for the combination of the real domain and augmented domains are calculated. The $\mathsf{L}_{2}$ losses are defined as $\mathsf{L}_{2_{T_{fog}}}$, $\mathsf{L}_{2_{T_{night}}}$, $\mathsf{L}_{2_{R_{fog}}}$, and $\mathsf{L}_{2_{R_{night}}}$ and put together as $\mathsf{L}_{2_{total}}$ and an objective to minimize the error during the prediction of the final pose is evaluated. This loss, given as $\mathsf{L}_{\mathcal{P}}$ \cite{kendall2017geometric}, is calculated for the prediction of $\widehat{\mathcal{P}_{real}}$, $\widehat{\mathcal{P}_{fog}}$, and $\widehat{\mathcal{P}_{night}} \thinspace \in \thinspace \mathbb{R}^{7}$ with respect to the ground truth pose $\mathcal{P}_{GT}$ and combined as $\mathsf{L}_{\mathcal{P}_{total}}$. The final training objective $\mathsf{L}_{\mathcal{DA}}$ for network optimization is the combination of the three objectives {$\mathsf{L}_{\mathcal{BT}_{total}}$}, {$\mathsf{L}_{2_{total}}$}, and {$\mathsf{L}_{\mathcal{P}_{total}}$} as following

{\small
\begin{gather}
\resizebox{.8 \columnwidth}{!}{$
    \mathsf{L}_{\mathcal{BT}_{total}} = \mathsf{L}_{\mathcal{BT}_{T_{fog}}} + \mathsf{L}_{\mathcal{BT}_{T_{night}}} + \mathsf{L}_{\mathcal{BT}_{R_{fog}}} + \mathsf{L}_{\mathcal{BT}_{R_{night}}}$, \label{eq:L-BT-total}
}\\
\resizebox{.8 \columnwidth}{!}{$
    \mathsf{L}_{2_{total}} = \mathsf{L}_{2_{T_{fog}}} + \mathsf{L}_{2_{T_{night}}} + \mathsf{L}_{2_{R_{fog}}} + \mathsf{L}_{2_{R_{night}}}$, \label{eq:L-2-total}
}\\
    \mathsf{L}_{\mathcal{P}_{total}} = \mathsf{L}_{\mathcal{P}_{real}} + \mathsf{L}_{\mathcal{P}_{fog}} + \mathsf{L}_{\mathcal{P}_{night}}, \label{eq:L-p-total} \\
    \mathsf{L}_{\mathcal{DA}} = \mathsf{L}_{\mathcal{BT}_{total}} + \mathsf{L}_{2_{total}} + \mathsf{L}_{\mathcal{P}_{total}}. \label{eq:L-DA-total}
\end{gather}
}

The network is trained end-to-end with the loss function $\mathsf{L}_{\mathcal{DA}}$ given in~(\ref{eq:L-DA-total}).

\noindent
\textbf{Inference:} An important aspect of the proposed framework for absolute pose estimation is that the architecture of each parallel branch uses shared weights during training. Since the weights are shared, the trained model can be used as a single-branch model during inference time. Additionally, it can be used for inference for an image with any data distribution. In Fig.~\ref{fig:02_fig-architecture}(b), the inference process for pose estimation of a single image is shown. This property of the proposed framework ensures that no additional complexity during inference is added and we achieve faster runtime as will be presented in the next section.

\input{./tables/01_table-Cambridge.tex}
\input{./tables/02_table-7Scenes.tex}

\input{./figs/03_fig-trajectory.tex}

%% file: tables/01_table-Cambridge.tex
\begin{table*}[hbt!]
\centering
\caption{Results for outdoors Cambridge Landmark dataset. The last column represents the overall rank of different methods.}
\label{tab:01_table-Cambridge}
\resizebox{0.96\textwidth}{!}{%
\centering
\setlength{\extrarowheight}{2pt}
\begin{tabular}{|l|cc|cc|cc|cc|cc|cc|c|}
\hline
\textbf{Method}                    & \multicolumn{2}{|c|}{\textbf{KingsCollege}} & \multicolumn{2}{c|}{\textbf{OldHospital}} & \multicolumn{2}{c|}{\textbf{ShopFacade}} & \multicolumn{2}{c|}{\textbf{StMarysChurch}} & \multicolumn{2}{c|}{\textbf{Average}} & \multicolumn{2}{c|}{\textbf{Rank}}        & \textbf{Overall Rank} \\
\multicolumn{1}{|l|}{}               & \textbf{T(m)}      & \textbf{R(deg)}      & \textbf{T(m)}      & \textbf{R(deg)}     & \textbf{T(m)}     & \textbf{R(deg)}     & \textbf{T(m)}       & \textbf{R(deg)}      & \textbf{T(m)}    & \textbf{R(deg)}   & \textbf{T} & \textbf{R} & \multicolumn{1}{|l|}{}  \\
\hline
\textbf{MSPN~\cite{blanton2020extending}}                      & 1.73               & 3.65                 & 2.55               & 4.05                & 2.92              & 7.49                & 2.67                & 6.18                 & 2.47             & 5.34              & 13                   & 10                & 12                    \\
\textbf{IR-Baseline~\cite{sattler2019understanding}}               & 1.48               & 4.45                 & 2.68               & 4.63                & 0.90              & 4.32                & 1.62                & 6.06                 & 1.67             & 4.87              & 9                    & 8                 & 9                     \\
\textbf{PoseNet~\cite{kendall2015posenet}}                   & 1.92               & 5.40                 & 2.31               & 5.38                & 1.46              & 8.08                & 2.65                & 8.48                 & 2.09             & 6.84              & 12                   & 13                & 13                    \\
\textbf{PoseNet(Learnable)~\cite{kendall2017geometric}}        & 0.99               & 1.06                 & 2.17               & 2.94                & 1.05              & 3.97                & 1.49                & 3.43                 & 1.43             & 2.85              & 6                    & 2                 & 3                     \\
\textbf{GeoPoseNet~\cite{kendall2017geometric}}                & 0.88               & 1.04                 & 3.20               & 3.29                & 0.88              & 3.78                & 1.57                & 3.32                 & 1.63             & 2.86              & 8                    & 3                 & 5                     \\
\textbf{LSTM-PN~\cite{walch2017image}}                   & 0.99               & 3.65                 & 1.51               & 4.29                & 1.18              & 7.44                & 1.52                & 6.68                 & 1.30             & 5.52              & 3                    & 11                & 8                     \\
\textbf{GPoseNet~\cite{cai2019hybrid}}                  & 1.61               & 2.29                 & 2.62               & 3.89                & 1.14              & 5.73                & 2.93                & 6.46                 & 2.08             & 4.59              & 11                   & 7                 & 10                    \\
\textbf{BayesianPN~\cite{kendall2016modelling}}                & 1.74               & 4.06                 & 2.57               & 5.14                & 1.25              & 7.54                & 2.11                & 8.38                 & 1.92             & 6.28              & 10                   & 12                & 11                    \\
\textbf{SVS-Pose~\cite{naseer2017deep}}                  & 1.06               & 2.81                 & 1.50               & 4.03                & 0.63              & 5.73                & 2.11                & 8.11                 & 1.33             & 5.17              & 4                    & 9                 & 6                     \\
\textbf{MapNet~\cite{brahmbhatt2018geometry}}                    & 1.07               & 1.89                 & 1.94               & 3.91                & 1.49              & 4.22                & 2.00                & 4.53                 & 1.63             & 3.64              & 7                    & 6                 & 6                     \\
\textbf{IRPNet~\cite{shavit2021we}}                    & 1.18               & 2.19                 & 1.87               & 3.38                & 0.72              & 3.47                & 1.87                & 4.94                 & 1.42             & 3.45              & 5                    & 5                 & 4                     \\
\textbf{$\mathcal{DA}$-model (Ours)}        & 0.74               & 2.81                 & 1.88               & 2.97                & 0.84              & 3.71                & 1.31                & 4.29                 & 1.19             & 3.44              & 2                    & 4                 & 2                     \\
\textbf{MS-Transformer(Optimized)~\cite{shavit2021learning,shavit2022camera}} & *(0.83)            & 1.47                 & *(1.81)            & 2.39                & *(0.86)           & 3.07                & *(1.62)             & 3.99                 & 0.96(1.28)       & 2.73              & 1                    & 1                 & 1  \\
\hline
\end{tabular}%
}
{\footnotesize {\raggedright *Scene-specific optimized translation vectors are unknown; however, for ranking, we used the overall average result from~\cite{shavit2022camera} \par}}
\end{table*}

%% file: tables/02_table-7Scenes.tex

\begin{table*}[hbt!] 
\centering
\caption{Results for indoor 7Scenes dataset. The last column represents the overall rank of different methods.}
\label{tab:02_table-7Scenes}
\resizebox{0.96\textwidth}{!}{%
\centering
\setlength{\extrarowheight}{3pt}
\begin{tabular}{|l|cc|cc|cc|cc|cc|cc|cc|cc|cc|c|}
\hline
\textbf{Method}   & \multicolumn{2}{c|}{\textbf{Chess}} & \multicolumn{2}{c|}{\textbf{Fire}} & \multicolumn{2}{c|}{\textbf{Heads}} & \multicolumn{2}{c|}{\textbf{Office}} & \multicolumn{2}{c|}{\textbf{Pumpkin}} & \multicolumn{2}{c|}{\textbf{Kitchen}} & \multicolumn{2}{c|}{\textbf{Stairs}} & \multicolumn{2}{c|}{\textbf{Average}} & \multicolumn{2}{c|}{\textbf{Rank}}        & \textbf{Overall Rank} \\
                                   & \textbf{T(m)}   & \textbf{R(deg)}  & \textbf{T(m)}  & \textbf{R(deg)}  & \textbf{T(m)}   & \textbf{R(deg)}  & \textbf{T(m)}   & \textbf{R(deg)}   & \textbf{T(m)}    & \textbf{R(deg)}   & \textbf{T(m)}    & \textbf{R(deg)}   & \textbf{T(m)}   & \textbf{R(deg)}   & \textbf{T(m)}    & \textbf{R(deg)}   & \textbf{T(m)} & \textbf{R(deg)} & \multicolumn{1}{l|}{}  \\
\hline
\textbf{MSPN~\cite{blanton2020extending}}                      & 0.09            & 4.76             & 0.29           & 10.50            & 0.16            & 13.10            & 0.16            & 6.80              & 0.19             & 5.50              & 0.21             & 6.61              & 0.31            & 11.63             & 0.20             & 8.41              & 4                    & 6                 & 4                     \\
\textbf{IR-Baseline~\cite{sattler2019understanding}}               & 0.18            & 10.00            & 0.33           & 12.40            & 0.15            & 14.30            & 0.25            & 10.10             & 0.26             & 9.42              & 0.27             & 11.10             & 0.24            & 14.70             & 0.24             & 11.72             & 10                   & 14                & 12                    \\
\textbf{PoseNet~\cite{kendall2015posenet}}                   & 0.32            & 8.12             & 0.47           & 14.40            & 0.29            & 12.00            & 0.48            & 7.68              & 0.47             & 8.42              & 0.59             & 8.64              & 0.47            & 13.80             & 0.44             & 10.43             & 13                   & 13                & 14                    \\
\textbf{PoseNet(Learnable)~\cite{kendall2017geometric}}        & 0.14            & 4.50             & 0.27           & 11.80            & 0.18            & 12.10            & 0.20            & 5.77              & 0.25             & 4.82              & 0.24             & 5.52              & 0.37            & 10.60             & 0.24             & 7.87              & 9                    & 4                 & 7                     \\
\textbf{GeoPoseNet~\cite{kendall2017geometric}}                & 0.13            & 4.48             & 0.27           & 11.30            & 0.17            & 13.00            & 0.19            & 5.55              & 0.26             & 4.75              & 0.23             & 5.35              & 0.35            & 12.40             & 0.23             & 8.12              & 6                    & 5                 & 6                     \\
\textbf{LSTM-PN~\cite{walch2017image}}                   & 0.24            & 5.77             & 0.34           & 11.90            & 0.21            & 13.70            & 0.30            & 8.08              & 0.33             & 7.00              & 0.37             & 8.83              & 0.40            & 13.70             & 0.31             & 9.86              & 12                   & 11                & 10                    \\
\textbf{GPoseNet~\cite{cai2019hybrid}}                  & 0.20            & 7.11             & 0.38           & 12.30            & 0.21            & 13.80            & 0.28            & 8.83              & 0.37             & 6.94              & 0.35             & 8.15              & 0.37            & 12.50             & 0.31             & 9.95              & 11                   & 12                & 10                    \\
\textbf{BayesianPN~\cite{kendall2016modelling}}                & 0.37            & 7.24             & 0.43           & 13.70            & 0.31            & 12.00            & 0.48            & 8.04              & 0.61             & 7.08              & 0.58             & 7.54              & 0.48            & 13.10             & 0.47             & 9.81              & 14                   & 10                & 12                    \\
\textbf{Hourglass~\cite{melekhov2017image}}                 & 0.15            & 6.17             & 0.27           & 10.80            & 0.19            & 11.60            & 0.21            & 8.48              & 0.25             & 7.01              & 0.27             & 10.20             & 0.29            & 12.50             & 0.23             & 9.53              & 7                    & 9                 & 9                     \\
\textbf{MapNet~\cite{brahmbhatt2018geometry}}                    & 0.08            & 3.25             & 0.27           & 11.70            & 0.18            & 13.30            & 0.17            & 5.15              & 0.22             & 4.02              & 0.23             & 4.93              & 0.30            & 12.10             & 0.21             & 7.79              & 5                    & 3                 & 3                     \\
\textbf{AtLoc~\cite{wang2020atloc}}                     & 0.10            & 4.07             & 0.25           & 11.40            & 0.16            & 11.80            & 0.17            & 5.34              & 0.21             & 4.37              & 0.23             & 5.42              & 0.26            & 10.50             & 0.20             & 7.56              & 3                    & 2                 & 2                     \\
\textbf{IPRNet~\cite{shavit2021we}}                    & 0.13            & 5.64             & 0.25           & 9.67             & 0.15            & 13.10            & 0.24            & 6.33              & 0.22             & 5.78              & 0.30             & 7.29              & 0.34            & 11.60             & 0.23             & 8.49              & 7                    & 7                 & 8                     \\
\textbf{$\mathcal{DA}$-model (Ours)} & 0.11            & 6.06             & 0.26           & 11.67            & 0.15            & 12.81            & 0.17            & 8.34              & 0.21             & 6.37              & 0.19             & 8.90              & 0.26            & 12.21             & 0.19             & 9.48              & 2                    & 8                 & 4                     \\
\textbf{MS-Transformer(Optimized)~\cite{shavit2021learning, shavit2022camera}}            & *(0.11)         & 4.66             & *(0.24)        & 9.60             & *(0.14)         & 12.19            & *(0.17)         & 5.66              & *(0.18)          & 4.44              & *(0.17)          & 5.94              & *(0.26)         & 8.45              & 0.15(0.18)       & 7.28              & 1                    & 1                 & 1    \\
\hline
\end{tabular}%
}
{\footnotesize {\raggedright *Scene-specific optimized translation vectors are unknown; however, for ranking, we used the overall average result from~\cite{shavit2022camera} \par}}
\end{table*}

%% file: figs/03_fig-trajectory.tex
\begin{figure*}[ht]
\captionsetup[subfigure]{labelformat=empty}

\centering
\begin{tabular}{ccccc}
{\rotatebox{90}{\qquad\quad$\mathcal{DA}$-model}}&
\subfloat[]{\includegraphics[width = 0.20\textwidth]{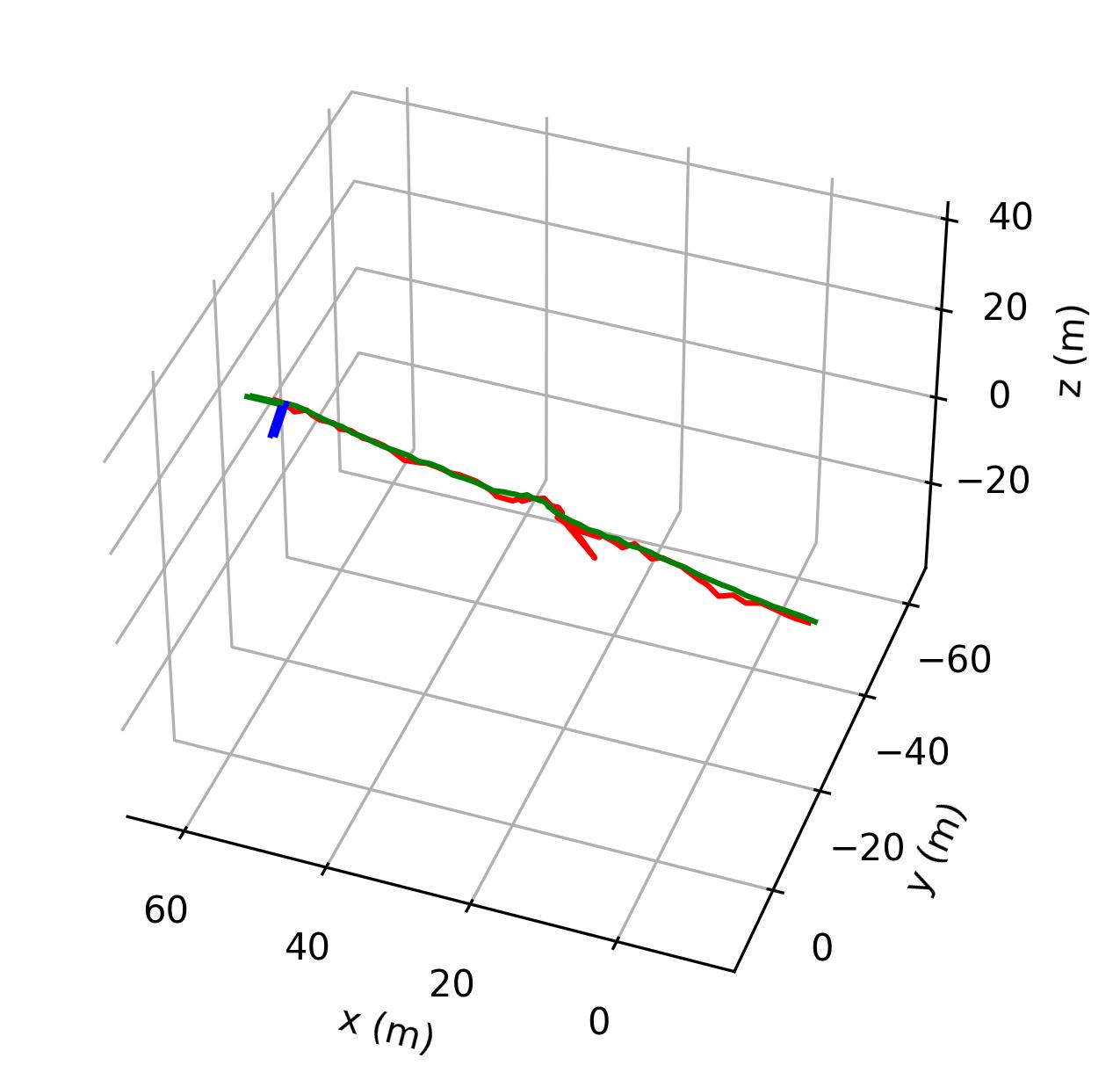}} &
\subfloat[]{\includegraphics[width = 0.20\textwidth]{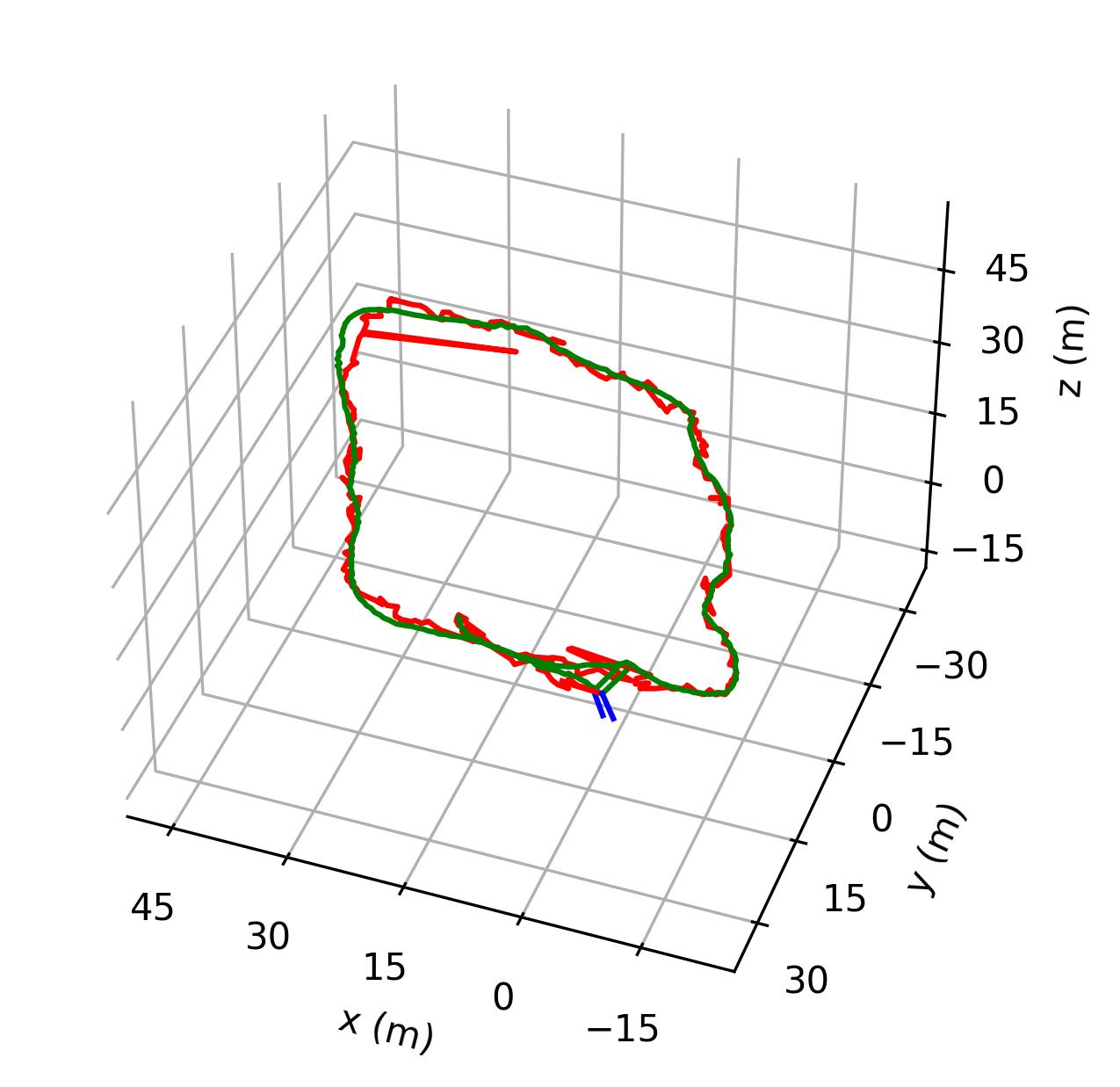}} &
\subfloat[]{\includegraphics[width = 0.20\textwidth]{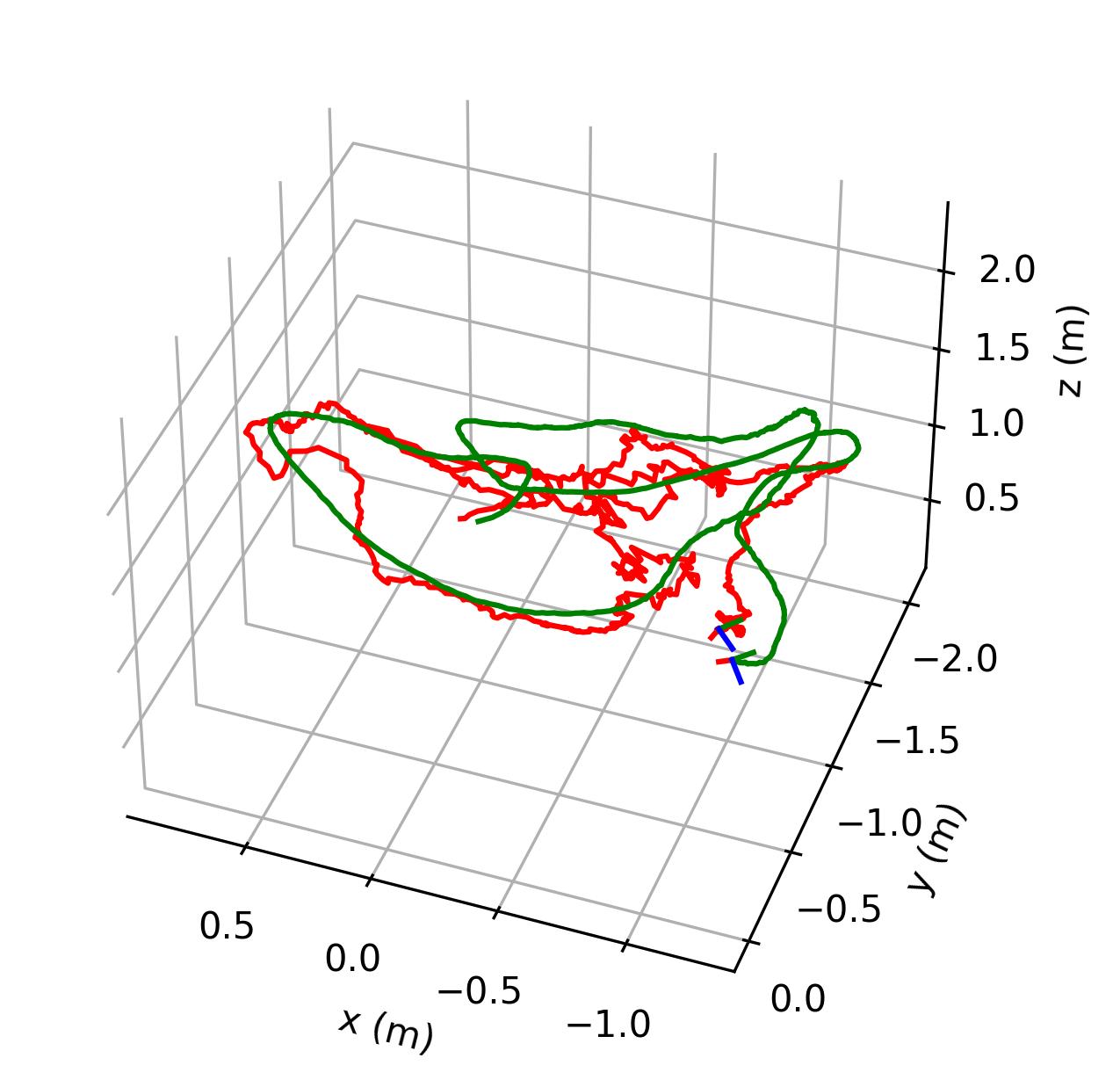}} &
\subfloat[]{\includegraphics[width = 0.20\textwidth]{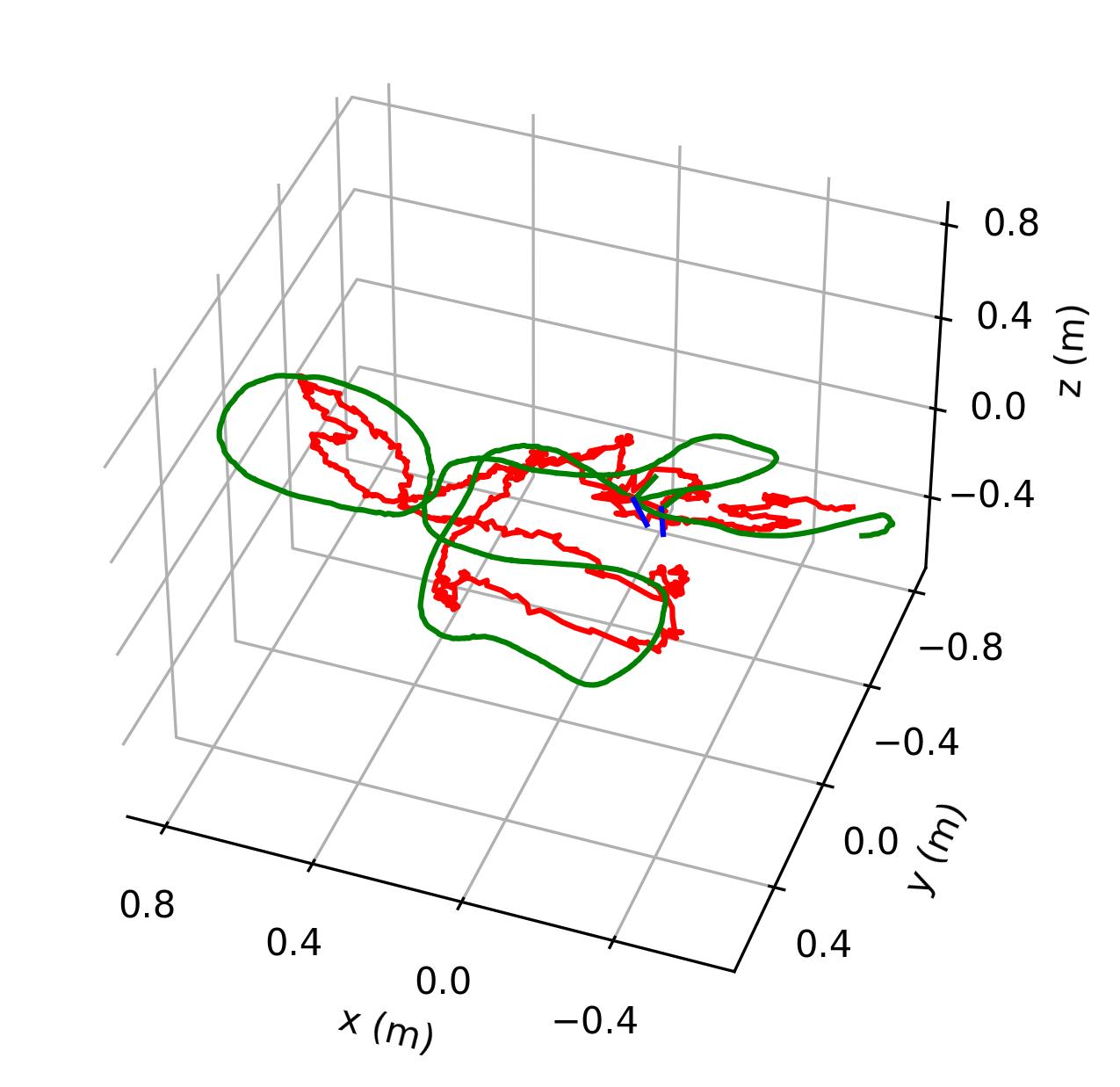}} \\[-3.61ex]
{\rotatebox{90}{\qquad MS-Transformer}}&
\subfloat[]{\includegraphics[width = 0.20\textwidth]{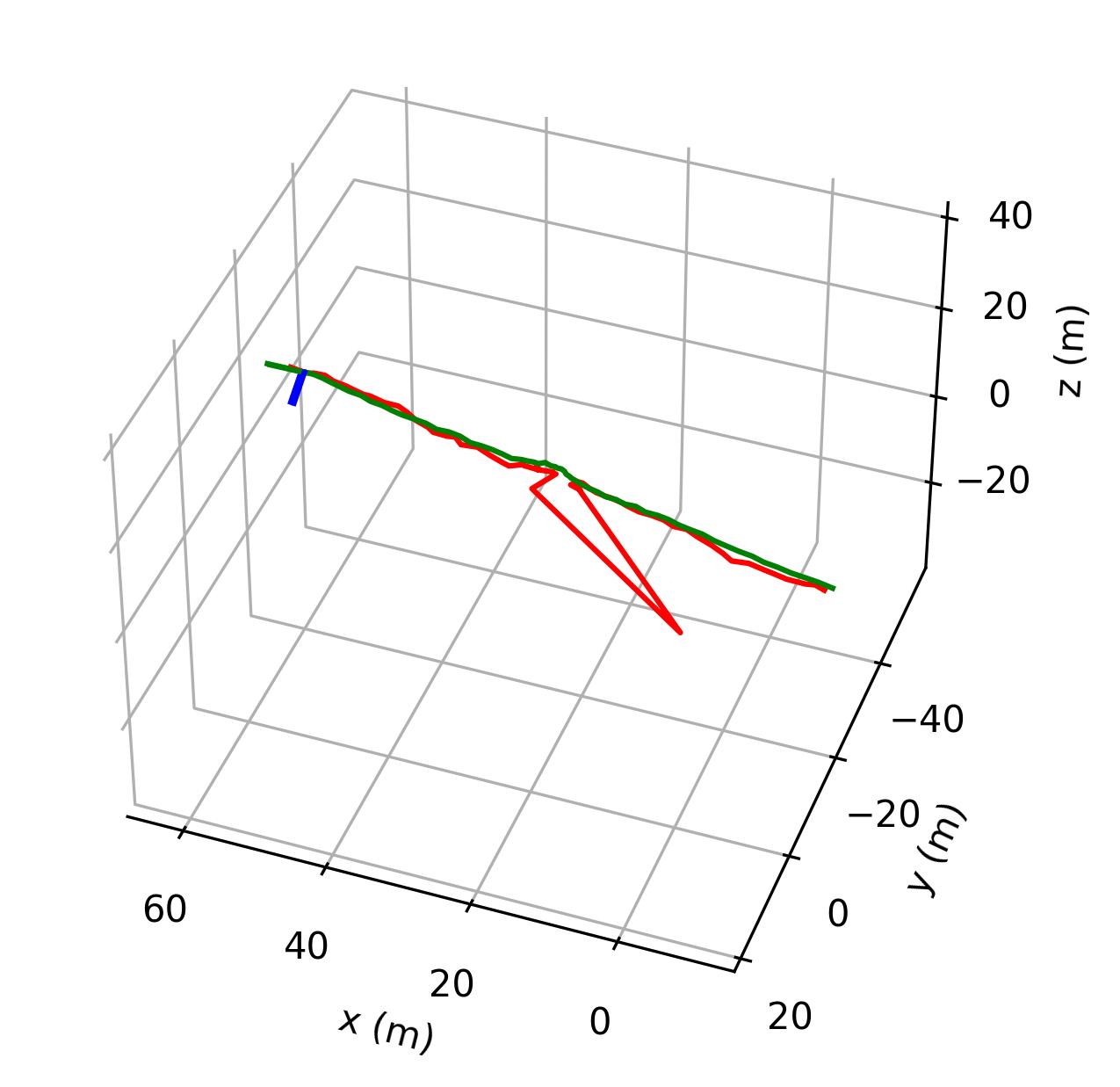}} &
\subfloat[]{\includegraphics[width = 0.20\textwidth]{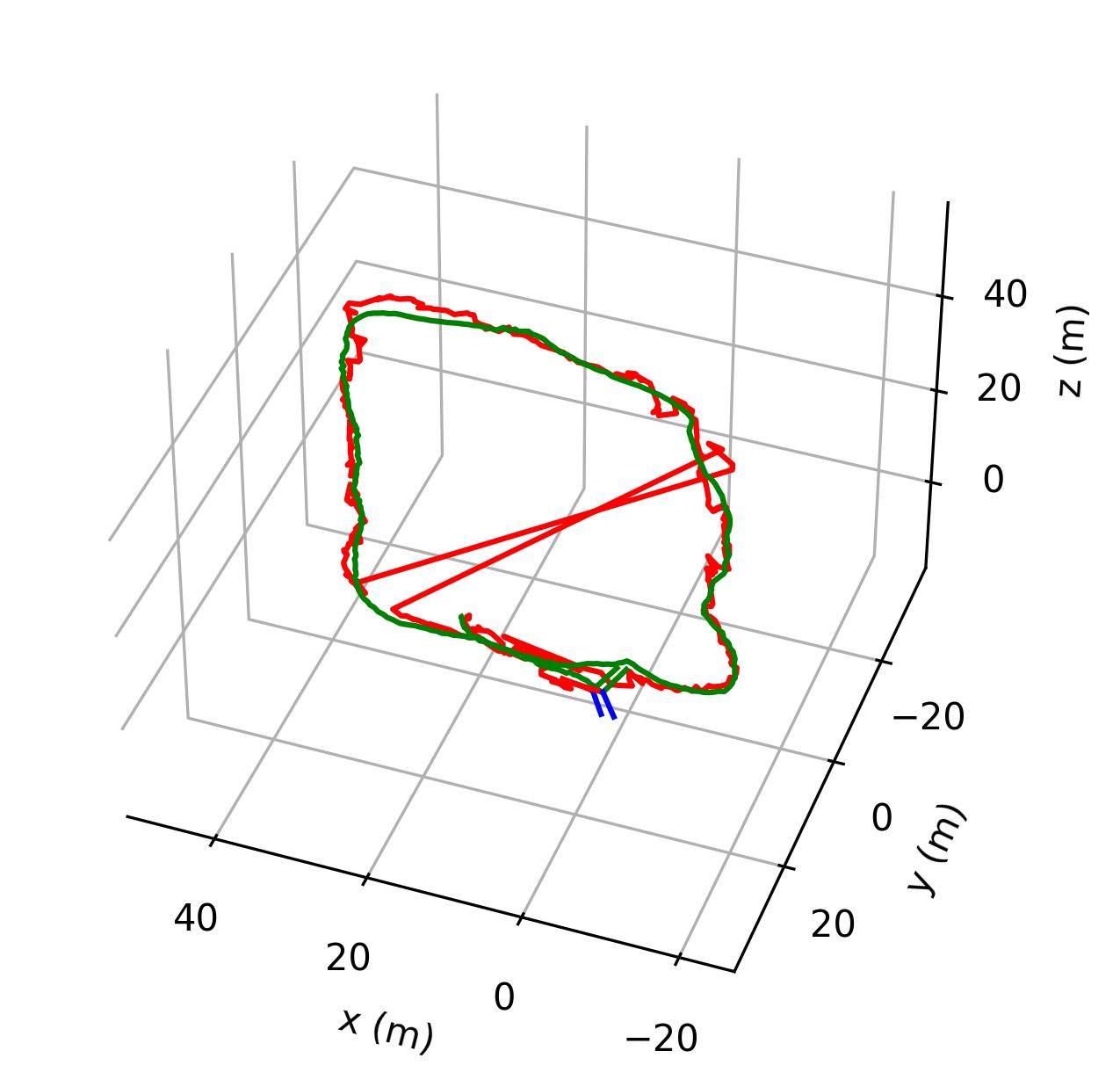}} &
\subfloat[]{\includegraphics[width = 0.20\textwidth]{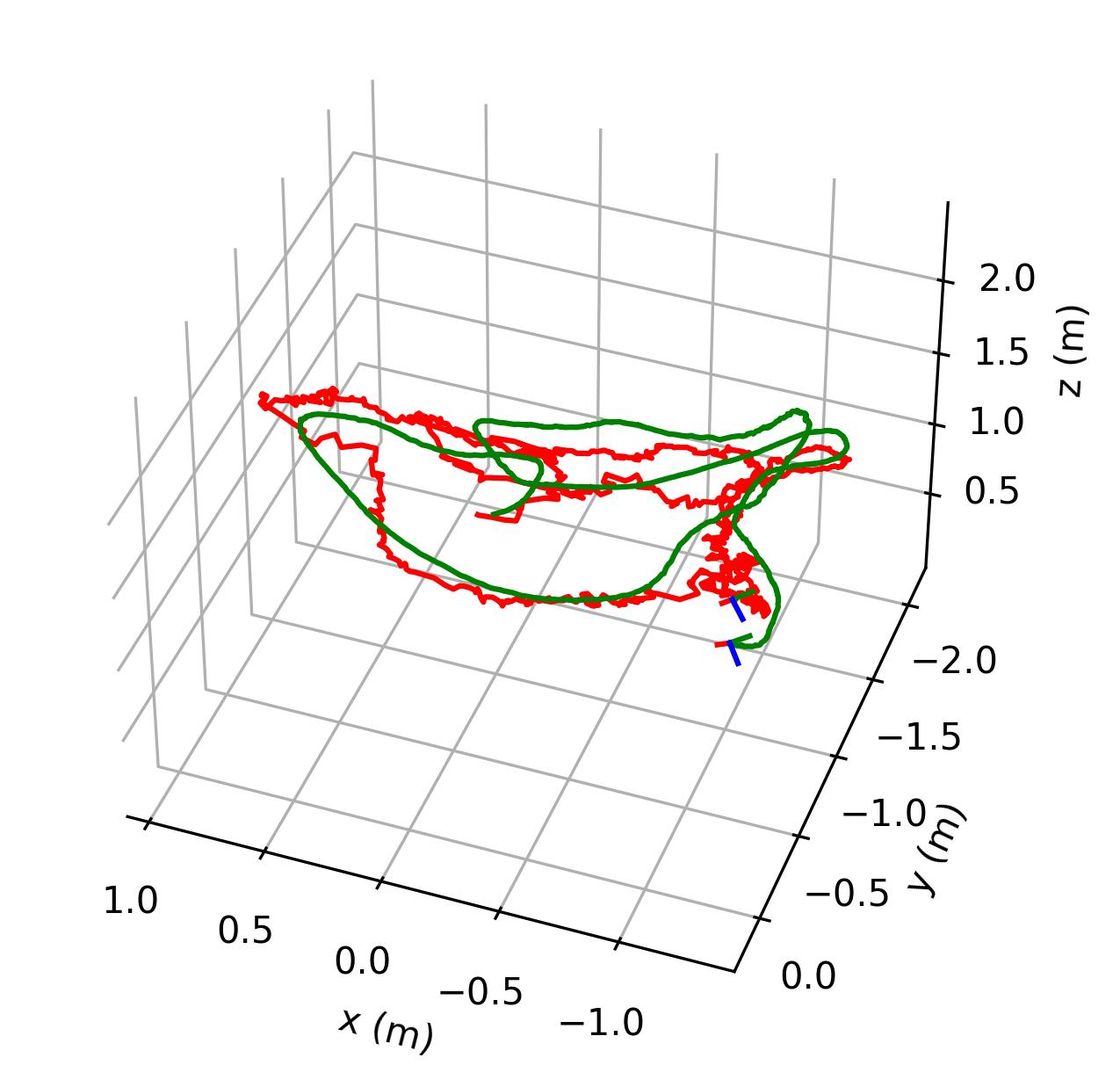}} &
\subfloat[]{\includegraphics[width = 0.20\textwidth]{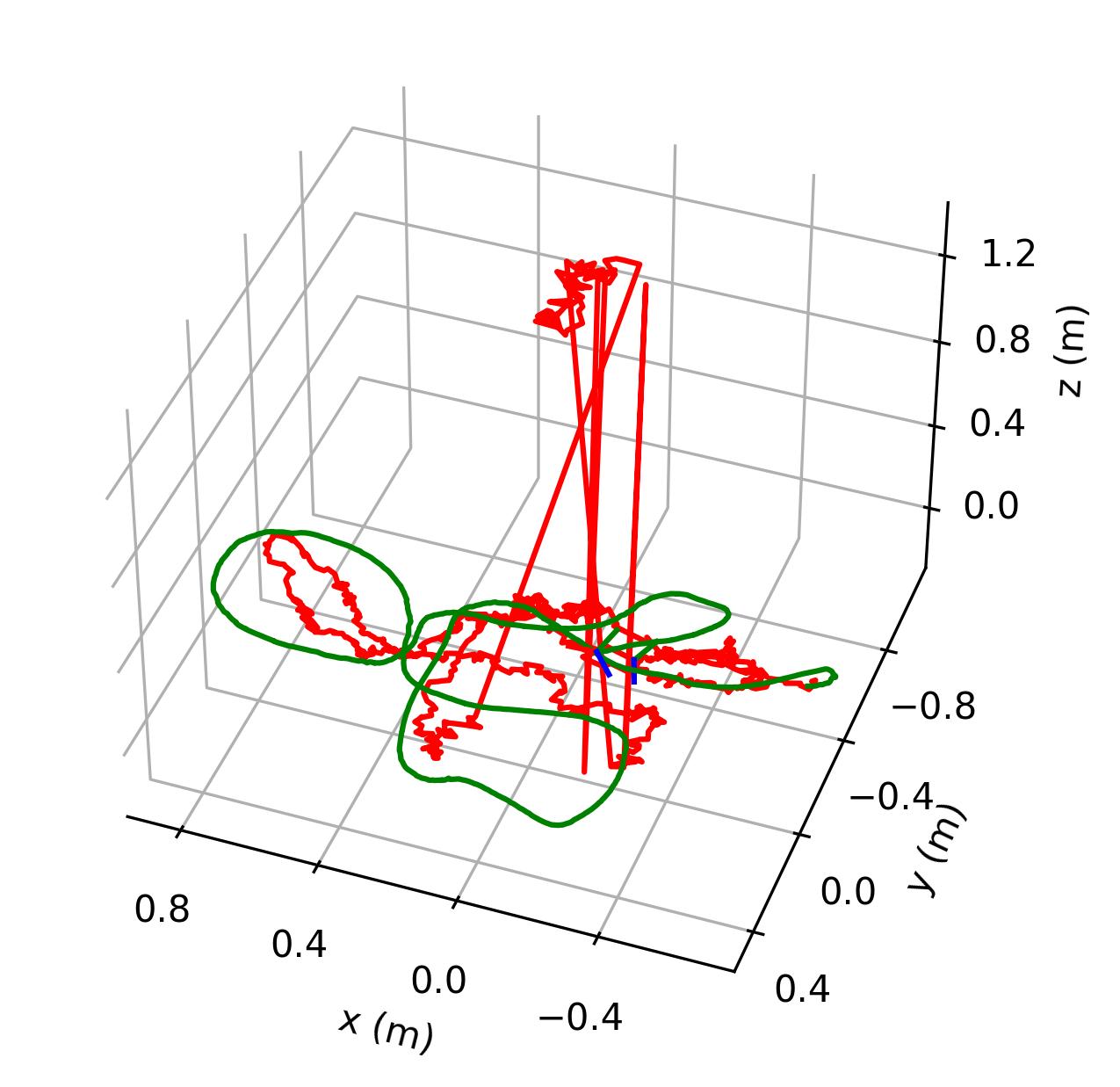}}
\end{tabular}

\caption{Camera trajectory visualization~\cite{fabisch2019pytransform3d} on real Cambridge Landmarks and 7Scenes datasets~\cite{glocker2013real, kendall2015posenet}. Each plot shows the camera trajectory, green for the ground truth and red for the prediction. From left to right, the testing trajectories are for the scenes KingsCollege-seq-02, StMarysChurch-seq-13, Office-seq-06, and Heads-seq-01. From the $1^{st}$ row, it is clear that the $\mathcal{DA}$-model inherently leads to fewer discontinuities and more stable trajectories than the MS-Transformer\cite{shavit2021learning} ($2^{nd}$ row). More trajectories and analysis are given in Appendix~D.}
\vspace{-1.0em}
\label{fig:trajectory}
\end{figure*}

%% file: 04_experiment.tex
\section{Experiments}
\label{sec:experiment}

\noindent
\textbf{Datasets:}  Our approach is evaluated using the Cambridge Landmarks\cite{kendall2015posenet} and 7Scenes\cite{glocker2013real} datasets, which have recently been used as camera pose estimation benchmark sets. The Cambridge Landmarks dataset includes six scenes \((\sim900 - 5500m^{2})\) taken in urban areas. For the comparative analysis, we considered four scenes (KingsCollege, OldHospital, ShopFace, and StMarysChurch) from the Cambridge Landmarks dataset that are used for APR assessments. Captured with a handheld Kinect RGB-D camera, the 7Scenes dataset contains seven small-scale scenes \((\sim1 - 10m^{2})\) depicting indoor environments. We use all seven scenes from the 7Scenes dataset (Chess, Fire, Heads, Office, Pumpkin, Kitchen, and Stairs) for a comprehensive performance assessment. \textcolor{black}{Official train/test splits are used from the datasets.} \textcolor{black}{Samples of dataset images and domain augmentations are shown in Appendix~I.}

\noindent
\textbf{Environment:}  Following the preprocessing steps performed in \cite{shavit2021learning, shavit2022camera}, the scene images contained in the dataset are re-scaled so that the dimension of the shorter side of the image is set to have 256 pixels. The re-scaled image is then cropped randomly into a square shape (224 $\times$ 224) in order to train the APR network. Similar to the baselines, it is evaluated using center crops at test time. The loss function given in~(\ref{eq:L-DA-total}) is used to optimize the network by using the Adam optimizer~\cite{kingma2014adam} with the learning rate set to $1e^{-3}$ and weight decay set to $1e^{-4}$. All $\lambda$ in $\mathsf{L}_{\mathcal{BT}}$ are kept at $5.1e^{-3}$ while $\alpha_{1}$ and $\alpha_{2}$ are set at $1e^{-7}$ and $1e^{-3}$. More details on the environment and hyperparameter settings are given in Appendix~B.

\subsection{Comparison with baseline methods}
\label{subsec:exp-sota}

The visual localization results for the Cambridge Landmarks and 7scenes dataset are presented in Tables~\ref{tab:01_table-Cambridge} and \ref{tab:02_table-7Scenes}, respectively. Median value\textcolor{red}{} of translation error(meters) and rotation error(degrees) are shown for each target scene. The overall average of median errors calculated across scenes is also added for comparison. \textcolor{black}{We rank the proposed method and the baseline methods similarly to~\cite{shavit2021learning}.} Table~\ref{tab:01_table-Cambridge} shows the results for outdoor scenes. In Table~\ref{tab:01_table-Cambridge} it is seen that the proposed method achieves the second-best result for translation prediction while achieving the fourth-best result for rotation prediction. It should be noted that in the last row of Table~\ref{tab:01_table-Cambridge} we report the refined (optimized) pose estimation results from \cite{shavit2022camera}. The refinement in \cite{shavit2022camera} requires additional training for pose encoder and an iterative post-processing step in addition to the inference performed by the transformer network. On the other hand, our network works as a simple APR in terms of inference, without any optimization step (see Section~\ref{subsec:exp-runtime} for runtime analysis). Even with a lightweight architecture, the proposed method outperforms computationally heavy transformer-based architecture without a refinement step \cite{shavit2021learning} in terms of translation vector estimation for outdoor scenes.

Table~\ref{tab:02_table-7Scenes} shows the results for indoor scenes. In Table~\ref{tab:02_table-7Scenes} it is shown that the proposed method achieves the second-best result for translation prediction while achieving the eighth-best result for rotation prediction. Even though the proposed method achieves consistent results for translation prediction, for indoor scenes we observed limitations in the rotation prediction. As it is presented in subsection~\ref{subsec:exp-ablation}, the addition of the MHA component improves rotation prediction by having access to the features learned by the CNN backbone. To evaluate how the error in rotation prediction affects the generation of camera trajectories, we evaluate the camera trajectory for two outdoor scenes and two indoor scenes using the proposed method and MS-Transformer. The visualization of camera trajectories is presented in Fig.~\ref{fig:trajectory}. It can be observed that, even with a higher predicted error and without any temporal information, the proposed method leads to less discontinuities and more stable trajectories.

\subsection{Computational Complexity Analysis}
\label{subsec:exp-runtime}

We perform a computational complexity analysis with the proposed method and three baseline methods, MS-Transformer, AtLoc, and MapNet. Flop count, activation count, number of architecture parameters, and memory footprint are presented, similar to the analysis shown in \cite{radosavovic2020designing}. Flop count and activation count are calculated using the tool~\cite{fvcore}. The results obtained are presented in Table~\ref{tab:03_table-FLOPs}. It is shown that our method is highly efficient regarding FLOPs, number of parameters and memory footprint when compared to baseline methods. The proposed  method uses nearly 24 times fewer FLOPs than MS-Transformer and nearly 13 times fewer than AtLoc. Our model also achieves equivalence to other methods while having 5-7 times fewer parameters and a memory footprint. In terms of activations, our model has approximately 12 times fewer activations when compared to the MS-Transformer while having approximately 16\% more activations when compared to AtLoc and MapNet. More details on various CNN backbones impact on domain adaptive framework is presented in Appendix~G.

\subsection{Ablation studies}
\label{subsec:exp-ablation}

\input{./tables/03_table-FLOPs.tex}
\input{./tables/04_table-Ablation.tex}
\noindent
\textbf{Impact of Proposed Components:} Table~\ref{tab:04_table-Ablation} shows how individual components in our domain adaptive training framework impact the model performance on outdoor scenes: \textit{ShopFacade}, \textit{OldHospital}, and indoor scenes: \textit{Chess} and \textit{Heads}. It is observed that the proposed components $\mathsf{L}_{BT}$, $\mathsf{L}_{2}$, and MHA impact especially rotation estimation in the scenes analysed.

\input{./tables/05_table-FogNight.tex}

\noindent
\textbf{MS-Transformer vs proposed \texorpdfstring{$\mathcal{SB}$-}-model vs proposed \texorpdfstring{$\mathcal{DA}$-}-model on domains seen during training:} Next, we compare the proposed $\mathcal{DA}$-model with MS-Transformer and with a single-branch $\mathcal{SB}$-model trained with the same APR architecture as $\mathcal{DA}$-model and without the domain adaptive framework on the original datasets. The results are shown in Table~\ref{tab:05_table-FogNight}. In this comparison, we evaluate using real domain and domain augmentation corresponding to foggy and night images from the test set. The average median errors calculated on the three domains are reported. It is seen that the $\mathcal{DA}$-model outperforms other methods, which are not domain adaptive, in 16 out of 20 instances. It is important to note that MS-Transformer and $\mathcal{SB}$-model do not have access to foggy and night-view images during the training process.

\noindent
\textbf{MS-Transformer vs proposed \texorpdfstring{$\mathcal{SB}$-}-model vs proposed \texorpdfstring{$\mathcal{DA}$-}-model on domains unseen during training:} Table~\ref{tab:06_table-MosaicUdnieStarry} shows the comparison between MS-Transformer, $\mathcal{SB}$-model, and $\mathcal{DA}$-model in three domains unseen during training. The three domains, mosaic, udnie, and starry are defined as
\label{subsubsec:exp-abl-Unseen}
\begin{gather}
    \mathcal{I}_{mosaic} = \mathbb{G}_{mosaic}(\mathcal{I}_{real}), \label{eq:mosaic}\\
    \mathcal{I}_{udnie} = \mathbb{G}_{udnie}(\mathcal{I}_{real}), \label{eq:udnie} \\
    \mathcal{I}_{starry} = \mathbb{G}_{starry}(\mathcal{I}_{real}). \label{eq:starry}
\end{gather}
where $\mathcal{I}_{mosaic}$, $\mathcal{I}_{udnie}$, and $\mathcal{I}_{starry}$ are the unseen domain images while $\mathbb{G}_{mosaic}$, $\mathbb{G}_{udnie}$, and $\mathbb{G}_{starry}$ represent the GAN checkpoints adopted from~\cite{rusty2018faststyletransfer} for image translation. It is seen in Table~\ref{tab:06_table-MosaicUdnieStarry} that the $\mathcal{DA}$-model consistently outperforms the $\mathcal{SB}$-model for domains unseen during training. When compared to the MS-Transformer, the $\mathcal{DA}$-model achieves improved performance for 17 of 20 instances. We also analyzed if the learned embeddings by the proposed method are robust to domains seen and unseen during training. For that, we qualitatively analyze by plotting embeddings, using dimensionality reduction techniques. The results in Fig.~\ref{fig:tsne-plots} show that the $\mathcal{DA}$-model generates similar embeddings for domains seen and unseen during training, leading to better pose estimation in general. On the other hand, the $\mathcal{SB}$-model generates dissimilar embeddings for domains seen and unseen during training, as observed in the second row of Fig.~\ref{fig:tsne-plots}. We also evaluate AtLoc~\cite{wang2020atloc} for the same task. However, only the \textit{Stairs} scene pretrained model is publicly available, so we provide a detailed evaluation only for that specific scene (see Table~\ref{tab:07_table-AtLoc}).

\input{./tables/06_table-MosaicUdnieStarry.tex}

\input{./tables/07_table-AtLoc}

\input{./figs/04_fig-tsne-plots.tex}

%% file: tables/03_table-FLOPs.tex
\begin{table}[t!]
\centering
\caption{Computational complexity analysis in terms of FLOPs, activations, parameters, and memory. Bold numbers represent the highest efficiency.}
\label{tab:03_table-FLOPs}
\resizebox{0.9\columnwidth}{!}{%
\begin{tabular}{|l|c|c|c|c|}
\hline
\textbf{Methods} & \makecell{\textbf{FLOPs} \\ (G)}  & \makecell{\textbf{Activations} \\ (millions)} & \makecell{\textbf{Params} \\ (millions)} & \makecell{\textbf{Memory} \\ (MB)} \\
\hline
\textbf{Ours}      & \textbf{0.237} & 4.4          & \textbf{3.847} & \textbf{14.2} \\
\textbf{MS-Transformer \cite{shavit2021learning}} & 5.578          & 52.2         & 18.542         & 74.5          \\
\textbf{AtLoc \cite{wang2020atloc}}          & 3.068          & 3.8 & 24.448         & 97.9   \\
\textbf{MapNet \cite{brahmbhatt2018geometry}}          & 3.672          & \textbf{3.7} & 22.348         & 89.5   \\
\hline
\end{tabular}%
}
\end{table}

%% file: tables/04_table-Ablation.tex

\begin{table}[t!]
\centering
\caption{Impact of individual components $\mathsf{L}_{BT}$, $\mathsf{L}_{2}$, and MHA}
\label{tab:04_table-Ablation}
\resizebox{\columnwidth}{!}{%
\setlength{\tabcolsep}{6pt} 

\begin{tabular}{|l|cc|cc|cc|cc|c|}
\hline
\textbf{Components} &
  \multicolumn{2}{c|}{\textbf{ShopFacade}} &
  \multicolumn{2}{c|}{\textbf{OldHospital}} &
  \multicolumn{2}{c|}{\textbf{Chess}} &
  \multicolumn{2}{c|}{\textbf{Heads}} & \makecell{\textbf{Memory}\\(MB)}\\ \cline{2-9}
 &
  \textbf{T(m)} &
  \textbf{R(deg)} &
  \textbf{T(m)} &
  \textbf{R(deg)} &
  \textbf{T(m)} &
  \textbf{R(deg)} &
  \textbf{T(m)} &
  \textbf{R(deg)} & \\
  \hline
\textbf{W/O $\mathsf{L}_{\mathcal{BT}}$}  & 0.85 & 4.05 & 2.14 & 3.04 & 0.11 & 6.76 & 0.16 & 13.21 & 14.2 \\
\textbf{W/O $\mathsf{L}_{2}$}  & 0.81 & 4.29 & 2.09 & 2.97 & 0.11 & 6.30 & 0.15 & 13.15 & 14.2\\
\textbf{W/O MHA} & 0.81 & 4.46 & 1.70 & 3.20 & 0.11 & 6.25 & 0.16 & 13.62 & 14.1\\
\textbf{Full}    & 0.84 & 3.71 & 1.88 & 2.97 & 0.11 & 6.06 & 0.15 & 12.81 & 14.2\\
\hline
\end{tabular}%
}
\end{table}

%% file: tables/05_table-FogNight.tex
\begin{table}[t!]
\caption{Average of median errors for multiple indoor and outdoor scenes on the domains used for training: real, foggy, and night. Bold numbers represent best performance. See more details in Appendix~E.}
\label{tab:05_table-FogNight}
\resizebox{\columnwidth}{!}{%
\begin{tabular}{|l|cc|cc|cc|}
\hline
{}                                                       & \multicolumn{6}{c|}{\textbf{Average of median errors for different methods}} \\
 & \multicolumn{2}{c}{\makecell{\textbf{MS-Transformer}}} &
  \multicolumn{2}{c}{\makecell{\textbf{$\mathcal{SB}$-model}}} &
  \multicolumn{2}{c|}{\makecell{\textbf{$\mathcal{DA}$-model}}} \\
  \cline{2-7}
{\textbf{Scenes}} &
  \textbf{T(m)} &
  \textbf{R(deg)} &
  \textbf{T(m)} &
  \textbf{R(deg)} &
  \textbf{T(m)} &
  \textbf{R(deg)} \\
  \hline
\textbf{KingsCollege}        & 1.38   & 2.89            & 1.53  & 3.19   & \textbf{0.82}  & \textbf{2.33}  \\
\textbf{OldHospital}         & 3.29   & 4.09            & 3.00  & 3.84   & \textbf{2.15}  & \textbf{2.87}  \\
\textbf{ShopFacade}          & 1.11   & 4.43            & 1.45  & 4.87   & \textbf{0.78}  & \textbf{2.83}  \\
\textbf{StMarysChurch}       & 2.09   & 11.92           & 1.89  & 5.36   & \textbf{1.32}  & \textbf{3.78}  \\
\hline
\textbf{Average} & 1.97   & 5.83            & 1.97  & 4.31   & \textbf{1.27}  & \textbf{2.95}  \\
\hline
\textbf{chess}               & 0.12   & 6.70            & 0.16  & 7.32   & \textbf{0.11}  & \textbf{6.11}  \\
\textbf{office}              & 0.21   & \textbf{7.80}   & 0.24  & 9.03   & \textbf{0.17}  & 8.27           \\
\textbf{pumpkin}             & 0.24   & \textbf{5.42}   & 0.31  & 6.85   & \textbf{0.22}  & 6.34           \\
\textbf{stairs}              & 0.37   & \textbf{9.53}   & 0.32  & 11.40  & \textbf{0.27}  & 12.18          \\
\hline
\textbf{Average} & 0.24   & \textbf{7.36}   & 0.26  & 8.65   & \textbf{0.19}  & 8.23      \\
\hline
\end{tabular}%
}
\vspace{-1.0em}
\end{table}

%% file: tables/06_table-MosaicUdnieStarry.tex

\begin{table}[hbt!]
\centering
\caption{Average of median errors for multiple indoor and outdoor scenes on domains unused for training: mosaic, udnie and starry. Bold numbers represent best performance. See more details in Appendix~F.}
\label{tab:06_table-MosaicUdnieStarry}
\resizebox{0.9\columnwidth}{!}{%
\begin{tabular}{|l|cc|cc|cc|}
\hline
{} & \multicolumn{6}{c|}{\textbf{Average of median errors for different methods}}     \\ 
{}  & \multicolumn{2}{c}{\makecell{\textbf{MS-Transformer}}} &
  \multicolumn{2}{c}{\makecell{\textbf{$\mathcal{SB}$-model}}} &
  \multicolumn{2}{c|}{\makecell{\textbf{$\mathcal{DA}$-model}}} \\ \cline{2-7}
{\textbf{Scenes}} &
  \textbf{T(m)} &
  \textbf{R(deg)} &
  \textbf{T(m)} &
  \textbf{R(deg)} &
  \textbf{T(m)} &
  \textbf{R(deg)} \\
\hline
\textbf{KingsCollege}        & 2.93          & 4.97           & 7.66  & 9.38  & \textbf{2.49} & \textbf{4.32}  \\
\textbf{OldHospital}         & 4.45          & 6.21           & 4.56  & 5.19  & \textbf{4.29} & \textbf{4.53}  \\
\textbf{ShopFacade}          & 1.56          & \textbf{7.66}  & 3.39  & 11.81 & \textbf{2.99} & 9.08           \\
\textbf{StMarysChurch}       & 12.02         & 67.01          & 15.14 & 19.37 & \textbf{6.39} & \textbf{9.47}  \\
\hline
\textbf{Average} & 5.24          & 21.46          & 7.69  & 11.44 & \textbf{4.04} & \textbf{6.85}  \\
\hline
\textbf{chess}               & \textbf{0.20} & 12.88          & 0.34  & 11.15 & \textbf{0.20} & \textbf{7.96}  \\
\textbf{office}              & 0.37          & 20.46          & 0.48  & 18.74 & \textbf{0.33} & \textbf{12.51} \\
\textbf{pumpkin}             & 0.48          & 19.03          & 0.55  & 26.94 & \textbf{0.41} & \textbf{11.74} \\
\textbf{stairs}              & 0.44          & \textbf{13.49} & 0.51  & 17.72 & \textbf{0.37} & 15.76          \\
\hline
\textbf{Average} & 0.37          & 16.47          & 0.47  & 18.64 & \textbf{0.33} & \textbf{11.99} \\
\hline
\end{tabular} %
}
\end{table}

%% file: tables/07_table-AtLoc.tex
\begin{table}[hbt!]
\centering
\caption{Comparison with AtLoc~\cite{wang2020atloc}. Average of median errors for unseen domains with Stairs scene. Bold numbers represent best performance.}
\label{tab:07_table-AtLoc}
\resizebox{0.9\columnwidth}{!}{%
\setlength{\tabcolsep}{12pt} 
\begin{tabular}{|l|cc|cc|cc|}
\hline
\multicolumn{1}{|l}{\makecell{\textbf{Stairs} \\ \textbf{scene}}} &
  \multicolumn{2}{|c|}{\makecell{\textbf{AtLoc}}} &
  \multicolumn{2}{c|}{\makecell{\textbf{$\mathcal{SB}$-model}}} &
  \multicolumn{2}{c|}{\makecell{\textbf{$\mathcal{DA}$-model}}} \\
  \hline
\multicolumn{1}{|l|}{} &
  \textbf{T(m)} &
  \textbf{R(deg)} &
  \textbf{T(m)} &
  \textbf{R(deg)} &
  \textbf{T(m)} &
  \textbf{R(deg)} \\ \cline{2-7}
mosaic  & 0.71 & 17.69          & 0.55 & 20.70 & \textbf{0.43} & \textbf{17.16} \\
udnie   & 0.31 & 15.33          & 0.34 & 16.31 & \textbf{0.31} & \textbf{15.26} \\
starry  & 0.75 & \textbf{14.81} & 0.64 & 16.14 & \textbf{0.38} & 14.87          \\
Average & 0.59 & 15.94          & 0.51 & 17.72 & \textbf{0.37} & \textbf{15.76} \\
\hline
\end{tabular}%
}
\end{table}

%% file: figs/04_fig-tsne-plots.tex
\begin{figure}[t]
\centering
\captionsetup[subfigure]{labelformat=empty}
\begin{tabular}{cc}
    {\footnotesize \rotatebox{90}{\thinspace\thinspace$\mathcal{DA}$-model}}&\subfloat[]{\includegraphics[width = 0.85\columnwidth]{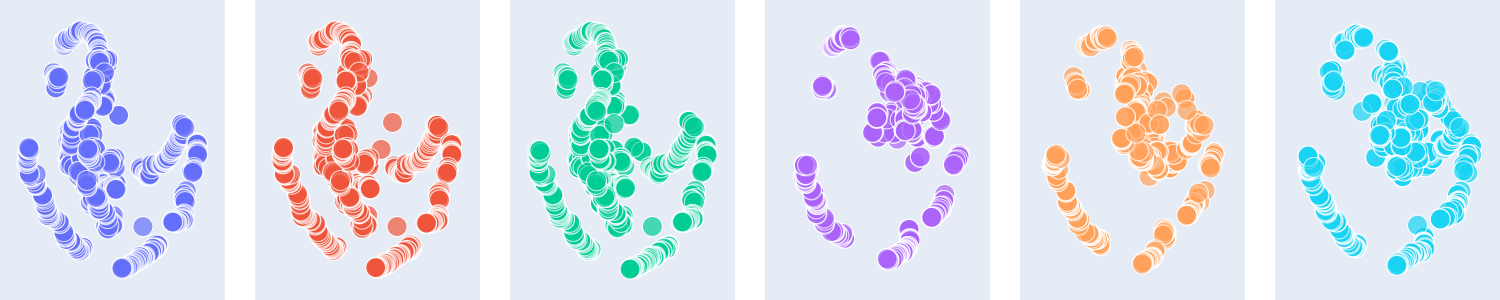}} \\[-2.5ex]
    {\footnotesize \rotatebox{90}{\thinspace\thinspace$\mathcal{SB}$-model}}&\subfloat[]{\includegraphics[width = 0.85\columnwidth]{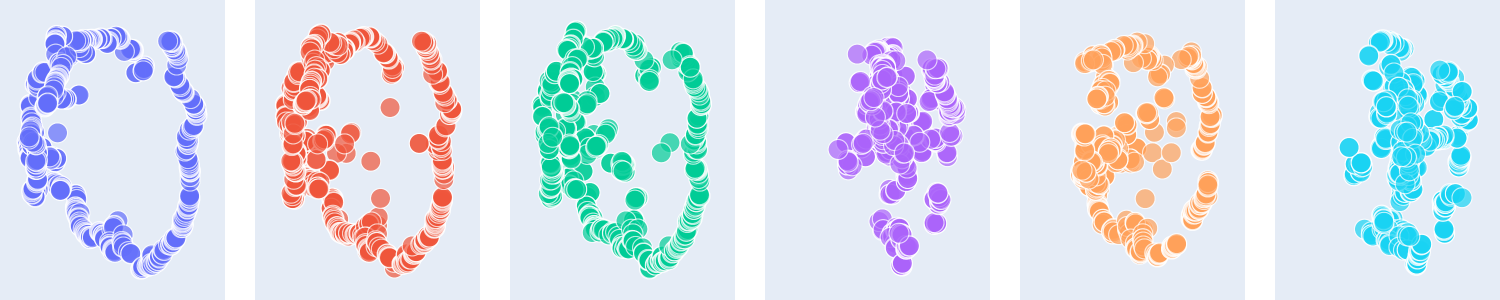}}
\end{tabular}
\caption{Visualization of the embeddings produced by $\mathcal{SB}$-model and $\mathcal{DA}$-model for the test split of StMarysChurch scene in 2 dimensions by using t-SNE~\cite{van2008visualizing, plotly} plots. Each subplot represents the latent embedding space of different domains. From the $1^{st}$ row, it can be seen that $\mathcal{DA}$-model produces similar embeddings to seen and unseen domains while the $2^{nd}$ row, shows that $\mathcal{SB}$-model produces dissimilar embeddings. From left to right, the domains are real, foggy, night, mosaic, udnie, and starry.}
\vspace{-1.8em}
\label{fig:tsne-plots}
\end{figure}

%% file: 10_conclusion.tex
\section{Conclusion}
\label{sec:conclusion}

In this paper, we propose a domain-adaptive training framework for improving generalization to unseen distributions during inference. Our model is as light as a single branch APR during inference. We exploit the \textsc{Barlow Twins} objective which is introduced for a self-supervised learning framework to enforce domain invariance and redundancy reduction in learned latent embeddings. Even with a lightweight architecture, the $\mathcal{DA}$-model ranks 2nd with the Cambridge landmarks dataset and ranks 4th with the 7Scenes dataset. Also, our method outperforms the state-of-the-art MS-Transformer with unseen domains. By analysis, we prove that our model is lightweight in terms of the number of parameters, memory, FLOPs and activations. Our domain adaptive framework is evaluated with a MobileNetV3-Large based APR. However, the proposed training framework is general and can be used to retrain any available APRs to improve their generalization to unseen domains.

%% file: 12_appendix.tex
{\raggedleft \LARGE \textbf{Appendix}}

\appendix
\label{sec:appendix}

\input{supplementary/A_MobileNetV3.tex}

\input{supplementary/B_Environment_hyperparameter.tex}

\input{supplementary/C_algo_pseudo_code.tex}

\input{supplementary/D_Analysis_hist_traj.tex}

\input{supplementary/E_seen_domains.tex}

\input{supplementary/F_unseen_domains.tex}

\input{supplementary/G_CNN_backbones.tex}

\input{supplementary/H_review_sfm_IR.tex}

\input{supplementary/I_Gan_samples.tex}

%% file: supplementary/A_MobileNetV3.tex
\section{MobileNetV3 Architecture}

\newcommand\relus{\operatorname{ReLU6}}
\newcommand\relu{\operatorname{ReLU}}
\newcommand{\sigmoid}{\operatorname{\sigma}}
\newcommand{\swish}{\operatorname{swish}}
\newcommand{\hardsigmoid}{\operatorname{h-sigmoid}}
\newcommand{\hardswish}{\operatorname{h-swish}}
\newcommand{\externalfigtab}{\textcolor{red}}

In~\cite{howard2019searching}, Howard et al. proposed a MobileNet architecture with a complementary approach using search procedures such as platform-aware network architecture search(NAS)~\cite{zoph2016neural} and NetAdapt~\cite{yang2018netadapt}. Combining those techniques and novel architecture advances, MobileNetV3 demonstrated its efficacy in improving the state of the art in classification, object detection and dense prediction tasks i.e., segmentation. The MobileNet model architecture found through the search procedure was observed to be computationally heavy in earlier layers and a few of the last layers. Therefore, to enhance efficiency, those specific layers are modified such that the model maintains accuracy. Furthermore, layers were upgraded with a novel version of the swish~(\ref{eq:swish})\cite{ramachandran2017searching} nonlinear activation functions called h-swish(\ref{eq:hardswish}) to enable faster computation with quantization-friendly properties. It also leveraged squeeze and excitation blocks~\cite{hu2018squeeze} and replaced the sigmoid function with an efficient hard-sigmoid(\ref{eq:hardsigmoid})~\cite{avenash2019semantic}.
\begin{gather}
\swish (x) = x \cdot \sigmoid(x) \label{eq:swish}\\
\hardsigmoid [x] =\frac{\relus(x + 3)}{6} \label{eq:hardsigmoid}\\
\hardswish [x] = x \frac{\relus(x + 3)}{6} \label{eq:hardswish}
\end{gather}

where $\sigmoid(.)$ represents the conventional sigmoid function while $\relus$ is adopted from ~\cite{krizhevsky2010convolutional}. Please check Tables \externalfigtab{1}, \externalfigtab{2} and Fig. \externalfigtab{6} of~\cite{howard2019searching} for layer-wise details, similarity between $\swish$ and $\hardswish$ and similarity between sigmoid($\sigmoid$) and hard-sigmoid. Also, additional explanations on the efficient redesign of the original last stage architectures found by search methods are presented in Fig.~\externalfigtab{5} of~\cite{howard2019searching}. Two MobileNetV3 models were investigated in \cite{howard2019searching}: MobileNetV3-Large and MobileNetV3-Small. More details regarding the search performed in this paper to select the final CNN backbone are presented in Appendix~G.

%% file: supplementary/B_Environment_hyperparameter.tex
\section{Environment and Hyperparameters}

The proposed method is implemented in PyTorch \cite{paszke2017automatic} and inspired by \cite{shavit2022camera}. The models are trained on a \textit{Ubuntu 18.04} platform \textit{32GB RAM} machine with 2 $\times$ \textit{NVIDIA GeForce GTX 1080 Ti} GPUs. In the multi-head attention module, the number of heads is set to be \textit{7} as the embedding dimension(\textit{49}) needs to be divisible by the number of heads according to \cite{vaswani2017attention, paszke2017automatic}. Attention dropout is set at \textit{0.0025}. We trained all the 7Scenes APRs up to \textit{120} epochs with a learning rate scheduler~\cite{paszke2017automatic} with the step-size set to \textit{20}, gamma set to \textit{0.5}, and batch size set to \textit{32}. Cambridge Landmarks APRs are trained for \textit{600} epochs with a learning rate scheduler with the step-size set to \textit{150}, gamma set to  \textit{0.1}, and batch size set to \textit{64}. The latent embedding dimension $D$ is equal to \textit{256} for Cambridge Landmarks and equal to \textit{512} for 7Scenes.

%% file: supplementary/C_algo_pseudo_code.tex
\section{Review of details in the algorithm}

\subsection{\textsc{Barlow Twins} objective~\cite{zbontar2021barlow}}
The cross-correlation matrix($\mathcal{C}$), for latent embeddings ($z^A$ and $z^B$), is calculated similarly to~\cite{zbontar2021barlow} as follows

\begin{gather}
\mathcal{C}_{ij} \triangleq \frac{
\sum_b z^A_{b,i} z^B_{b,j}}
{\sqrt{\sum_b {(z^A_{b,i})}^2} \sqrt{\sum_b {(z^B_{b,j})}^2}} \label{eq:cross-corr-calc}
\end{gather}

where $i,j$ are indexes to the embeddings and $b$ are indexes to the batch samples in eqn.(\ref{eq:cross-corr-calc}) as explained in~\cite{zbontar2021barlow}. The \textsc{Barlow Twins} objective as presented in this section and used to train our method is shown in detail in  Algorithm~\ref{alg:barlow_twins_obj}.
\input{supplementary/C_L_BT.tex}

\subsection{Pose loss~\cite{kendall2017geometric}}
The geometric loss function adopted from~\cite{kendall2017geometric} is used in this work. $\mathsf{L}_{\mathbf{x}}$ measures the norm of deviation between predicted translation and ground truth translation while $\mathsf{L}_{\mathbf{q}}$ measures the deviation between ground truth quaternion and unit normalized quaternion of the prediction. 

\begin{gather}
\mathsf{L}_{\mathcal{P}}=\mathsf{L}_{\mathbf{x}}\exp (-\hat{s_{\mathbf{x}}})+\hat{s_{\mathbf{x}}}+\mathsf{L}_{\mathbf{q}}\exp (-\hat{s_{\mathbf{q}}})+\hat{s_{\mathbf{q}}} \label{eq:geometric-learnable-pose-loss}
\end{gather}

where $\hat{s_{\mathbf{x}}}$ and 
$\hat{s_{\mathbf{q}}}$ take arbitrary initial values in eqn.(\ref{eq:geometric-learnable-pose-loss}) as they are parameters learned during training.

\subsection{Foggy and Night view images augmentation}
\label{subsec:supp_fog_night_view_aug}

ManiFest~\cite{pizzati2021manifest} code and checkpoints are publicly available\footnote{\url{https://github.com/cv-rits/ManiFest}}. Also, CoMoGAN~\cite{pizzati2021comogan} code and checkpoints are publicly available \footnote{\url{https://github.com/cv-rits/CoMoGAN}}. 
The \texttt{clear2fog.pth} checkpoint from ManiFest is used to augment all the images in the original datasets~\cite{kendall2015posenet,glocker2013real} to foggy images. Similarly, in the case of CoMoGAN, the checkpoint in the official repository is used with the sun's angle $\phi$ being set to \textit{2.1} to augment all the images in the original datasets to night view images.

%% file: supplementary/C_L_BT.tex
\definecolor{codegreen}{rgb}{0,0.6,0}
\definecolor{codegray}{rgb}{0.5,0.5,0.5}
\definecolor{codepurple}{rgb}{0.58,0,0.82}
\definecolor{backcolour}{rgb}{0.95,0.95,0.92}
\lstdefinestyle{mystyle}{
    backgroundcolor=\color{backcolour},   
    commentstyle=\color{codegreen},
    keywordstyle=\color{magenta},
    numberstyle=\tiny\color{codegray},
    stringstyle=\color{codepurple},
    basicstyle=\ttfamily\footnotesize,
    breakatwhitespace=false,         
    breaklines=true,                 
    captionpos=b,                    
    keepspaces=true,                 
    numbers=left,                    
    numbersep=5pt,                  
    showspaces=false,                
    showstringspaces=false,
    showtabs=false,                  
    tabsize=2
}

\begin{algorithm*}[!ht]

   \caption{PyTorch pseudo-code for the Barlow Twins objective used to train the proposed method. The pseudo-code is inspired by~\cite{zbontar2021barlow}}
   \label{alg:barlow_twins_obj}
   
    \definecolor{codeblue}{rgb}{0.25,0.5,0.5}

\begin{lstlisting}[language=python,  style=mystyle]
# lambda: weight on the off-diagonal terms
# N: batch size
# D: dimensionality of the embeddings
# mm: matrix-matrix multiplication
# off_diagonal: off-diagonal elements of a matrix
# eye: identity matrix
class BarlowTwinsLoss(nn.Module):
    """
    A class to represent simple Barlow Twins objective
    """
    def __init__(self, lambd=0.0051, eps= 1e-5):
        super(BarlowTwinsLoss, self).__init__()
        self.lambd = lambd
        self.eps = eps # epsilon for numerical stability
    def off_diagonal(self, x):
        # return a flattened view of the off-diagonal elements of a square matrix
        n, m = x.shape
        assert n == m
        return x.flatten()[:-1].view(n - 1, n + 1)[:, 1:].flatten()
    def forward(self, z_a, z_b):
        """
        :param z_a: (torch.Tensor) batch of real latents, a NxD tensor
        :param z_b: (torch.Tensor) batch of augmented domain latents, a NxD tensor
        :return: invariance term, redundancy reduction term
        """
        # normalize repr. along the batch dimension
        z_a_norm = (z_a - z_a.mean(0)) / (z_a.std(0) + self.eps) # NxD
        z_b_norm = (z_b - z_b.mean(0)) / (z_b.std(0) + self.eps) # NxD

        # cross-correlation matrix
        N = z_a.shape[0]
        c = torch.mm(z_a_norm.T, z_b_norm) / N # DxD
     
        # loss
        D = z_a.shape[1] # Embedding dimension
        c_diff = (c - torch.eye(D).to('cuda')).pow(2) # DxD

        # multiply off-diagonal elements of c_diff by lambda
        redundancy_reduction = self.off_diagonal(c_diff)
        invariance_term = torch.diagonal(c_diff)
        return invariance_term.sum(), self.lambd * redundancy_reduction.sum() 
\end{lstlisting}

\end{algorithm*}

%% file: supplementary/D_Analysis_hist_traj.tex
\section{Qualitative analysis}
\input{./figs/05_supp_fig-histogram.tex}
\input{./figs/06_supp_fig-sb-trajectory.tex}
\input{./figs/07_supp_fig-tsne-plots.tex}

In Fig.~\ref{fig:histogram}, a histogram with the predictions of $\mathcal{DA}$-model, $\mathcal{SB}$-model and MS-Transformer~\cite{shavit2021learning} in the test splits of Cambridge landmarks~\cite{kendall2015posenet} and 7Scenes~\cite{glocker2013real} benchmarks on real distributions is shown. It can be seen that overall, the $\mathcal{DA}$-model has a better estimate than the $\mathcal{SB}$-model. However, MS-Transformer has better estimation than other methods. Trajectories of the $\mathcal{SB}$-model for the same sequences as that of Fig. \externalfigtab{3}(main text) are presented in Fig.~\ref{fig:sb-trajectory}. It can be observed that even the $\mathcal{SB}$-model can generate stable trajectories in the test sequences, unlike MS-Transformer. Similar to Fig. \externalfigtab{4}(main text), the latent embedding visualizations are provided for additional scenes in Fig.~\ref{fig:tsne-plots-additional}. It is seen that $\mathcal{DA}$-model almost obtains identical embeddings for seen domains(real, foggy, night) and reasonably closer embeddings to unseen domains(mosaic, udnie, starry). To unseen domains, however, the SB-model limitations are apparently visible.

%% file: figs/05_supp_fig-histogram.tex
\begin{figure*}[ht]
\captionsetup[subfigure]{labelformat=empty}
\centering
\newcommand{\histwidth}{0.29}

\begin{tabular}{cccc}
& $\mathcal{DA}$-model(Ours) & $\mathcal{SB}$-model(Ours) & MS-Transformer~\cite{shavit2021learning} \\
{\rotatebox{90}{\quad Cambridge T(m)}} &
\subfloat[]{\includegraphics[width = \histwidth \textwidth]{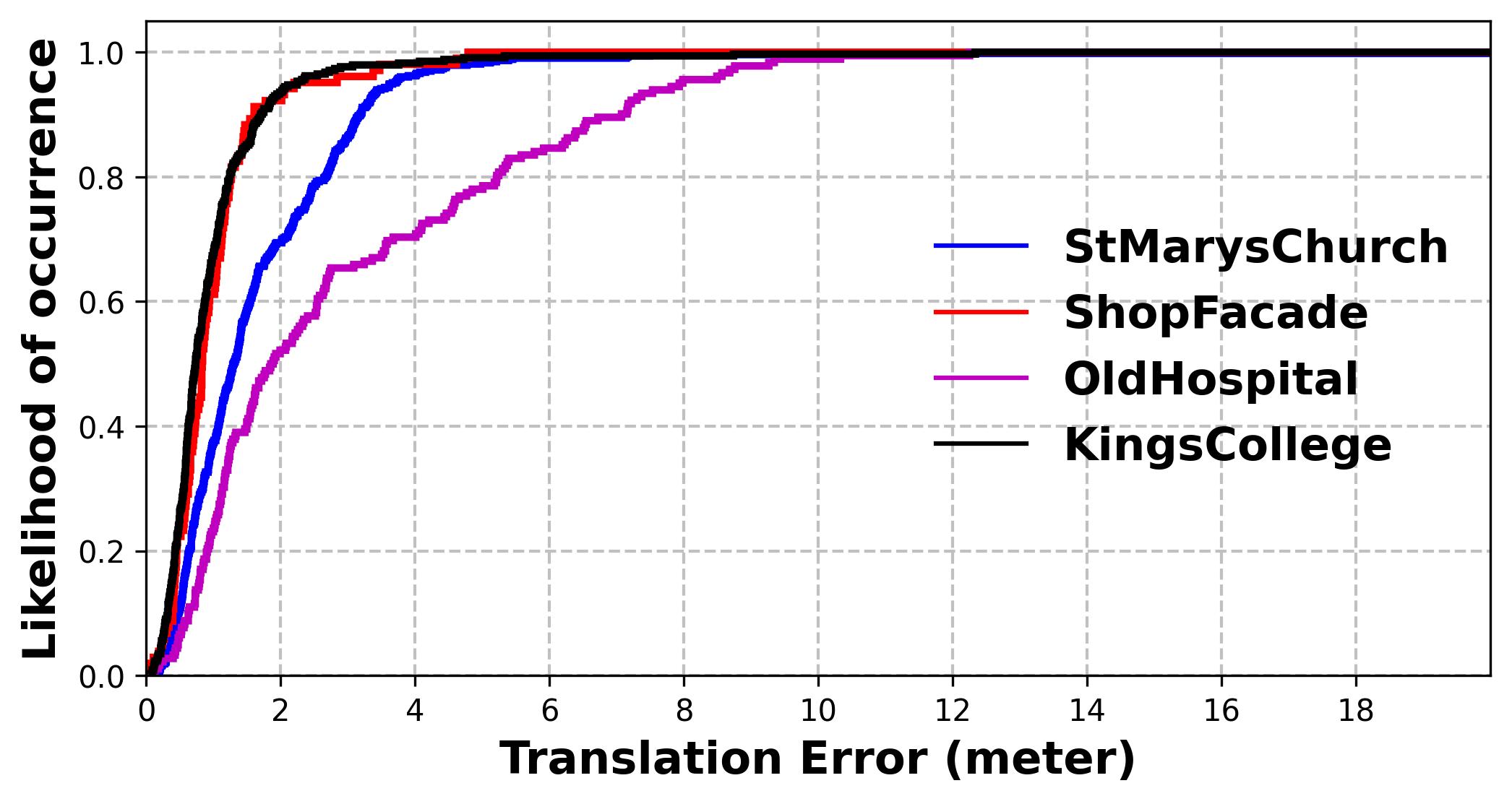}} &
\subfloat[]{\includegraphics[width = \histwidth \textwidth]{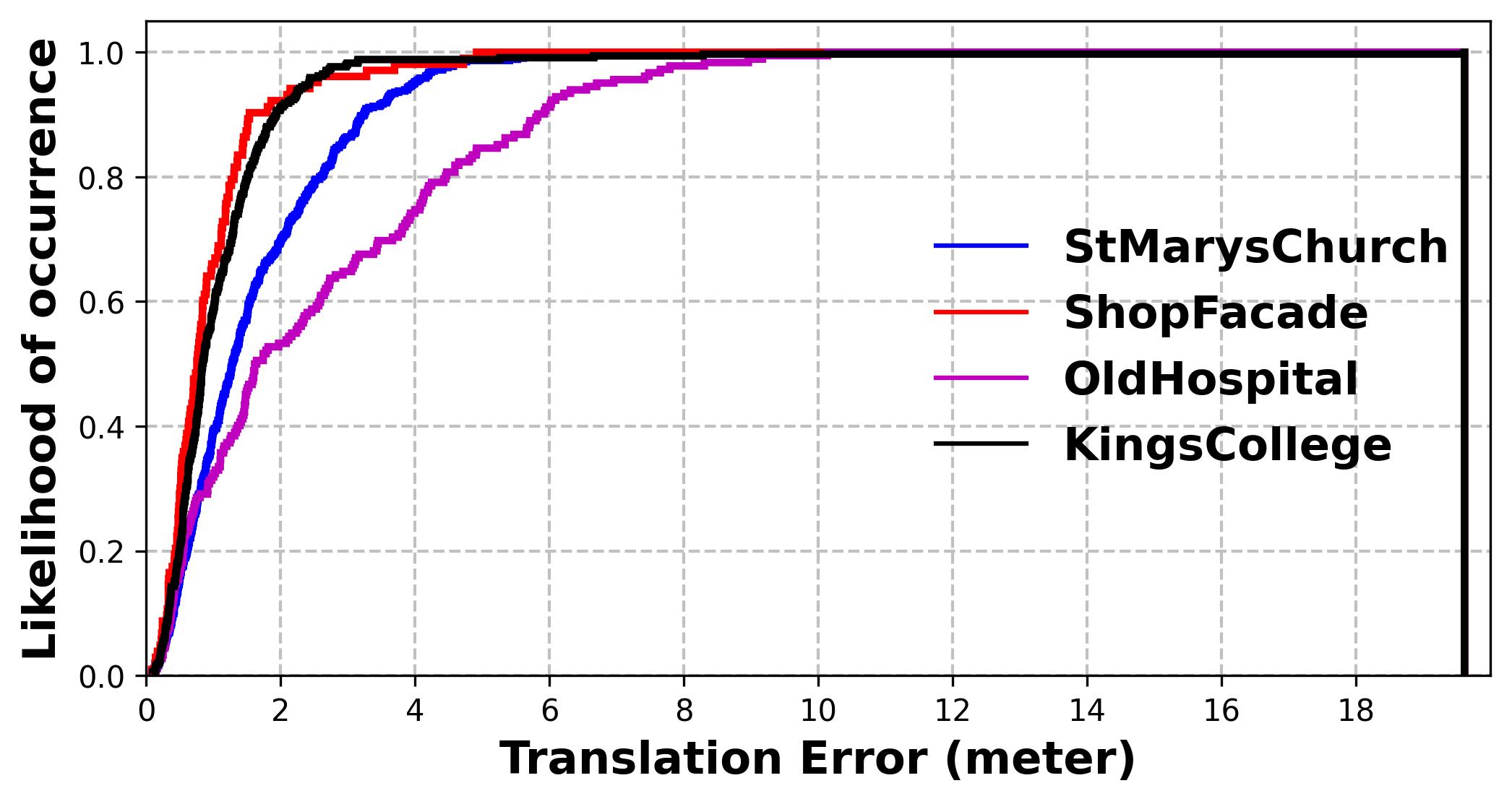}} &
\subfloat[]{\includegraphics[width = \histwidth \textwidth]{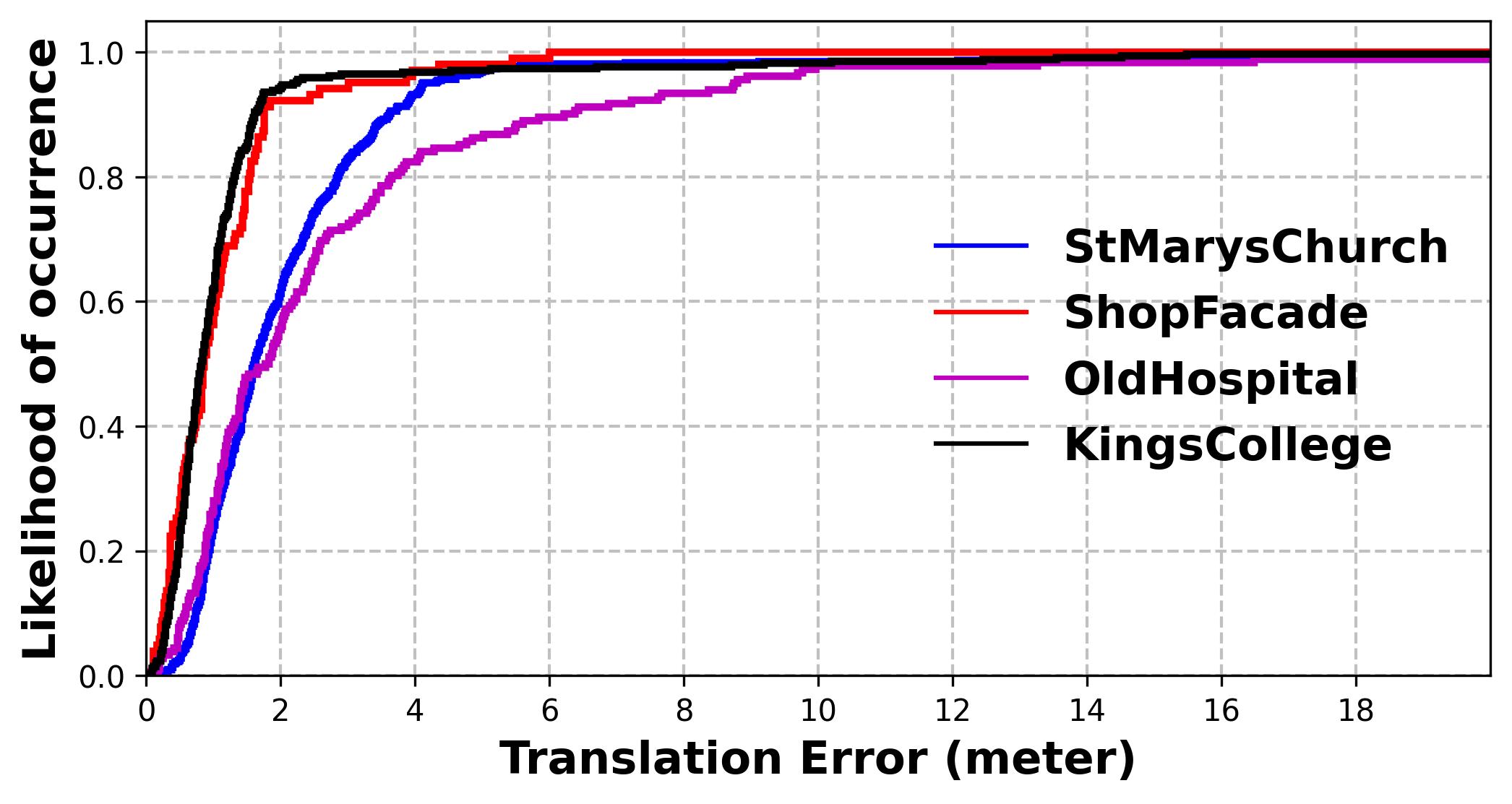}} \\[-4.0ex]
{\rotatebox{90}{\quad Cambridge R(deg)}} &
\subfloat[]{\includegraphics[width = \histwidth \textwidth]{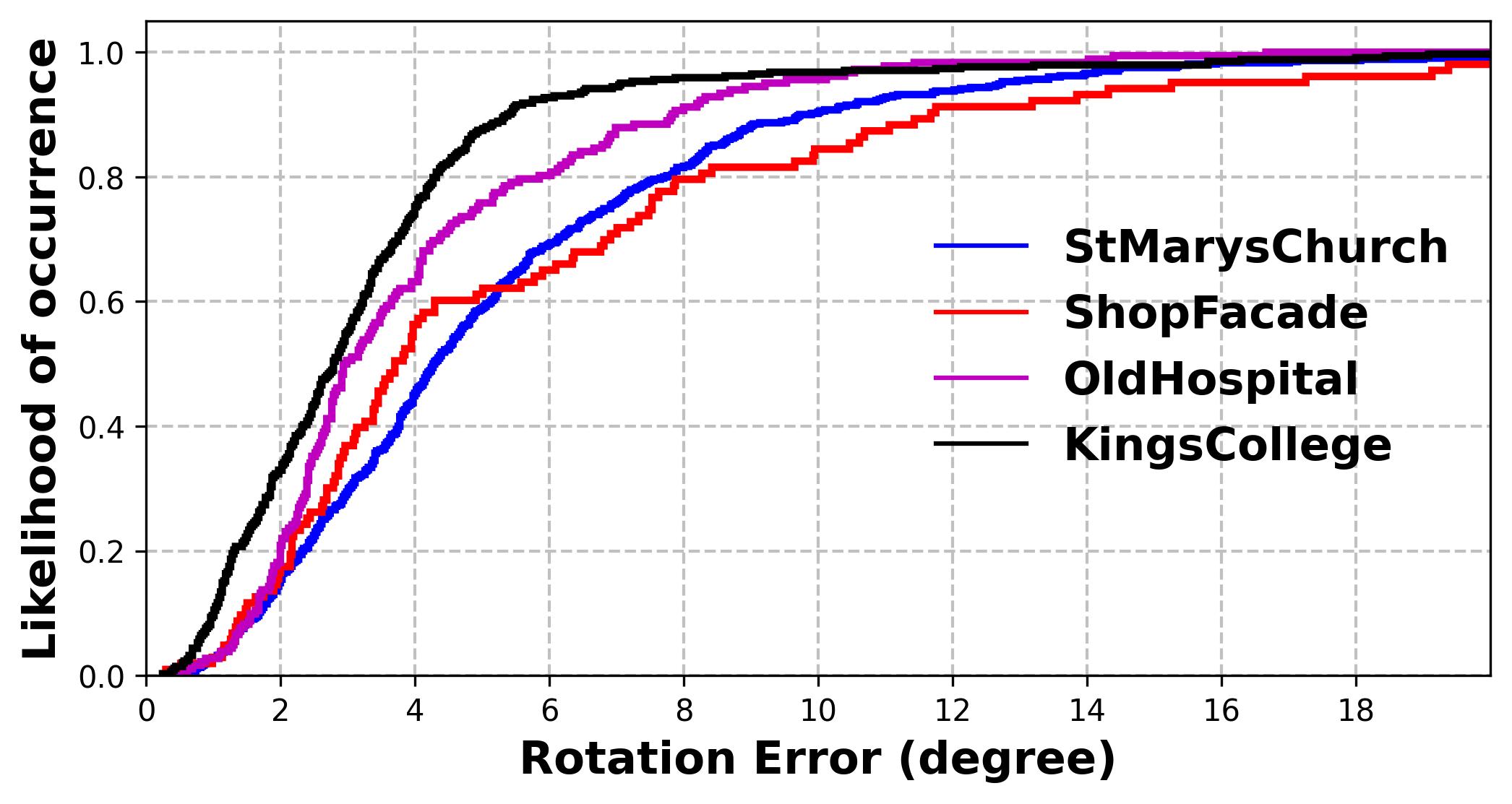}} &
\subfloat[]{\includegraphics[width = \histwidth \textwidth]{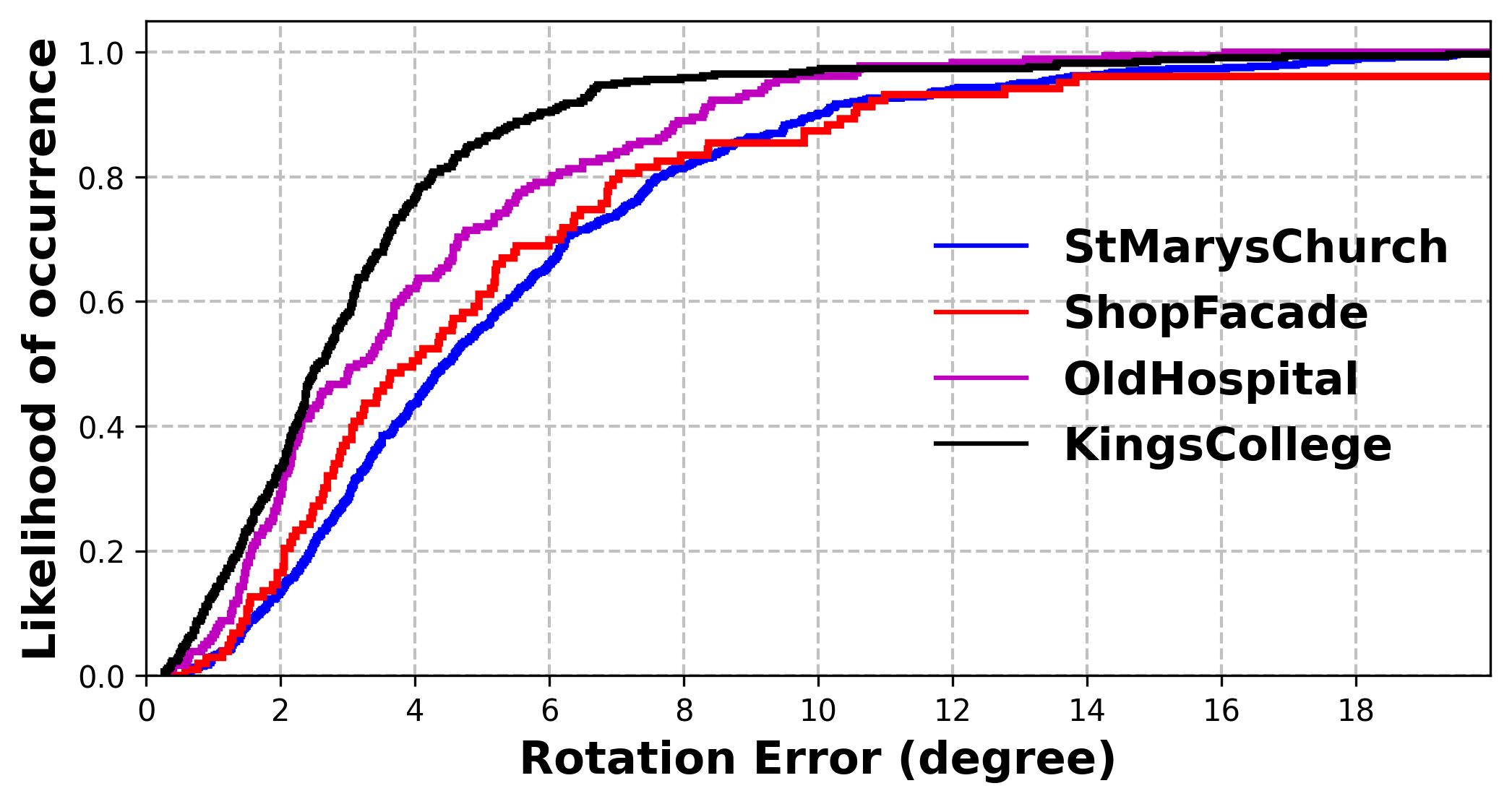}} &
\subfloat[]{\includegraphics[width = \histwidth \textwidth]{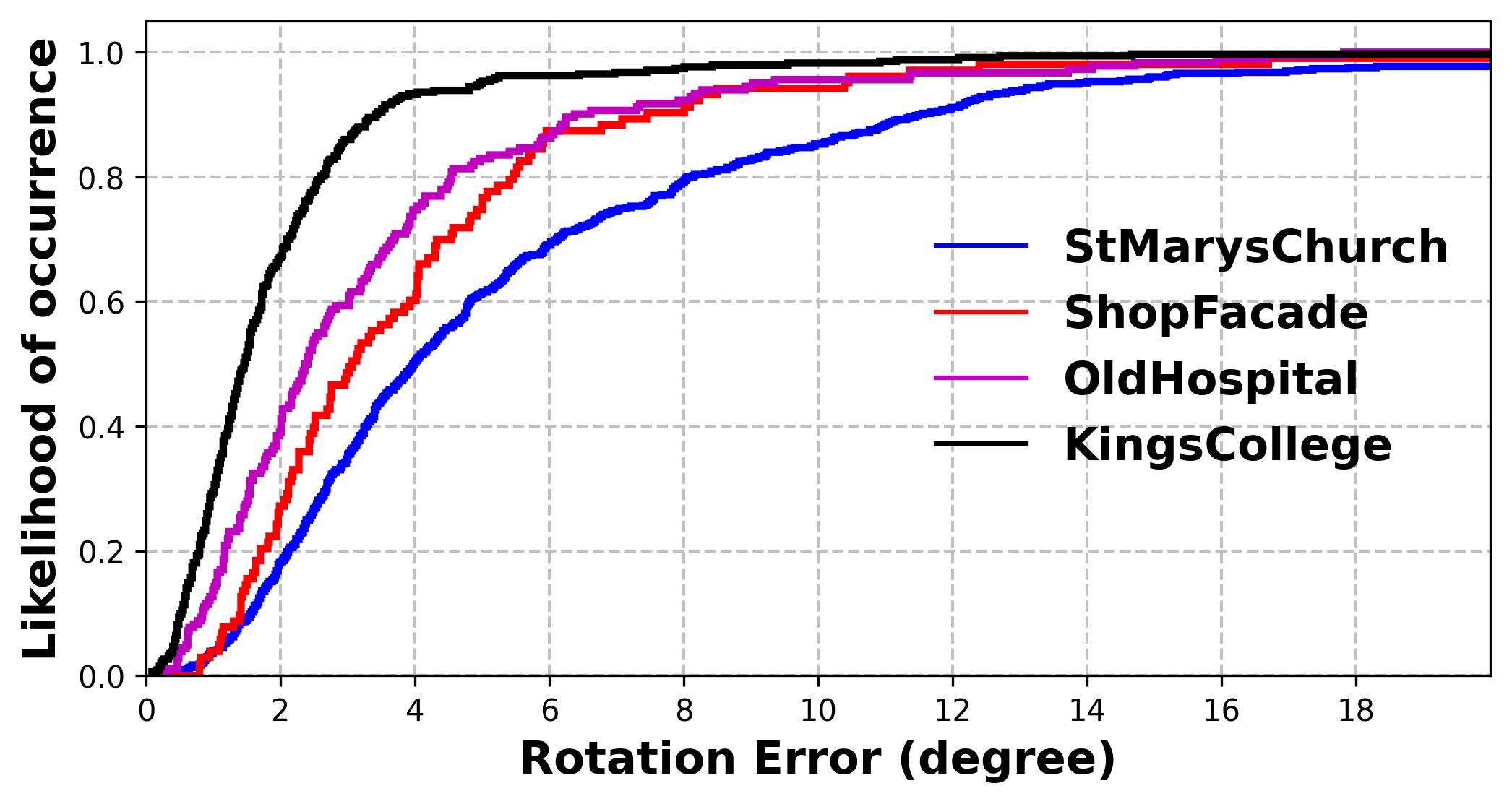}} \\ [-2.5ex]
{\rotatebox{90}{\qquad 7Scenes T(m)}} &
\subfloat[]{\includegraphics[width = \histwidth \textwidth]{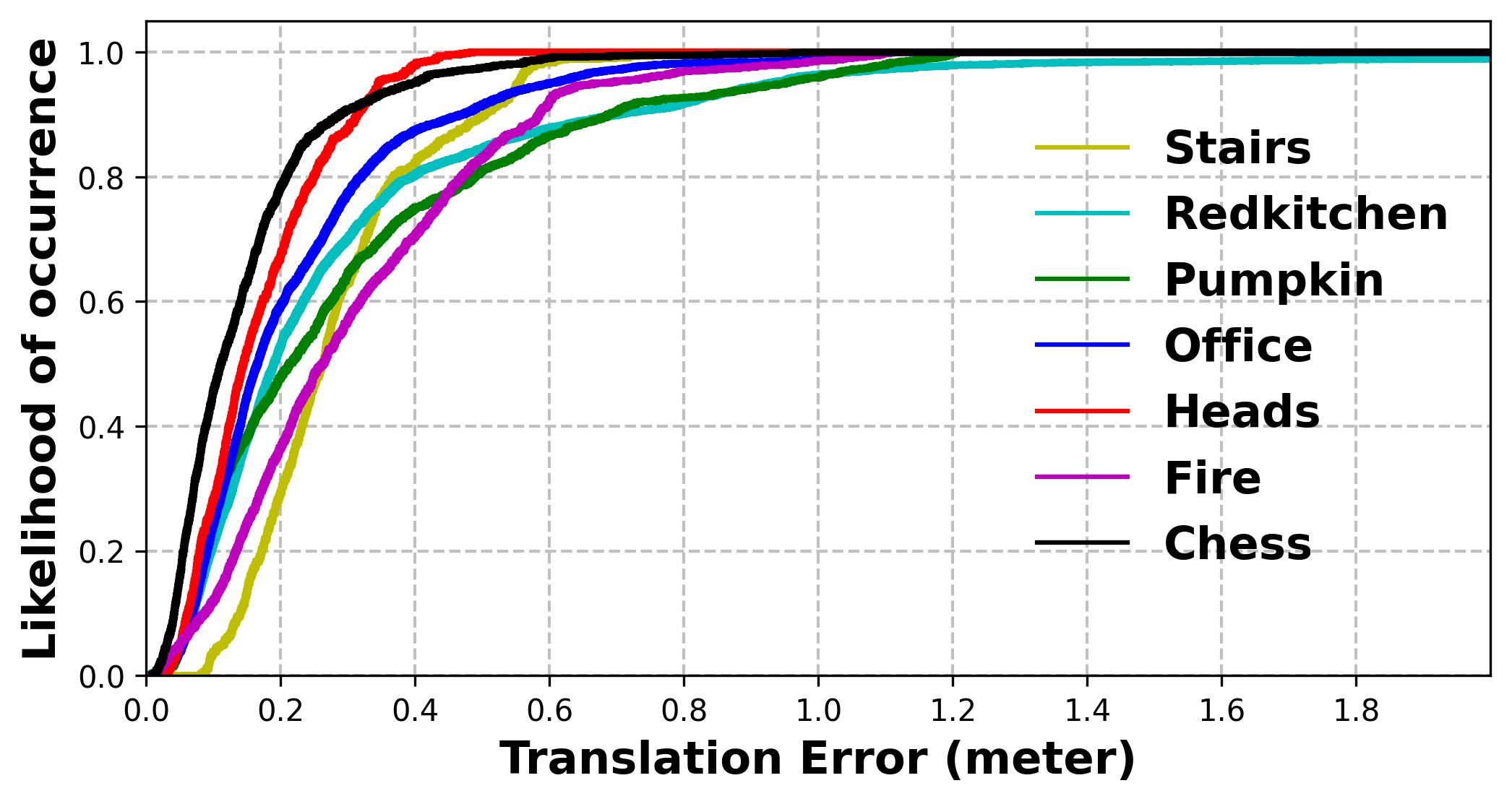}} &
\subfloat[]{\includegraphics[width = \histwidth \textwidth]{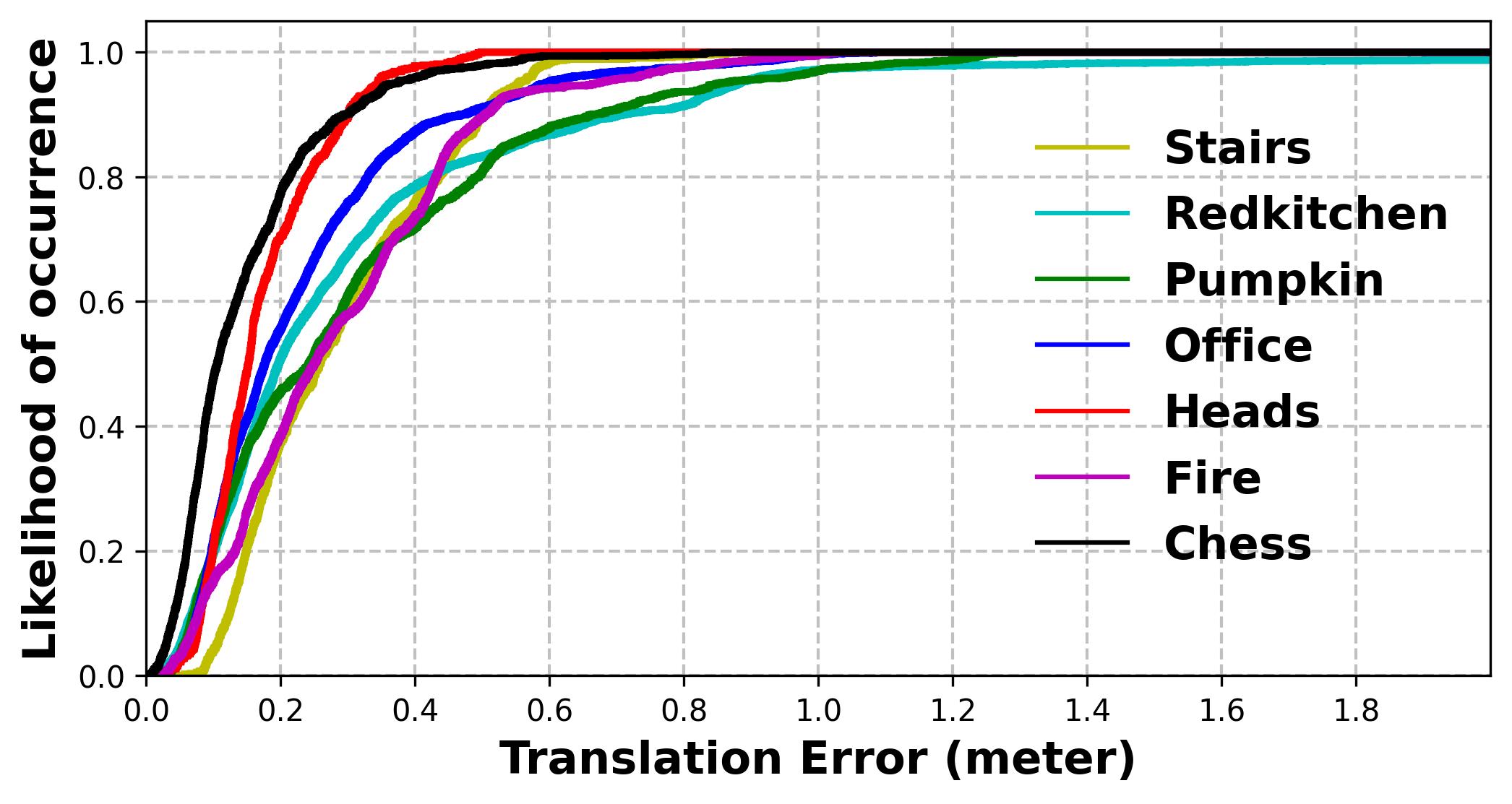}} &
\subfloat[]{\includegraphics[width = \histwidth \textwidth]{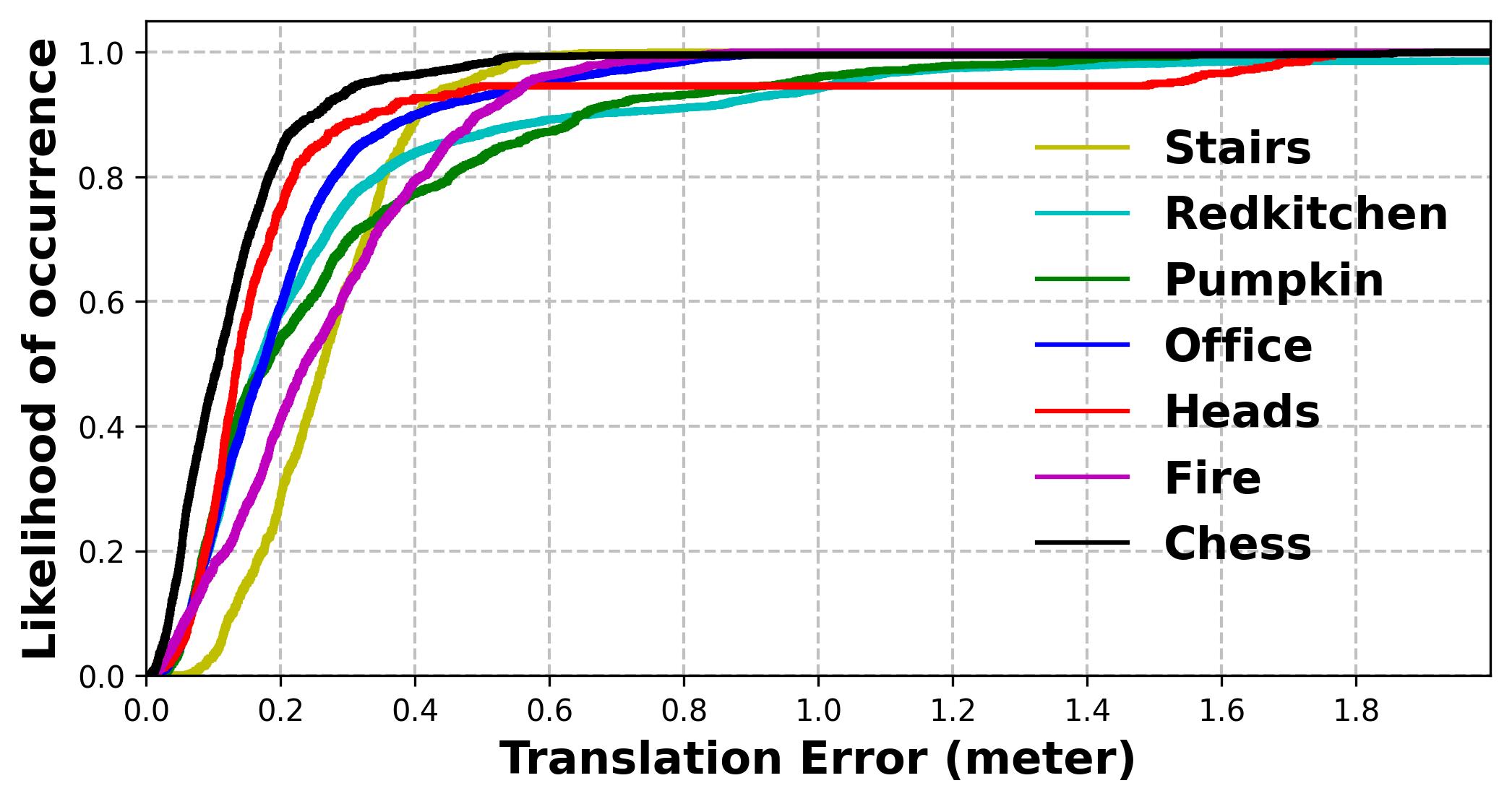}} \\[-4.0ex]
{\rotatebox{90}{\qquad 7Scenes R(deg)}} &
\subfloat[]{\includegraphics[width = \histwidth \textwidth]{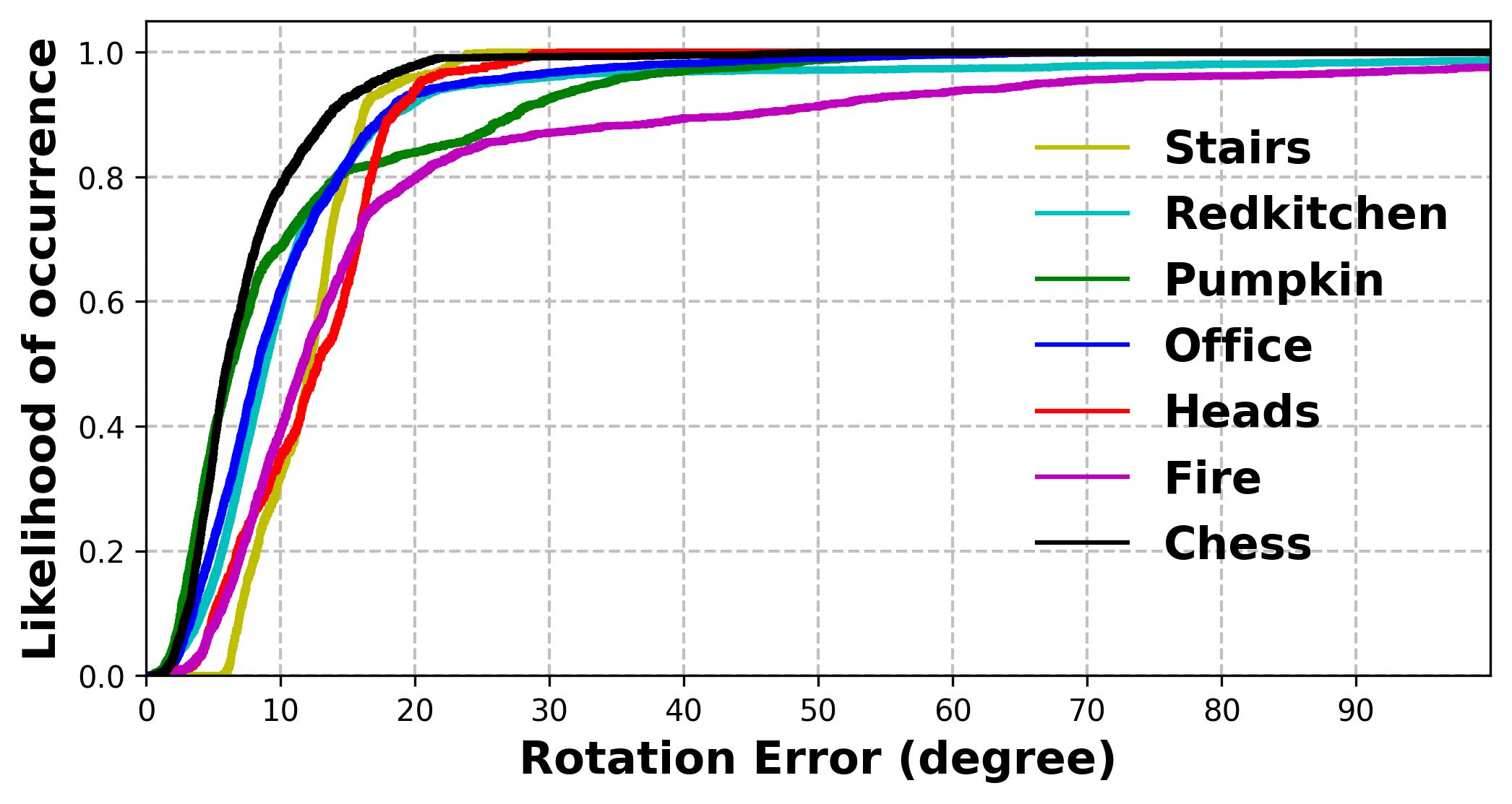}} &
\subfloat[]{\includegraphics[width = \histwidth \textwidth]{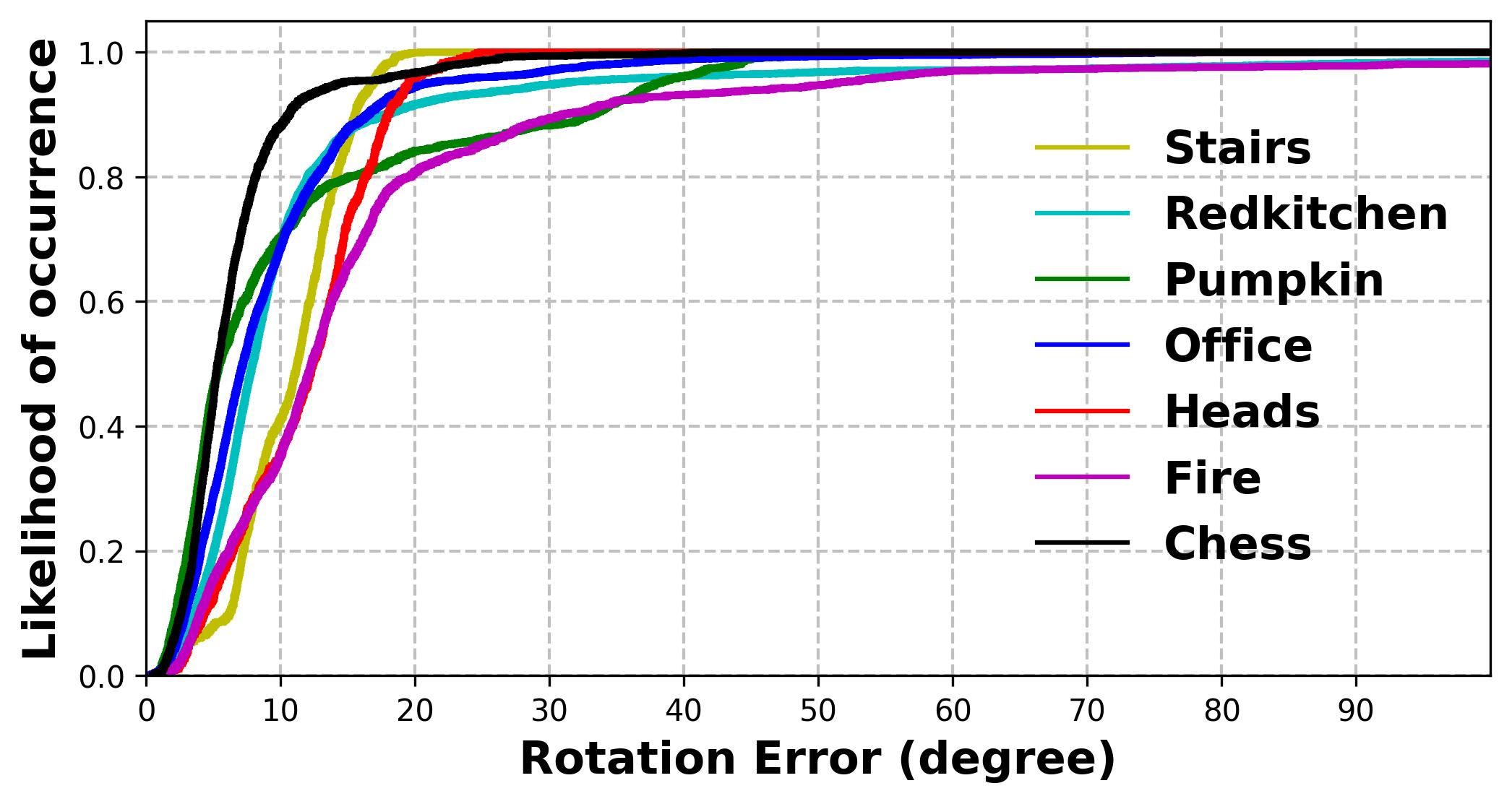}} &
\subfloat[]{\includegraphics[width = \histwidth \textwidth]{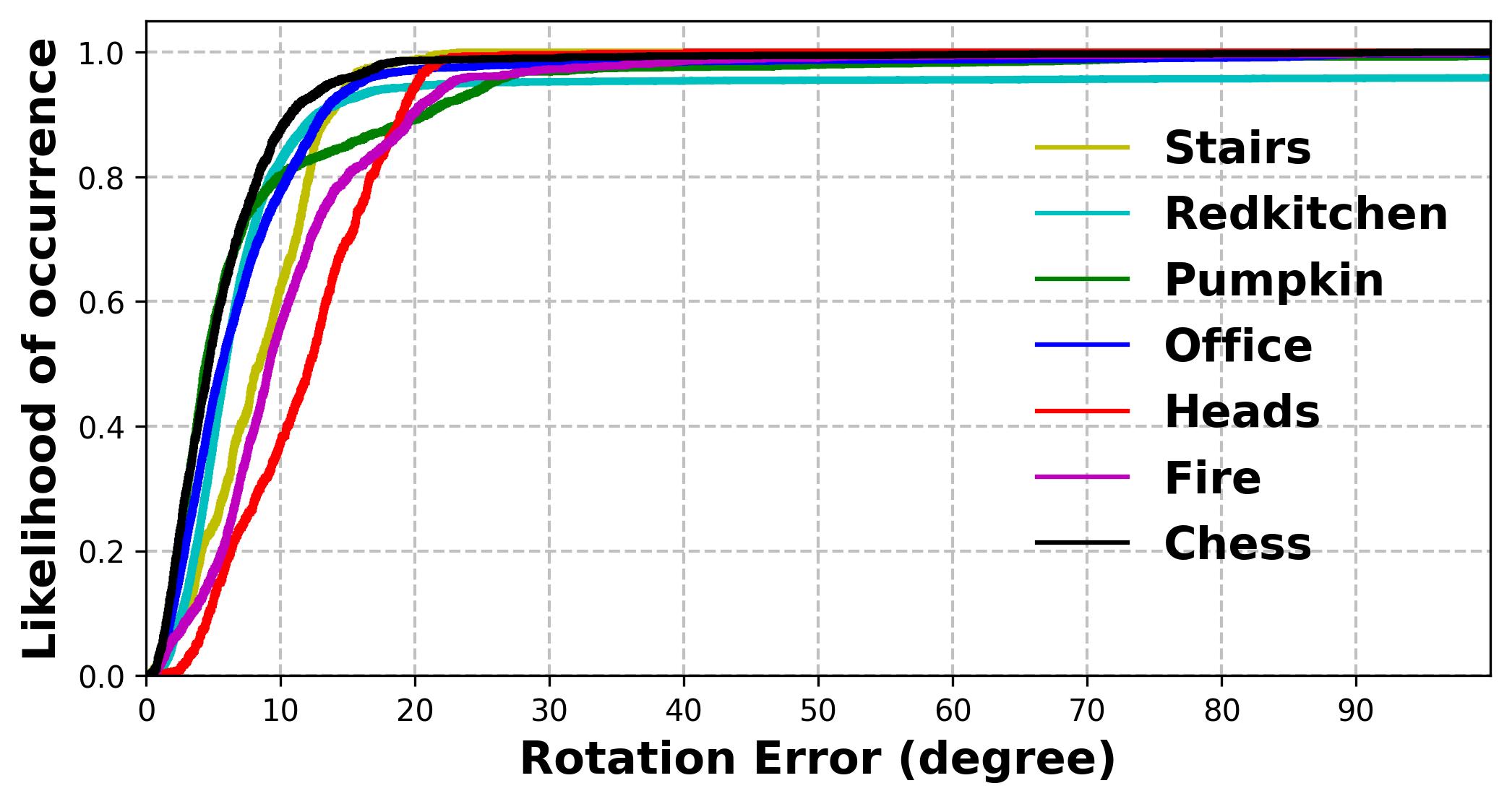}} \\

\end{tabular}

\caption{Cumulative histogram of prediction errors in the test splits with rotation(deg) and translation(m). $1^{st}$, $2^{nd}$, $3^{rd}$ columns, depict the performance of $\mathcal{DA}$-model, $\mathcal{SB}$-model, and MS-Transformer~\cite{shavit2021learning}, respectively. The accuracy of the estimation increases with the area under the curve.}
\label{fig:histogram}
\end{figure*}

%% file: figs/06_supp_fig-sb-trajectory.tex
\begin{figure*}[ht]
\captionsetup[subfigure]{labelformat=empty}
\centering
\begin{tabular}{ccccc}
{\rotatebox{90}{\qquad\quad$\mathcal{SB}$-model}}&
\subfloat[]{\includegraphics[width = 0.20\textwidth]{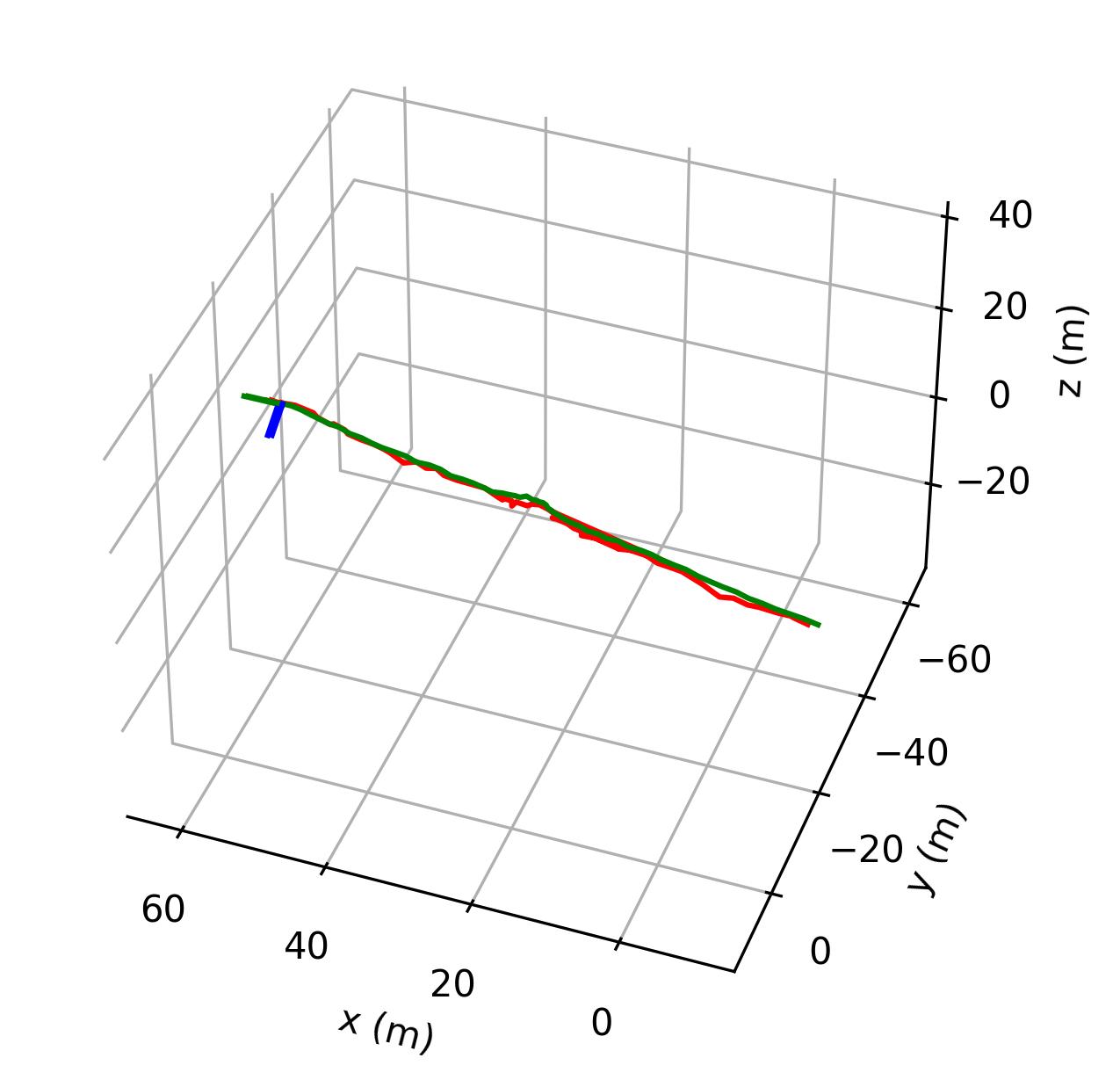}} &
\subfloat[]{\includegraphics[width = 0.20\textwidth]{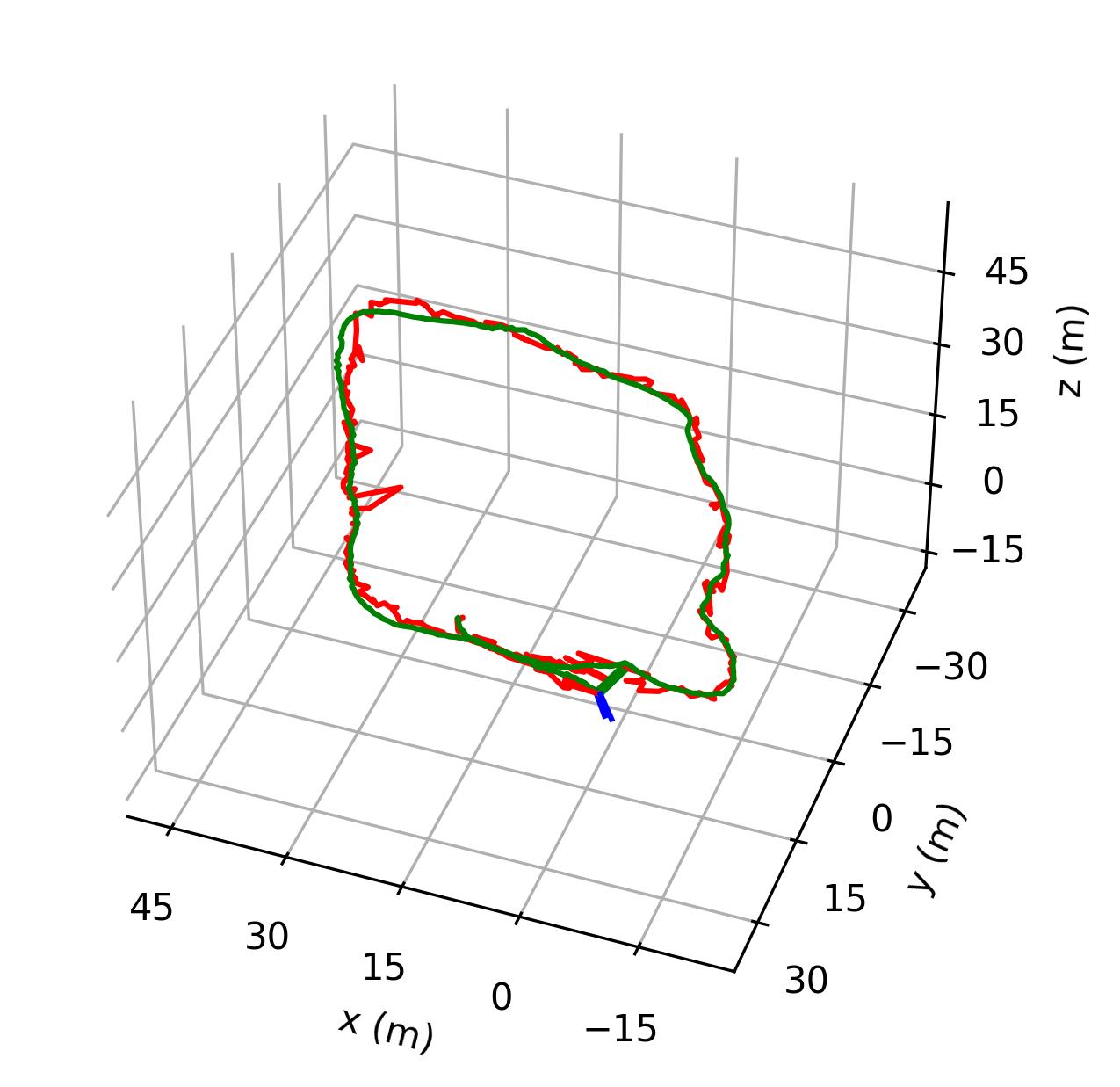}} &
\subfloat[]{\includegraphics[width = 0.20\textwidth]{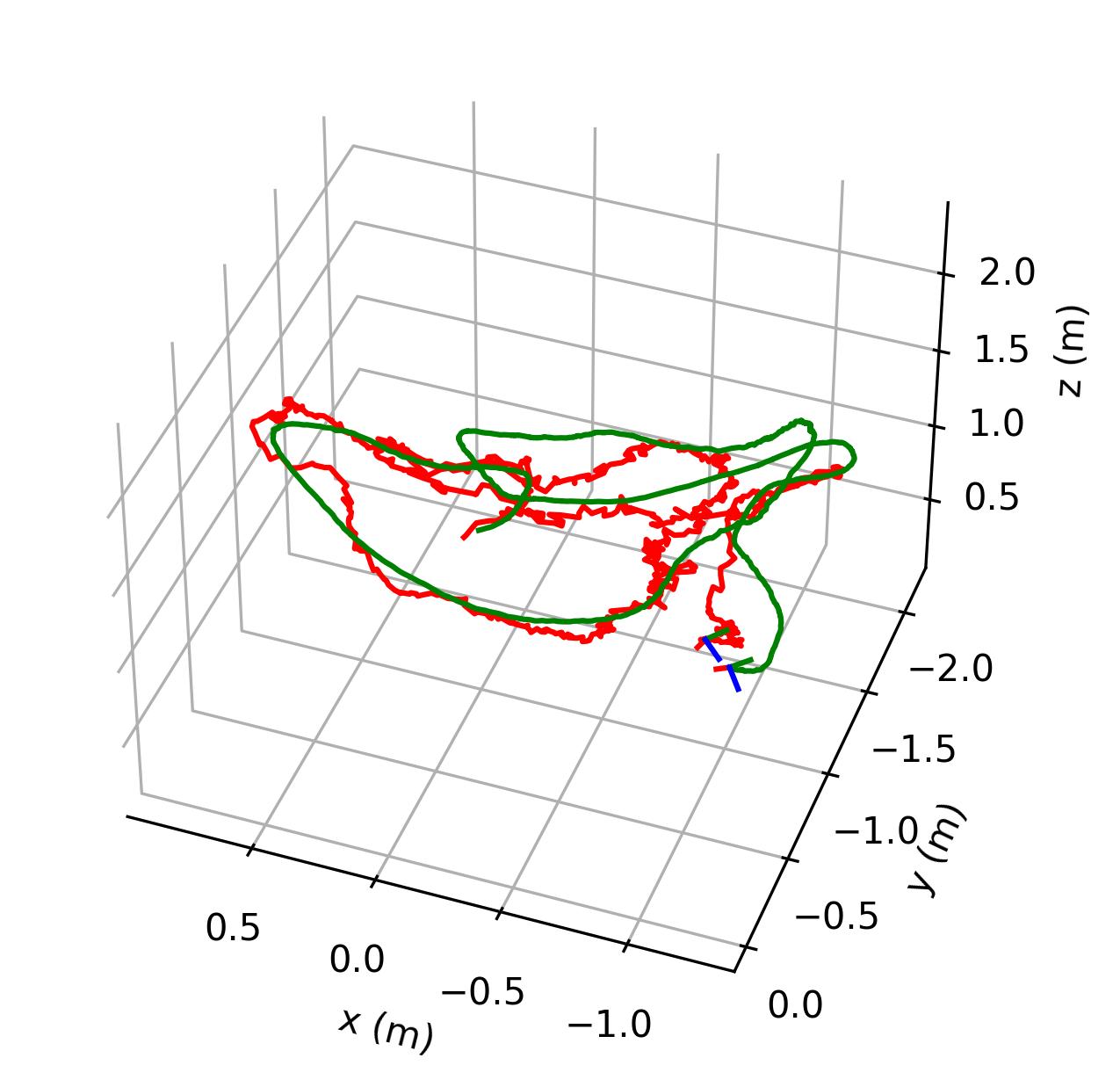}} &
\subfloat[]{\includegraphics[width = 0.20\textwidth]{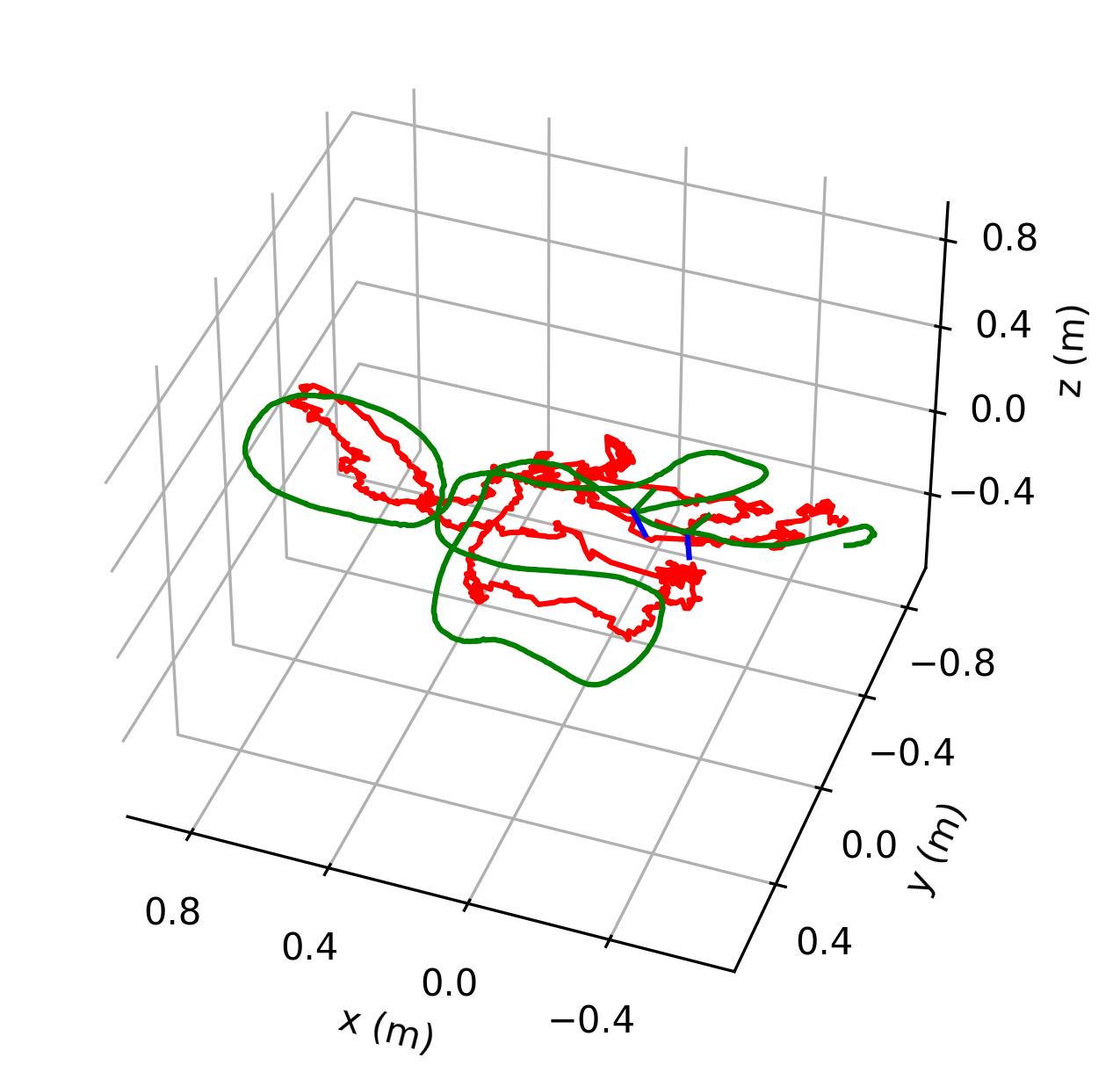}} \\[-3.61ex]
\end{tabular}
\caption{$\mathcal{SB}$-model Camera Trajectory visualization on Cambridge and 7Scenes dataset~\cite{glocker2013real, kendall2015posenet}. Each plot shows the camera trajectory (green for the ground truth and red for the prediction). From left to right, the testing sequences are KingsCollege-seq-02, StMarysChurch-seq-13, Office-seq-06 and Heads-seq-01.}
\label{fig:sb-trajectory}
\end{figure*}

%% file: figs/07_supp_fig-tsne-plots.tex
\begin{figure}[t]
\centering
\captionsetup[subfigure]{labelformat=empty}
\begin{center}
    \textit{Chess}
\end{center}
\begin{tabular}{cc}
    {\footnotesize \rotatebox{90}{\thinspace\thinspace$\mathcal{DA}$-model}}&\subfloat[]{\includegraphics[width = 0.85\columnwidth]{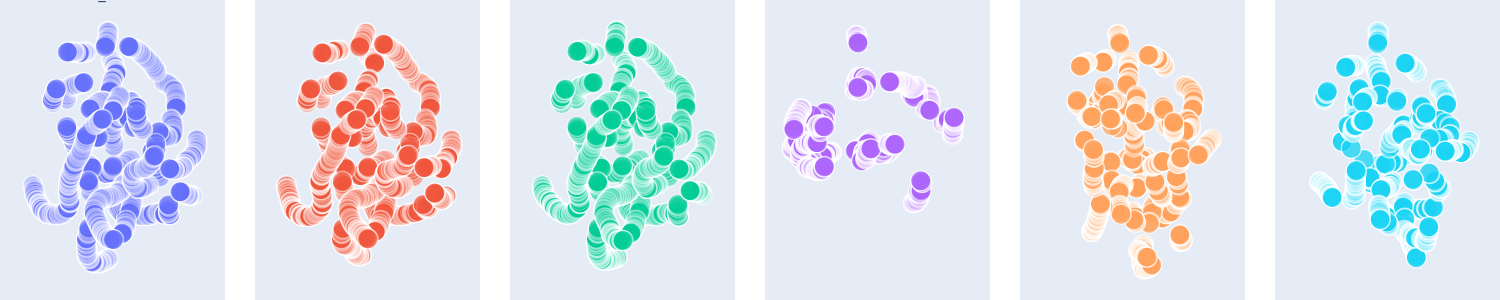}} \\[-2.5ex]
    {\footnotesize \rotatebox{90}{\thinspace\thinspace$\mathcal{SB}$-model}}&\subfloat[]{\includegraphics[width = 0.85\columnwidth]{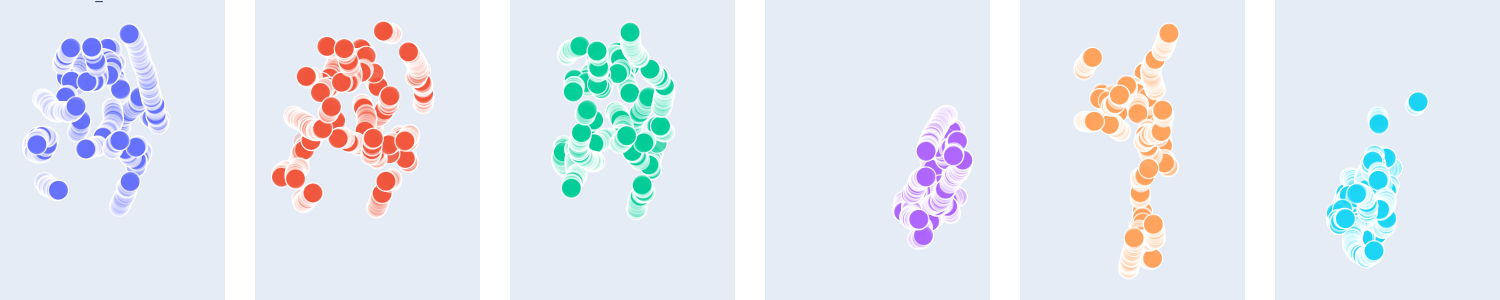}}
\end{tabular}
\begin{center}
    \textit{Redkitchen}
\end{center}
\begin{tabular}{cc}
    {\footnotesize \rotatebox{90}{\thinspace\thinspace$\mathcal{DA}$-model}}&\subfloat[]{\includegraphics[width = 0.85\columnwidth]{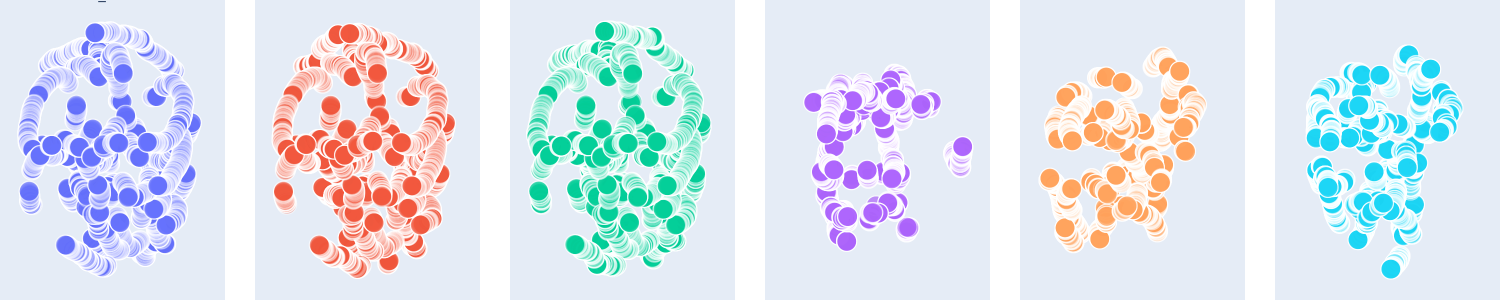}} \\[-2.5ex]
    {\footnotesize \rotatebox{90}{\thinspace\thinspace$\mathcal{SB}$-model}}&\subfloat[]{\includegraphics[width = 0.85\columnwidth]{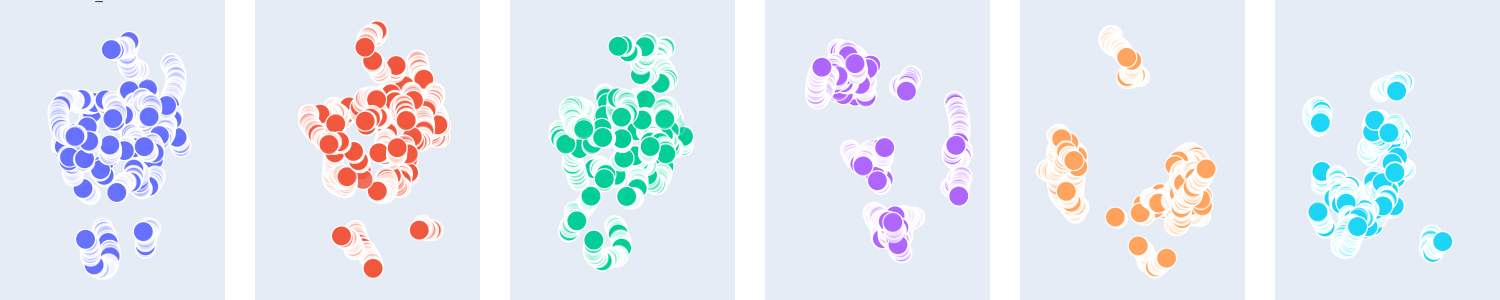}}
\end{tabular}
\caption{Visualization of the embeddings produced by $\mathcal{SB}$-model and $\mathcal{DA}$-model for the test split of \textit{Chess} and \textit{Redkitchen} scene in 2 dimensions by using t-SNE~\cite{van2008visualizing, plotly} plots. Each subplot represents the latent embedding space of different domains. From the $1^{st}$ row, it can be seen that the $\mathcal{DA}$-model produces similar embeddings to seen and unseen domains while the $2^{nd}$ row, shows that $\mathcal{SB}$-model produces dissimilar embeddings. From left to right, the domains are real, foggy, night, mosaic, udnie, and starry.}

\label{fig:tsne-plots-additional}
\end{figure}

%% file: supplementary/E_seen_domains.tex
\section{Results from Scenes seen during Training}
\input{tables/08_supp_table-FogNight.tex}

As given in~\ref{subsec:supp_fog_night_view_aug}, the original datasets are augmented to foggy and night view images. Collectively, the original dataset i.e, (real domain) and augmented domains(foggy, night) take part in the domain adaptive training framework as explained in the main text. As per one of the goals of the domain adaptive training framework, the $\mathcal{DA}$-model is expected to produce similar prediction performance for all these domains. Table~\ref{tab:08_supp_table-FogNight} shows the full results for Table \externalfigtab{5}(main text) as presented in the manuscript. It presents the scene-wise and domain-wise predictions for both datasets along with averages. In the detailed results, the $\mathcal{DA}$-model produces similar predictions across all three domains whereas the  $\mathcal{SB}$-model and the MS-Transformer prediction results seem to be dissimilar. Also, variance in prediction for different domains is higher in MS-Transformer when compared to $\mathcal{SB}$-model.

%% file: tables/08_supp_table-FogNight.tex
\begin{table*}[ht!]
\caption{Median errors for multiple indoor and outdoor scenes on domains used for training: Real, Foggy, Night. Full Version of Table 5(main text).}
\label{tab:08_supp_table-FogNight}
\begin{adjustbox}{width=.7\textwidth,center}

\begin{tabular}{|c|c|cc|cc|cc|cc|}
\hline
\multicolumn{2}{|c|}{} &
  \multicolumn{2}{c|}{\textbf{Real}} &
  \multicolumn{2}{c|}{\textbf{Foggy}} &
  \multicolumn{2}{c|}{\textbf{Night}} &
  \multicolumn{2}{c|}{\textit{\textbf{Average}}} \\ \cline{3-10}
\multicolumn{2}{|c|}{\multirow{-2}{*}{\textbf{Methods / Scenes}}} &
  \textbf{T(m)} &
  \textbf{R(deg)} &
  \textbf{T(m)} &
  \textbf{R(deg)} &
  \textbf{T(m)} &
  \textbf{R(deg)} &
  \textbf{T(m)} &
  \textbf{R(deg)} \\
  \hline
 &
  KingsCollege &
  0.83 &
  1.47 &
  1.27 &
  3.47 &
  2.03 &
  3.72 &
  1.38 &
  2.89 \\
 &
  OldHospital &
  1.81 &
  2.39 &
  4.65 &
  6.02 &
  3.39 &
  3.87 &
  3.29 &
  4.09 \\
 &
  ShopFacade &
  0.86 &
  3.07 &
  1.27 &
  4.43 &
  1.20 &
  5.78 &
  1.11 &
  4.43 \\
 &
  StMarysChurch &
  1.62 &
  3.99 &
  2.39 &
  22.36 &
  2.26 &
  9.39 &
  2.09 &
  11.92 \\
 &
  \cellcolor[HTML]{B7B7B7}\textit{\textbf{Average}} &
  \cellcolor[HTML]{B7B7B7}1.28 &
  \cellcolor[HTML]{B7B7B7}2.73 &
  \cellcolor[HTML]{B7B7B7}2.40 &
  \cellcolor[HTML]{B7B7B7}9.07 &
  \cellcolor[HTML]{B7B7B7}2.22 &
  \cellcolor[HTML]{B7B7B7}5.69 &
  \cellcolor[HTML]{B7B7B7}1.97 &
  \cellcolor[HTML]{B7B7B7}5.83 \\
 &
  Chess &
  0.11 &
  4.66 &
  0.14 &
  8.42 &
  0.13 &
  7.03 &
  0.12 &
  6.70 \\
 &
  Office &
  0.17 &
  5.66 &
  0.20 &
  7.71 &
  0.26 &
  10.03 &
  0.21 &
  7.80 \\
 &
  Pumpkin &
  0.18 &
  4.44 &
  0.23 &
  5.90 &
  0.31 &
  5.90 &
  0.24 &
  5.42 \\
 &
  Stairs &
  0.26 &
  8.45 &
  0.55 &
  9.83 &
  0.30 &
  10.31 &
  0.37 &
  9.53 \\
\multirow{-10}{*}{\textbf{{\rotatebox{90}{MS-Transformer}}}} &
  \cellcolor[HTML]{B7B7B7}\textit{\textbf{Average}} &
  \cellcolor[HTML]{B7B7B7}0.18 &
  \cellcolor[HTML]{B7B7B7}5.80 &
  \cellcolor[HTML]{B7B7B7}0.28 &
  \cellcolor[HTML]{B7B7B7}7.97 &
  \cellcolor[HTML]{B7B7B7}0.25 &
  \cellcolor[HTML]{B7B7B7}8.32 &
  \cellcolor[HTML]{B7B7B7}0.24 &
  \cellcolor[HTML]{B7B7B7}7.36 \\
  \hline
 &
  KingsCollege &
  0.84 &
  2.58 &
  1.43 &
  2.77 &
  2.32 &
  4.21 &
  1.53 &
  3.19 \\
 &
  OldHospital &
  1.63 &
  3.20 &
  4.08 &
  3.72 &
  3.29 &
  4.59 &
  3.00 &
  3.84 \\
 &
  ShopFacade &
  0.76 &
  3.98 &
  1.66 &
  4.47 &
  1.94 &
  6.16 &
  1.45 &
  4.87 \\
 &
  StMarysChurch &
  1.28 &
  4.47 &
  1.97 &
  5.67 &
  2.44 &
  5.92 &
  1.89 &
  5.36 \\
 &
  \cellcolor[HTML]{B7B7B7}\textit{\textbf{Average}} &
  \cellcolor[HTML]{B7B7B7}1.13 &
  \cellcolor[HTML]{B7B7B7}3.56 &
  \cellcolor[HTML]{B7B7B7}2.28 &
  \cellcolor[HTML]{B7B7B7}4.16 &
  \cellcolor[HTML]{B7B7B7}2.50 &
  \cellcolor[HTML]{B7B7B7}5.22 &
  \cellcolor[HTML]{B7B7B7}1.97 &
  \cellcolor[HTML]{B7B7B7}4.31 \\
 &
  Chess &
  0.11 &
  5.34 &
  0.19 &
  8.29 &
  0.17 &
  8.32 &
  0.16 &
  7.32 \\
 &
  Office &
  0.18 &
  7.22 &
  0.23 &
  8.34 &
  0.32 &
  11.52 &
  0.24 &
  9.03 \\
 &
  Pumpkin &
  0.24 &
  5.47 &
  0.35 &
  8.23 &
  0.34 &
  6.84 &
  0.31 &
  6.85 \\
 &
  Stairs &
  0.26 &
  11.25 &
  0.38 &
  11.37 &
  0.32 &
  11.57 &
  0.32 &
  11.40 \\
\multirow{-10}{*}{\textbf{{\rotatebox{90}{$\mathcal{SB}$-model(Ours)}}}} &
  \cellcolor[HTML]{B7B7B7}\textit{\textbf{Average}} &
  \cellcolor[HTML]{B7B7B7}0.20 &
  \cellcolor[HTML]{B7B7B7}7.32 &
  \cellcolor[HTML]{B7B7B7}0.29 &
  \cellcolor[HTML]{B7B7B7}9.06 &
  \cellcolor[HTML]{B7B7B7}0.29 &
  \cellcolor[HTML]{B7B7B7}9.56 &
  \cellcolor[HTML]{B7B7B7}0.26 &
  \cellcolor[HTML]{B7B7B7}8.65 \\
  \hline
 &
  KingsCollege &
  0.74 &
  2.81 &
  0.85 &
  2.61 &
  0.88 &
  3.07 &
  0.82 &
  2.33 \\
 &
  OldHospital &
  1.88 &
  2.97 &
  2.54 &
  3.29 &
  2.02 &
  3.07 &
  2.15 &
  2.87 \\
 &
  ShopFacade &
  0.84 &
  3.71 &
  0.72 &
  3.52 &
  0.79 &
  3.32 &
  0.78 &
  2.83 \\
 &
  StMarysChurch &
  1.31 &
  4.29 &
  1.36 &
  5.05 &
  1.28 &
  4.48 &
  1.32 &
  3.78 \\
 &
  \cellcolor[HTML]{B7B7B7}\textit{\textbf{Average}} &
  \cellcolor[HTML]{B7B7B7}1.19 &
  \cellcolor[HTML]{B7B7B7}3.45 &
  \cellcolor[HTML]{B7B7B7}1.37 &
  \cellcolor[HTML]{B7B7B7}3.62 &
  \cellcolor[HTML]{B7B7B7}1.24 &
  \cellcolor[HTML]{B7B7B7}3.49 &
  \cellcolor[HTML]{B7B7B7}1.27 &
  \cellcolor[HTML]{B7B7B7}2.95 \\
 &
  Chess &
  0.11 &
  6.06 &
  0.11 &
  5.90 &
  0.11 &
  6.37 &
  0.11 &
  6.11 \\
 &
  Office &
  0.17 &
  8.34 &
  0.17 &
  8.17 &
  0.18 &
  8.30 &
  0.17 &
  8.27 \\
 &
  Pumpkin &
  0.21 &
  6.37 &
  0.22 &
  6.41 &
  0.23 &
  6.24 &
  0.22 &
  6.34 \\
 &
  Stairs &
  0.26 &
  12.21 &
  0.28 &
  11.97 &
  0.26 &
  12.37 &
  0.27 &
  12.18 \\
\multirow{-10}{*}{\textbf{{\rotatebox{90}{$\mathcal{DA}$-model(Ours)}}}} &
  \cellcolor[HTML]{B7B7B7}\textit{\textbf{Average}} &
  \cellcolor[HTML]{B7B7B7}0.19 &
  \cellcolor[HTML]{B7B7B7}8.25 &
  \cellcolor[HTML]{B7B7B7}0.19 &
  \cellcolor[HTML]{B7B7B7}8.11 &
  \cellcolor[HTML]{B7B7B7}0.20 &
  \cellcolor[HTML]{B7B7B7}8.32 &
  \cellcolor[HTML]{B7B7B7}0.19 &
  \cellcolor[HTML]{B7B7B7}8.23 \\
  \hline
\end{tabular}%

\end{adjustbox}

\end{table*}

%% file: supplementary/F_unseen_domains.tex
\section{Results from Scenes not seen during Training}
\input{tables/09_supp_table-MosaicUdnieStarry.tex}

Unseen domain augmentations are generated using the methodology and checkpoints presented in~\cite{rusty2018faststyletransfer}. To obtain mosaic, udnie and starry domains the following checkpoints are used\footnote{\url{https://github.com/rrmina/fast-neural-style-pytorch}}: \texttt{mosaic.pth}, \texttt{udnie.pth}, \texttt{starry.pth}. As another advantage of the domain adaptive framework, it is expected that the $\mathcal{DA}$-model should also work reasonably to unseen domains. Table~\ref{tab:09_supp_table-MosaicUdnieStarry} shows the full results for Table \externalfigtab{6}(main text) as presented in the manuscript. Similar to the previous section, it can be observed that the $\mathcal{DA}$-model outperforms other baseline methods when unseen domains are evaluated.

%% file: tables/09_supp_table-MosaicUdnieStarry.tex
\begin{table*}[hbt!]
\caption{Median errors for multiple indoor and outdoor scenes on domains not used for training: Mosaic, Starry, and Udnie. Full Version of Table 6(main text).}
\label{tab:09_supp_table-MosaicUdnieStarry}
\begin{adjustbox}{width=.7\textwidth,center}

\begin{tabular}{|c|c|cc|cc|cc|cc|}
\hline
\multicolumn{2}{|c|}{} &
  \multicolumn{2}{c|}{\textbf{Mosaic-style}} &
  \multicolumn{2}{c|}{\textbf{Udnie-style}} &
  \multicolumn{2}{c|}{\textbf{Starry-style}} &
  \multicolumn{2}{c|}{\textit{\textbf{Average}}} \\ \cline{3-10}
\multicolumn{2}{|c|}{\multirow{-2}{*}{\textbf{Methods / Scenes}}} &
  \textbf{T(m)} &
  \textbf{R(deg)} &
  \textbf{T(m)} &
  \textbf{R(deg)} &
  \textbf{T(m)} &
  \textbf{R(deg)} &
  \textbf{T(m)} &
  \textbf{R(deg)} \\
  \hline
 &
  KingsCollege &
  3.28 &
  4.44 &
  1.95 &
  3.18 &
  3.57 &
  7.28 &
  2.93 &
  4.97 \\
 &
  OldHospital &
  5.65 &
  6.02 &
  2.35 &
  3.80 &
  5.35 &
  8.82 &
  4.45 &
  6.21 \\
 &
  ShopFacade &
  2.11 &
  7.49 &
  1.13 &
  3.66 &
  1.45 &
  11.82 &
  1.56 &
  7.66 \\
 &
  StMarysChurch &
  10.52 &
  73.65 &
  7.34 &
  29.84 &
  18.19 &
  97.53 &
  12.02 &
  67.01 \\
 &
  \cellcolor[HTML]{B7B7B7}\textit{\textbf{Average}} &
  \cellcolor[HTML]{B7B7B7}5.39 &
  \cellcolor[HTML]{B7B7B7}22.90 &
  \cellcolor[HTML]{B7B7B7}3.19 &
  \cellcolor[HTML]{B7B7B7}10.12 &
  \cellcolor[HTML]{B7B7B7}7.14 &
  \cellcolor[HTML]{B7B7B7}31.36 &
  \cellcolor[HTML]{B7B7B7}5.24 &
  \cellcolor[HTML]{B7B7B7}21.46 \\
 &
  Chess &
  0.28 &
  17.80 &
  0.15 &
  9.00 &
  0.18 &
  11.86 &
  0.20 &
  12.88 \\
 &
  Office &
  0.55 &
  31.44 &
  0.26 &
  14.21 &
  0.30 &
  15.73 &
  0.37 &
  20.46 \\
 &
  Pumpkin &
  0.52 &
  20.16 &
  0.48 &
  19.90 &
  0.44 &
  17.03 &
  0.48 &
  19.03 \\
 &
  Stairs &
  0.60 &
  16.54 &
  0.37 &
  13.08 &
  0.36 &
  10.85 &
  0.44 &
  13.49 \\
\multirow{-10}{*}{\textbf{{\rotatebox{90}{MS-Transformer}}}} &
  \cellcolor[HTML]{B7B7B7}\textit{\textbf{Average}} &
  \cellcolor[HTML]{B7B7B7}0.49 &
  \cellcolor[HTML]{B7B7B7}21.49 &
  \cellcolor[HTML]{B7B7B7}0.32 &
  \cellcolor[HTML]{B7B7B7}14.05 &
  \cellcolor[HTML]{B7B7B7}0.32 &
  \cellcolor[HTML]{B7B7B7}13.87 &
  \cellcolor[HTML]{B7B7B7}0.37 &
  \cellcolor[HTML]{B7B7B7}16.47 \\
  \hline
 &
  KingsCollege &
  9.55 &
  8.70 &
  3.89 &
  5.58 &
  9.55 &
  13.87 &
  7.66 &
  9.38 \\
 &
  OldHospital &
  5.56 &
  6.29 &
  3.74 &
  4.75 &
  4.37 &
  4.53 &
  4.56 &
  5.19 \\
 &
  ShopFacade &
  4.90 &
  14.44 &
  2.42 &
  10.68 &
  2.84 &
  10.30 &
  3.39 &
  11.81 \\
 &
  StMarysChurch &
  19.78 &
  24.11 &
  5.17 &
  9.31 &
  20.46 &
  24.70 &
  15.14 &
  19.37 \\
 &
  \cellcolor[HTML]{B7B7B7}\textit{\textbf{Average}} &
  \cellcolor[HTML]{B7B7B7}9.95 &
  \cellcolor[HTML]{B7B7B7}13.38 &
  \cellcolor[HTML]{B7B7B7}3.81 &
  \cellcolor[HTML]{B7B7B7}7.58 &
  \cellcolor[HTML]{B7B7B7}9.30 &
  \cellcolor[HTML]{B7B7B7}13.35 &
  \cellcolor[HTML]{B7B7B7}7.69 &
  \cellcolor[HTML]{B7B7B7}11.44 \\
 &
  Chess &
  0.44 &
  12.87 &
  0.24 &
  8.42 &
  0.33 &
  12.15 &
  0.34 &
  11.15 \\
 &
  Office &
  0.65 &
  29.18 &
  0.39 &
  13.24 &
  0.41 &
  13.80 &
  0.48 &
  18.74 \\
 &
  Pumpkin &
  0.65 &
  32.62 &
  0.50 &
  25.98 &
  0.49 &
  22.24 &
  0.55 &
  26.94 \\
 &
  Stairs &
  0.55 &
  20.70 &
  0.34 &
  16.31 &
  0.64 &
  16.14 &
  0.51 &
  17.72 \\
\multirow{-10}{*}{\textbf{{\rotatebox{90}{$\mathcal{SB}$-model(Ours)}}}} &
  \cellcolor[HTML]{B7B7B7}\textit{\textbf{Average}} &
  \cellcolor[HTML]{B7B7B7}0.57 &
  \cellcolor[HTML]{B7B7B7}23.84 &
  \cellcolor[HTML]{B7B7B7}0.36 &
  \cellcolor[HTML]{B7B7B7}15.99 &
  \cellcolor[HTML]{B7B7B7}0.47 &
  \cellcolor[HTML]{B7B7B7}16.08 &
  \cellcolor[HTML]{B7B7B7}0.47 &
  \cellcolor[HTML]{B7B7B7}18.64 \\
  \hline
 &
  KingsCollege &
  2.89 &
  4.64 &
  2.40 &
  3.74 &
  2.18 &
  4.57 &
  2.49 &
  4.32 \\
 &
  OldHospital &
  5.07 &
  5.48 &
  4.19 &
  3.75 &
  3.63 &
  4.36 &
  4.29 &
  4.53 \\
 &
  ShopFacade &
  4.31 &
  10.29 &
  2.28 &
  8.49 &
  2.38 &
  8.47 &
  2.99 &
  9.08 \\
 &
  StMarysChurch &
  10.91 &
  12.21 &
  3.89 &
  7.91 &
  4.35 &
  8.29 &
  6.39 &
  9.47 \\
 &
  \cellcolor[HTML]{B7B7B7}\textit{\textbf{Average}} &
  \cellcolor[HTML]{B7B7B7}5.80 &
  \cellcolor[HTML]{B7B7B7}8.15 &
  \cellcolor[HTML]{B7B7B7}3.19 &
  \cellcolor[HTML]{B7B7B7}5.97 &
  \cellcolor[HTML]{B7B7B7}3.13 &
  \cellcolor[HTML]{B7B7B7}6.42 &
  \cellcolor[HTML]{B7B7B7}4.04 &
  \cellcolor[HTML]{B7B7B7}6.85 \\
 &
  Chess &
  0.28 &
  9.89 &
  0.16 &
  6.87 &
  0.17 &
  7.12 &
  0.20 &
  7.96 \\
 &
  Office &
  0.48 &
  16.97 &
  0.26 &
  10.51 &
  0.27 &
  10.05 &
  0.33 &
  12.51 \\
 &
  Pumpkin &
  0.49 &
  15.23 &
  0.38 &
  9.96 &
  0.37 &
  10.02 &
  0.41 &
  11.74 \\
 &
  Stairs &
  0.43 &
  17.16 &
  0.31 &
  15.26 &
  0.38 &
  14.87 &
  0.37 &
  15.76 \\
\multirow{-10}{*}{\textbf{{\rotatebox{90}{$\mathcal{DA}$-model(Ours)}}}} &
  \cellcolor[HTML]{B7B7B7}\textit{\textbf{Average}} &
  \cellcolor[HTML]{B7B7B7}0.42 &
  \cellcolor[HTML]{B7B7B7}14.81 &
  \cellcolor[HTML]{B7B7B7}0.28 &
  \cellcolor[HTML]{B7B7B7}10.65 &
  \cellcolor[HTML]{B7B7B7}0.29 &
  \cellcolor[HTML]{B7B7B7}10.51 &
  \cellcolor[HTML]{B7B7B7}0.33 &
  \cellcolor[HTML]{B7B7B7}11.99 \\
  \hline
  
\end{tabular}%
\end{adjustbox}
\end{table*}

%% file: supplementary/G_CNN_backbones.tex
\section{CNN Backbones}
\input{tables/10_supp_table_cnn_backbone.tex}

\subsection{Investigated lightweight backbones}
Analyzed CNN backbones are the following: MobileNetV3-Large~\cite{howard2019searching}, MobileNetV3-Small~\cite{howard2019searching}, ShuffleNet V2 0.5$\times$~\cite{ma2018shufflenet}, ShuffleNet V2 1.0$\times$~\cite{ma2018shufflenet}, ShuffleNet V2 1.5$\times$~\cite{ma2018shufflenet}, MnasNet 0.5$\times$~\cite{tan2019mnasnet}, MnasNet 1.0$\times$~\cite{tan2019mnasnet}, and EfficientNet-B0~\cite{tan2019efficientnet}. Here, ShuffleNet V2 represents the next generation of ShuffleNet~\cite{zhang2018shufflenet} that exploited pointwise group convolution and channel shuffle for efficient operations. The MnasNet considered a new approach, mobile neural architecture search(MNAS), to find a low latency model. The family of EfficientNets are enabled by the compound scaling method investigated by Tan et al~\cite{tan2019efficientnet}.

\subsection{Impact of CNN backbones on \texorpdfstring{$\mathcal{DA}$-}-model}
A key contribution of our work is to maintain the lightweight of the neural network for the pose regression task. With that in consideration, multiple lightweight CNN backbone's impacts are studied with the proposed domain adaptive training framework. Table~\ref{tab:10_supp_table_cnn_backbone} demonstrates the detailed evaluation of localization performance in \textit{ShopFacade} outdoor scene, \textit{Chess} indoor scene and computational properties for various CNN backbones. We were able to achieve a better trade-off between model performance and latency requirements with MobileNetV3-Large, thus it was the architecture chosen for the entirety of our study.

%% file: tables/10_supp_table_cnn_backbone.tex
\begin{table*}[ht]
\caption{Various lightweight CNN backbones impact on the proposed domain adaptive framework's performance. MobileNetV3-Large demonstrates the better trade-off over other lightweight CNN backbones considered.}
\label{tab:10_supp_table_cnn_backbone}
\resizebox{\textwidth}{!}{%
\begin{tabular}{|c|cc|cc|c|c|c|c|}
\hline
\multirow{2}{*}{\textbf{Backbones}} &
  \multicolumn{2}{c|}{\textbf{ShopFacade}} &
  \multicolumn{2}{c|}{\textbf{Chess}} &
  \multirow{2}{*}{\textbf{FLOPs (G)}} &
  \multirow{2}{*}{\textbf{\begin{tabular}[c]{@{}c@{}}Activation \\ (millions)\end{tabular}}} &
  \multirow{2}{*}{\textbf{\begin{tabular}[c]{@{}c@{}}Params \\ (millions)\end{tabular}}} &
  \multirow{2}{*}{\textbf{Memory (MB)}} \\
                            & \textbf{T(m)} & \textbf{R(deg)} & \textbf{T(m)} & \textbf{R(deg)} &       &     &       &      \\
\hline
\textbf{MobileNetV3-Large~\cite{howard2019searching}}  & 0.84          & 3.71            & 0.11          & 6.06            & 0.237 & 4.4 & 3.847 & 14.2 \\
\textbf{MobileNetV3-Small~\cite{howard2019searching}}  & 1.01          & 4.56            & 0.12          & 6.63            & 0.062 & 1.4 & 1.246 & 5.1  \\
\textbf{ShuffleNet V2 0.5×~\cite{ma2018shufflenet}} & 0.92          & 5.32            & 0.11          & 6.85            & 0.045 & 1   & 0.89  & 3.7  \\
\textbf{ShuffleNet V2 1.0×~\cite{ma2018shufflenet}} & 0.87          & 5.63            & 0.11          & 7.00            & 0.154 & 1.9 & 1.802 & 7.4  \\
\textbf{ShuffleNet V2 1.5×~\cite{ma2018shufflenet}} & 0.75          & 5.48            & 0.13          & 8.10            & 0.309 & 2.8 & 3.027 & 12.3 \\
\textbf{MnasNet 0.5x~\cite{tan2019mnasnet}}       & 1.80          & 8.13            & 0.13          & 7.41            & 0.119 & 3.1 & 1.617 & 6.7  \\
\textbf{MnasNet 1.0x~\cite{tan2019mnasnet}}       & 8.68          & 6.51            & 0.11          & 6.64            & 0.341 & 5.5 & 3.782 & 15.4 \\
\textbf{EfficientNet-B0~\cite{tan2019efficientnet}}    & 0.83          & 4.47            & 0.10          & 6.52            & 0.421 & 6.7 & 5.968 & 24.2 \\
\hline
\end{tabular}%
}
\end{table*}

%% file: supplementary/H_review_sfm_IR.tex
\section{Review of Structure and IR based Methods}
\subsection{Structure Based Localization}
\label{subsec:related-sfm}

In structure-based localization methods, an image is processed to identify important points and corresponding associated features (descriptors) that are then used for feature matching based on visual similarity. To perform this 2D-2D correspondence, keypoint descriptors like SURF\cite{bay2008speeded}, SIFT\cite{lowe2004distinctive}, FAST\cite{philbin2007object}, and ORB\cite{rublee2011orb} are used. The results of the feature correspondence are used to acquire an essential matrix followed by rejecting outliers using RANSAC (Iterative model conditioning approach). To create a 3D understanding of the scene, epipolar geometric properties \textcolor{black}{are leveraged \cite{hartley2003multiple}}. The global 3D scene is obtained by using keypoint correspondence\cite{fischler1981random, hartley1997defense, nister2004efficient}. The final camera pose is obtained by using a perspective-n-point(PnP)~\cite{fischler1981random} method. Compared to the methods to be presented in this paper, traditional structure-based localization methods are computationally heavy, with slow inference time.

To replace the sub-components that are part of the aforementioned structure-based localization traditional pipeline, recently, deep learning-based approaches have been proposed. D2Net\cite{dusmanu2019d2} implements feature correspondence by separating detection and description. SuperGlue\cite{sarlin2020superglue} uses the graph representation of a scene to leverage graph neural networks to perform detection and description concurrently. Transformer-based methods like LoFTR\cite{sun2021loftr} and TransforMatcher\cite{kim2022transformatcher} carry out feature matching as well as semantic correspondence matching. Finally, methods such as DSAC \cite{brachmann2017dsac} and DSAC++ \cite{fischler1981random} propose a differentiable version of the original RANSAC algorithm.

\subsection{Image Retrieval}
\label{subsec:related-IR}

Image retrieval (IR) methods search for images on a large database and then retrieve images similar to the target image for pose estimation. The database, part of IR methods, consists of multiple images of the scene in which the camera pose is known. Intuitively, IR methods search for images that have poses closer to the query image. The similarity metrics in which the search is performed are usually performed by using features obtained with global descriptors enable by methods such as DenseVLAD\cite{torii201524}, NetVLAD\cite{arandjelovic2016netvlad}.

Poses retrieved from the database can be considered priors in the pose estimation process. This characteristic is powerful to scale this pipeline to both small-scale and large-scale scenes. However, it is important to account that the sample density of poses in the database significantly affects its accuracy. As an alternative to obtaining the final pose using IR, for example, the camera pose can be determined by the weighted average sum of poses from the top-$k$ most similar images retrieved from the database. If the density is high enough, another alternative is to consider the pose of the top-$1$ image as the pose of the query image. IR methods need the preparation of a carefully designed database and, also, the storage and search of these pose-labelled databases for inference. Methods that utilize IR encodings \cite{sarlin2019coarse, taira2018inloc} achieve state-of-the-art accuracy on large-scale datasets for visual localization. These methods are also inspired and hold a similar methodology with hierarchical localization approaches \cite{noh2017large, sattler2019understanding}.

%% file: supplementary/I_Gan_samples.tex
\section{Additional Samples of GAN generated images}
\input{./figs/08_supp_fig-gan-samples-full.tex}

Figure~\ref{fig:gan-samples-full} shows the samples for augmented images in different scenes. It can be seen that the augmentations that take part in the domain adaptive training framework are relatively weak augmentations(foggy, night) while the unseen domains used for evaluations are strong augmentations, i.e. augmentations with a significant deviation from the image's real distribution.

%% file: figs/08_supp_fig-gan-samples-full.tex
\begin{figure*}[ht]
\captionsetup[subfigure]{labelformat=empty}
\centering
\newcommand{\sampleganwidth}{0.16}
\newcommand{\samplegansepwidth}{0.1}

\begin{tabular}{c@{ }c@{ }c@{ }c@{ }c@{ }c@{ }c}
& Real & Foggy~\cite{pizzati2021manifest} & Night~\cite{pizzati2021comogan} & Mosaic~\cite{rusty2018faststyletransfer} & Udnie~\cite{rusty2018faststyletransfer} & Starry~\cite{rusty2018faststyletransfer} \\
{\rotatebox{90}{\qquad \quad Heads}} &
\subfloat[]{\includegraphics[width = \sampleganwidth \textwidth]{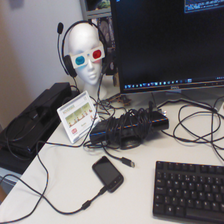}}&
\subfloat[]{\includegraphics[width = \sampleganwidth \textwidth]{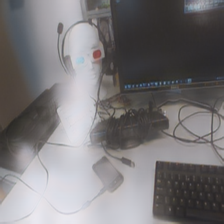}} &
\subfloat[]{\includegraphics[width = \sampleganwidth \textwidth]{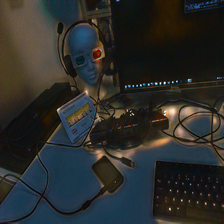}} &
\subfloat[]{\includegraphics[width = \sampleganwidth \textwidth]{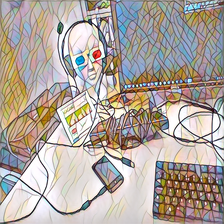}} &
\subfloat[]{\includegraphics[width = \sampleganwidth \textwidth]{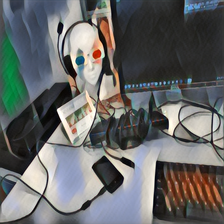}} &
\subfloat[]{\includegraphics[width = \sampleganwidth \textwidth]{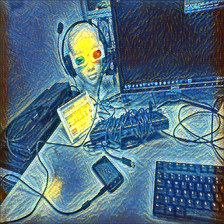}} \\ [-2.5ex]
{\rotatebox{90}{\qquad \quad Chess}} &
\subfloat[]{\includegraphics[width = \sampleganwidth \textwidth]{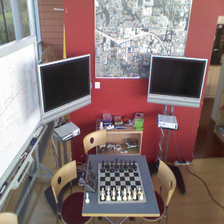}}&
\subfloat[]{\includegraphics[width = \sampleganwidth \textwidth]{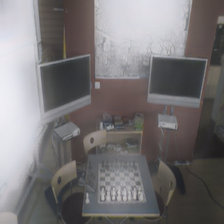}} &
\subfloat[]{\includegraphics[width = \sampleganwidth \textwidth]{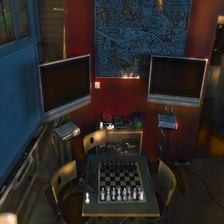}} &
\subfloat[]{\includegraphics[width = \sampleganwidth \textwidth]{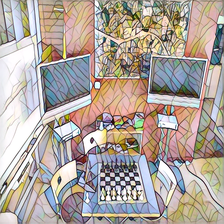}} &
\subfloat[]{\includegraphics[width = \sampleganwidth \textwidth]{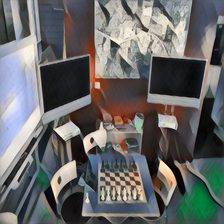}} &
\subfloat[]{\includegraphics[width = \sampleganwidth \textwidth]{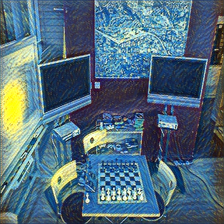}} \\ [-2.5ex]
{\rotatebox{90}{\quad ShopFacade}} &
\subfloat[]{\includegraphics[width = \sampleganwidth \textwidth]{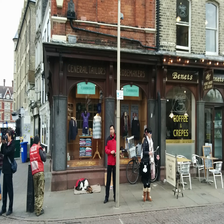}}&
\subfloat[]{\includegraphics[width = \sampleganwidth \textwidth]{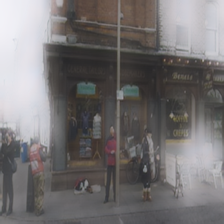}} &
\subfloat[]{\includegraphics[width = \sampleganwidth \textwidth]{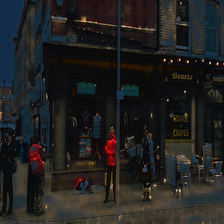}} &
\subfloat[]{\includegraphics[width = \sampleganwidth \textwidth]{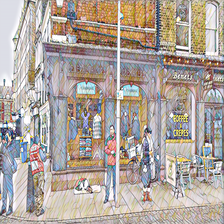}} &
\subfloat[]{\includegraphics[width = \sampleganwidth \textwidth]{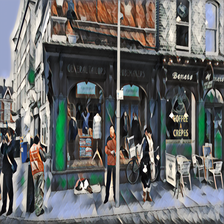}} &
\subfloat[]{\includegraphics[width = \sampleganwidth \textwidth]{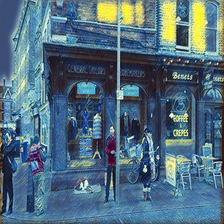}} \\ [-2.5ex]
{\rotatebox{90}{\quad KingsCollege}} &
\subfloat[]{\includegraphics[width = \sampleganwidth \textwidth]{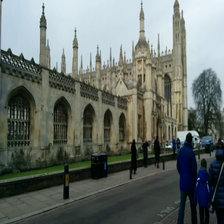}}&
\subfloat[]{\includegraphics[width = \sampleganwidth \textwidth]{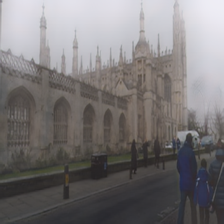}} &
\subfloat[]{\includegraphics[width = \sampleganwidth \textwidth]{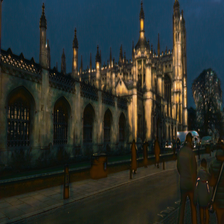}} &
\subfloat[]{\includegraphics[width = \sampleganwidth \textwidth]{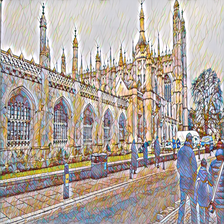}} &
\subfloat[]{\includegraphics[width = \sampleganwidth \textwidth]{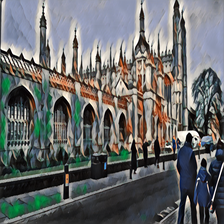}} &
\subfloat[]{\includegraphics[width = \sampleganwidth \textwidth]{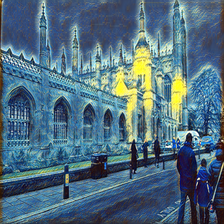}} \\ [-2.5ex]

\end{tabular}

\caption{Samples of GAN generated images. (a) Real image (b) Foggy image (c) Night image (d) Mosaic-style image (e) Udnie-style image (f) Starry-style image}
\label{fig:gan-samples-full}
\end{figure*}